\documentclass[journal, compsoc]{IEEEtran}
\usepackage{graphicx}
\usepackage{cite}
\usepackage{amssymb}
\usepackage{amsmath}
\usepackage{amsfonts}
\usepackage{enumerate}
\usepackage{booktabs}
\usepackage{algorithm}
\usepackage{algorithmic}
\usepackage{color}
\usepackage{colortbl}
\usepackage{bm}
\usepackage{bbm}
\usepackage{multirow}
\usepackage{array}
\usepackage{pifont}
\usepackage{ragged2e}

\usepackage{subfigure}

\usepackage{hyperref}
\hypersetup{colorlinks=true}

\usepackage{longtable}  
\usepackage{caption}    
\usepackage{makecell} 
\usepackage{hyperref}
\usepackage{threeparttable}
\usepackage{graphicx}

\usepackage[table]{xcolor} 

\begin{document}
%
\title{HyperSIGMA: Hyperspectral Intelligence Comprehension Foundation Model}

\author{

Di Wang$^*$, 
Meiqi Hu$^*$, 
Yao Jin$^*$,
Yuchun Miao$^*$, 
Jiaqi Yang$^*$, 
Yichu Xu$^*$, 
Xiaolei Qin$^*$, 
Jiaqi Ma$^*$, \\
Lingyu Sun$^*$, 
Chenxing Li$^*$, 
Chuan Fu, 
Hongruixuan Chen, 
Chengxi Han$\dagger$, 
Naoto Yokoya,\\~\IEEEmembership{Member,~IEEE,}
Jing Zhang$\dagger$,~\IEEEmembership{Senior Member,~IEEE,}
Minqiang Xu, 
Lin Liu,
Lefei Zhang,\\~\IEEEmembership{Senior Member,~IEEE,}
Chen Wu$\dagger$,~\IEEEmembership{Member,~IEEE,}
Bo Du$\dagger$,~\IEEEmembership{Senior Member,~IEEE,}\\
Dacheng Tao,~\IEEEmembership{Fellow,~IEEE}
and Liangpei Zhang$\dagger$,~\IEEEmembership{Fellow,~IEEE}

\thanks{D. Wang, M. Hu, Y. Jin, Y. Miao, J. Yang, Y. Xu, X. Qin, J. Ma, L. Sun, C. Li, C. Han, J. Zhang, L. Zhang, C. Wu, B. Du and L. Zhang are with the Wuhan University, China; C. Fu is with the Chongqing University, China; H. Chen and N. Yokoya are with The University of Tokyo, Japan; M. Xu and L.Liu are with the National Engineering Research Center of Speech and Language Information Processing, China; D. Tao is with the Nanyang Technological University, Singapore. (\{d\_wang, meiqi.hu, yao.jin, miaoyuchun, jqyang, xuyichu, qinxlei, jiaqima, lingyu.sun, chenxing.li, chengxihan, zhanglefei, chen.wu, dubo, zlp62\}@whu.edu.cn; fuchuan@cqu.edu.cn; \{qschrx, jingzhang.cv, dacheng.tao\}@gmail.com; yokoya@k.u-tokyo.ac.jp; \{mqxu7, linliu\}@iflytek.com).\\ $^*$: Equal contribution; $\dagger$: Corresponding author.}
}

\markboth{Journal of \LaTeX\ Class Files,~Vol.~X, No.~X, XXX XXX}%
{Shell \MakeLowercase{\textit{et al.}}: Bare Demo of IEEEtran.cls for IEEE Journals}



\IEEEtitleabstractindextext{
\begin{abstract}
\justifying
Accurate hyperspectral image (HSI) interpretation is critical for providing valuable insights into various earth observation-related applications such as urban planning, precision agriculture, and environmental monitoring. However, existing HSI processing methods are predominantly task-specific and scene-dependent, which severely limits their ability to transfer knowledge across tasks and scenes, thereby reducing the practicality in real-world applications. To address these challenges, we present HyperSIGMA, a vision transformer-based foundation model that unifies HSI interpretation across tasks and scenes, scalable to over one billion parameters. To overcome the spectral and spatial redundancy inherent in HSIs, we introduce a novel sparse sampling attention (SSA) mechanism, which effectively promotes the learning of diverse contextual features and serves as the basic block of HyperSIGMA. HyperSIGMA integrates spatial and spectral features using a specially designed spectral enhancement module. In addition, we construct a large-scale hyperspectral dataset, HyperGlobal-450K, for pre-training, which contains about 450K hyperspectral images, significantly surpassing existing datasets in scale. Extensive experiments on various high-level and low-level HSI tasks demonstrate HyperSIGMA's versatility and superior representational capability compared to current state-of-the-art methods. Moreover, HyperSIGMA shows significant advantages in scalability, robustness, cross-modal transferring capability, real-world applicability, and computational efficiency. The code and models will be released at \href{https://github.com/WHU-Sigma/HyperSIGMA}{HyperSIGMA}.

\end{abstract}

\begin{IEEEkeywords}
Remote sensing, Hyperspectral image, Foundation model, Attention, Vision transformer, Large-scale dataset
\end{IEEEkeywords}
}

\captionsetup{
    font=footnotesize,
    textfont=sf,
  }

\maketitle

\section{Introduction}

\begin{figure}[!tb]
\centering
\includegraphics[width=\linewidth]{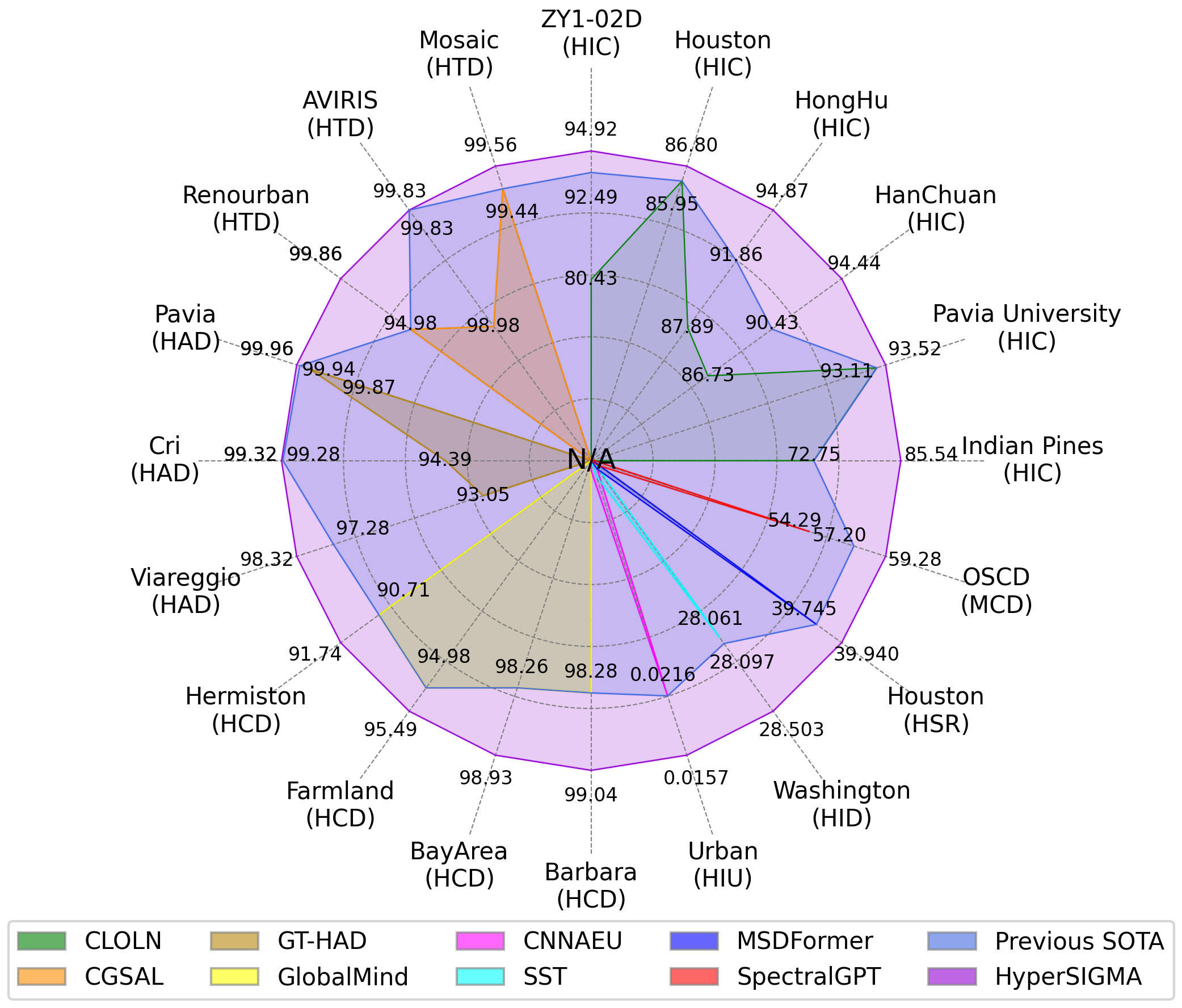}
\caption{
HyperSIGMA offers a universal solution for HSI processing, demonstrating superior performance across 20 datasets, including both high-level and low-level hyperspectral tasks, as well as multispectral scenes. It outperforms advanced models like SpectralGPT, even those specifically designed for these tasks. HIC: Hyperspectral Image Classification. HTD: Hyperspectral Target Detection. HAD: Hyperspectral Anomaly Detection. HCD: Hyperspectral Change Detection. HIU: Hyperspectral Image Unmixing. HID: Hyperspectral Image Denoising. HSR: Hyperspectral Super-Resolution. MCD: Multispectral Change Detection.
}
\label{fig: radar}
\end{figure}

\IEEEPARstart{T}{he} rapid advancements in aeronautical engineering, sensor technology, and computer science have made it possible to acquire massive hyperspectral remote sensing images (hereafter referred to as HSIs) with fine spectral resolution \cite{imaging_spectrometry,earth_data_sys_sci,spectralgpt}. HSIs cover a spectral range from visible near-infrared to short-wave and mid-infrared, capturing target features through continuous and fine spectral bands. This results in nearly continuous spectral curves that provide detailed surface information, enabling the distinction of subtle spectral differences between substances for precise land cover interpretation  \cite{rs_hsi_cls}. Hyperspectral imagery significantly enhances our capability for comprehensive, accurate, and timely earth observation and monitoring \cite{rs_globalwaste,hsi_forest_manage,hsi_monitor_biodiversity}. It offers crucial scientific insights and decision-making support for fields such as urban planning \cite{hsi_urban_plan}, precision agriculture \cite{agriculture_1}, and environmental monitoring \cite{hsi_env_monitor}.

\begin{figure}[!tb]
\centering
\includegraphics[width=0.75\linewidth]{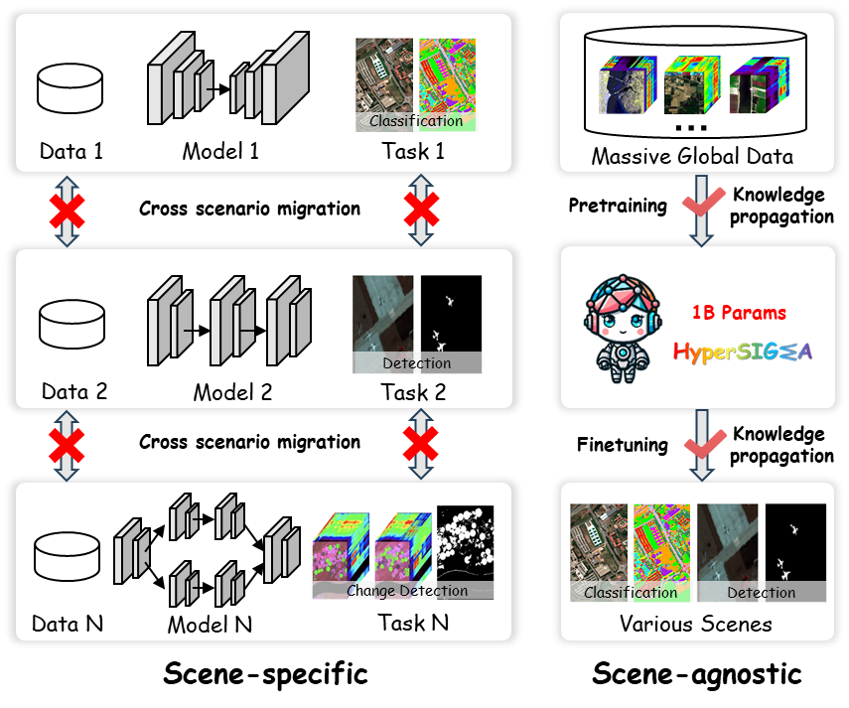}
\caption{
Previous HSI models are trained separately on different scenes, limiting cross-scene knowledge transfer. In contrast, our model acquires universal, scene-agnostic knowledge through pre-training with a large dataset of global HSIs, enabling effective transfer to various scenes through fine-tuning.
}
\label{fig: different_patterns}
\end{figure}

The challenges of processing HSIs primarily stem from their inherent characteristics: \textit{high dimensionality, data redundancy, and spatial variability} \cite{ica_rgf}. High dimensionality often leads to the \textit{Hughes phenomenon} \cite{hughes}, where machine learning models overfit as the number of channels increases, especially with limited training samples. Data redundancy, both spectral and spatial, results in unnecessary computations due to uninformative bands and pixels, often linked to the sensors' spectral and spatial resolutions. Spatial variability, influenced by imaging conditions like atmosphere, lighting, and topography, causes mismatches between object categories and spectral curves. To address these challenges, traditional HSI interpretation strategies typically involve two steps: \textit{dimensionality reduction} \cite{dr_fs} and \textit{feature extraction} \cite{dr_fe}, with feature extraction being a critical research focus due to its impact on subsequent tasks' performance. The HSI community continually seeks effective feature extractors, leading to numerous proposed methods \cite{hsi_mp,hsi_sp,uniadrs}. Among these, deep learning techniques have become predominant for their ability to automatically extract strong feature representations. To tackle spatial variability and enhance contextual feature representation, significant progress has been made with HSI interpretation methods based on convolutional neural networks (CNNs) \cite{3dcnn}, recurrent neural networks (RNNs) \cite{ssun}, transformers \cite{spectralformer}, and state space models \cite{hsidmamba}\footnote{The development history of HSI processing technology can be found in the appendix.}. Despite these advancements, most HSI methods remain scene-dependent, meaning models are trained and tested on the same specialized scenarios with limited cross-scene knowledge transfer, even for similar tasks. For instance, models for classifying Indian Pines and Pavia University scenes are trained separately, restricting their applicability. Therefore, developing universal HSI processing methods is crucial for advancing the hyperspectral community. Fig. \ref{fig: different_patterns} illustrates the differences between traditional and universal HSI processing schemes.

Foundation models, developed through training on vast datasets and optimized using substantial computational resources, have recently been transforming the field of artificial intelligence \cite{fm_agi,fm_definition}. Most foundation models utilize the transformer architecture \cite{selfattention}, known for its scalability and flexibility, allowing for rapid parameter scaling through the stacking of transformer blocks. In the field of computer vision (CV), numerous large-scale foundation models \cite{xu2021vitae, vitae_v2, vit_g} based on vision transformers ~\cite{swint, vit} have demonstrated exceptional performance across various tasks. Similarly, current remote sensing (RS) foundation models \cite{ringmo, rvsa} have proven capable of effectively handling multiple tasks, including scene classification, semantic segmentation, object detection, and change detection.

\begin{figure}[t]
\centering
\includegraphics[width=0.7\linewidth]{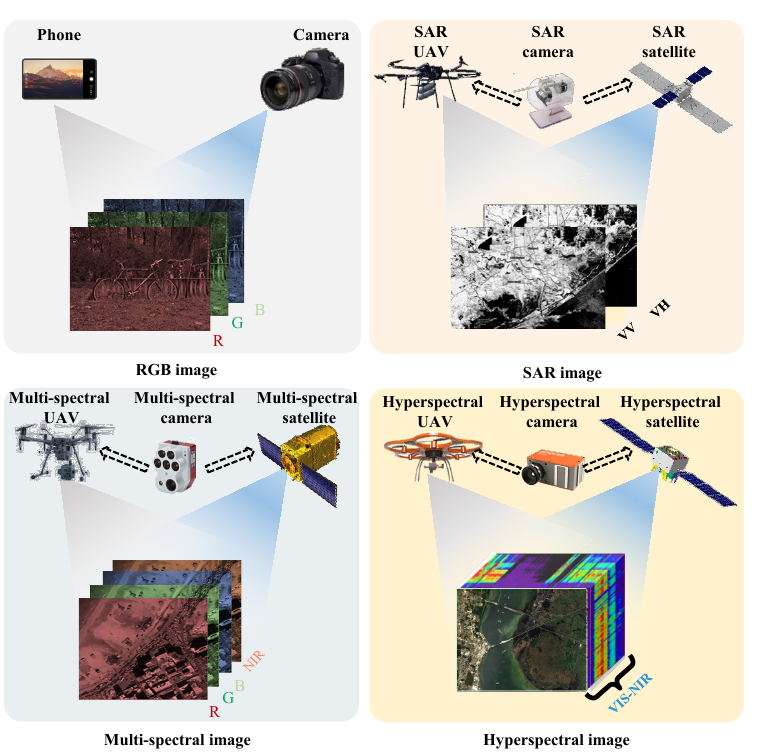}
\caption{Comparison of RGB, Synthetic-aperture radar (SAR), multi-spectral, and hyperspectral images.}
\label{fig: intro_HSI}
\end{figure}

Intuitively, developing hyperspectral foundation models may provide a promising solution for unified HSI interpretation across tasks and scenes. However, our literature review reveals a lack of foundation models specifically designed for HSI interpretation, despite their ability to capture terrestrial objects at a finer wavelength scale than aerial RGB, SAR, and multispectral images (see Fig. \ref{fig: intro_HSI}). The challenges to developing large-scale foundation models for HSI processing primarily lie in the unique characteristics of hyperspectral data, which complicate data collection, pre-training, and model design. Specifically, large-scale pre-training requires significant computational resources, while acquiring and processing hyperspectral data is labor- and time-intensive. Additionally, effective pre-training strategies and model architectures are expected to be tailored to the distinct characteristics of HSIs. These combined obstacles hinder the progress of large-scale hyperspectral foundation models in the RS community.

To address these challenges, we introduce \textbf{HyperSIGMA} (\textbf{HyperS}pectral \textbf{I}ntelli\textbf{G}ence co\textbf{M}prehension found\textbf{A}tion model), the first step towards hyperspectral foundation models tailored for HSI interpretation. HyperSIGMA integrates spatial and spectral features using a specially designed spectral enhancement module. To tackle the issues of spectral and spatial redundancy in HSIs, we introduce a novel sparse sampling attention (SSA) mechanism, which effectively promotes the learning of diverse contextual features and serves as the foundational block of HyperSIGMA. In addition, we have constructed a large-scale hyperspectral dataset, HyperGlobal-450K (\textbf{Hyper}spectral \textbf{Global} Image dataset), for pre-training. This dataset comprises about 450K hyperspectral images, equivalent to over 20 million trispectral images with non-overlapping channels. The appendix presents detailed comparisons between HyperGlobal-450K and existing large-scale RS datasets. Drawing inspiration from the scaling law \cite{scaling_law} and the success of existing large-scale RS foundation models \cite{bfm, spectralgpt, skysense}, we have scaled HyperSIGMA to over 1 billion parameters, supported by HyperGlobal-450K. Unlike existing RS foundation models focusing primarily on high-level tasks such as semantic segmentation and object detection, with minimal exploration of low-level tasks like image denoising and super-resolution, HyperSIGMA offers a unified solution to both high-level and low-level tasks (see Fig. \ref{fig: radar}). We hope this study provides valuable insights into developing foundation models for HSIs and anticipate that HyperSIGMA's strong representation capability will advance the hyperspectral RS field across diverse applications.

The main contributions of this paper are four-fold:
\begin{itemize}
  \item We construct a global hyperspectral image dataset, HyperGlobal-450K, facilitating large-scale pre-training of hyperspectral foundation models. HyperGlobal-450K surpasses existing multispectral and hyperspectral datasets in volume by orders.
  \item We develop a hyperspectral intelligence comprehension foundation model, HyperSIGMA, with over 1 billion parameters. It is the first billion-level foundation model specifically designed for HSI interpretation, offering a unified solution to both high-level and low-level tasks.
  \item We propose a novel attention mechanism, sparse sampling attention, addressing challenges inherent to hyperspectral images by effectively extracting strong feature representations with diverse contexts.
  \item Extensive experiments across diverse HSI tasks provide valuable insights into HyperSIGMA's remarkable versatility and superior representational capability compared to current state-of-the-art methods. Moreover, HyperSIGMA demonstrates significant advantages in scalability, robustness, cross-modal transferring capability, real-world applicability and computational efficiency.
  
\end{itemize}

\begin{figure*}[h]
  \centering
  \includegraphics[width=0.8\linewidth]{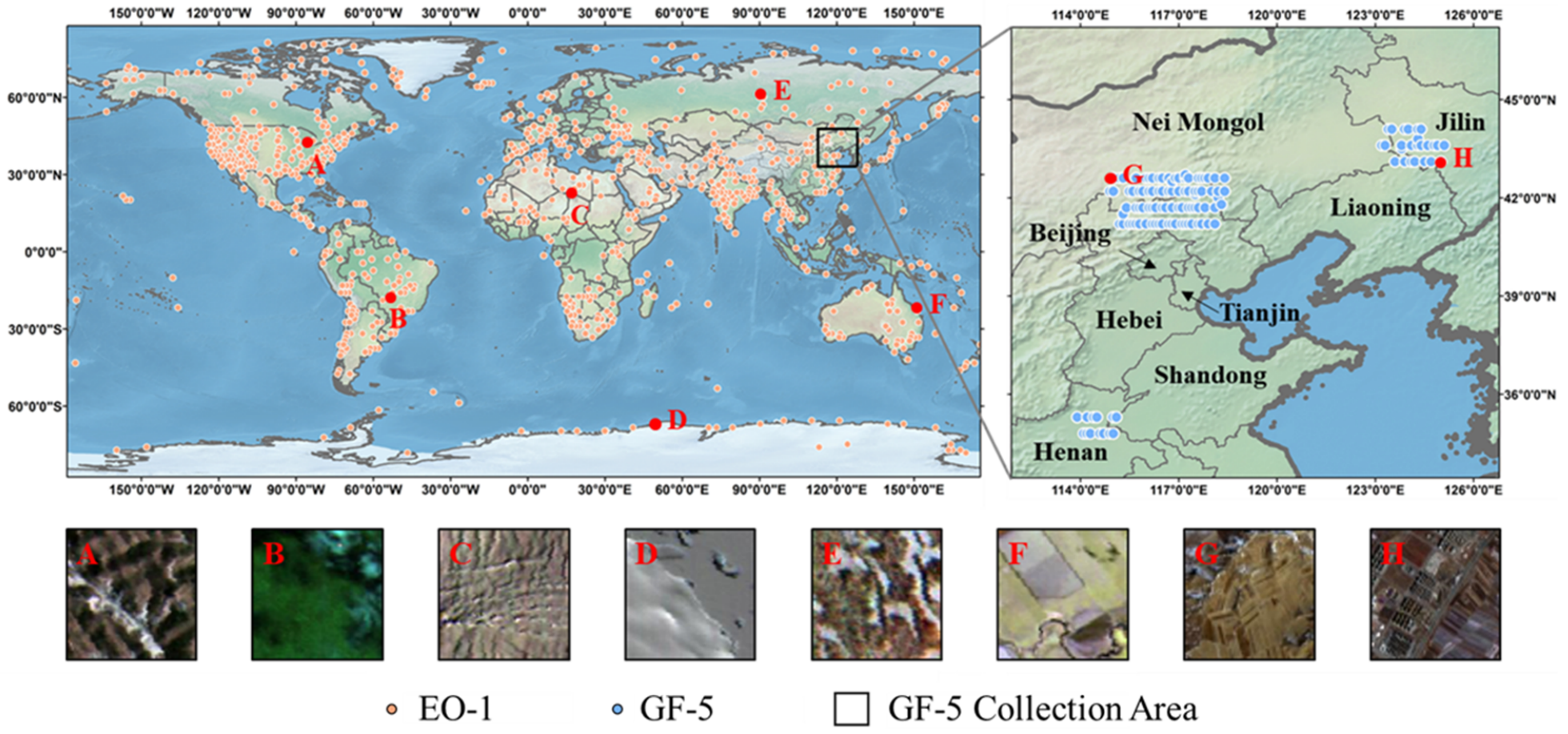}
  \caption{
  The distribution of HyperGlobal-450K samples across the globe. The sampled patches of typical landscapes from different regions, including forests, grasslands, barelands, and croplands, clearly exhibit the characteristics of their respective geographical regions.
  }
\label{fig: Pretrainingdataset}
\end{figure*}

\section{Related Work}
\label{sec:related_work}

Numerous foundation models have been proposed for natural images, with the vision transformer adopted as the mainstream network structure. Among them, scaling is an effective strategy. Typically, ViT \cite{vit} and Swin Transformer \cite{swint} have been scaled up to 1 or 10 billion parameters \cite{vit_g, swin_v2} to explore the potential of large-scale vision models. Another available approach is combining the advantages of both convolution and attention. For this purpose, ViTAE \cite{xu2021vitae} introduces the inductive bias of CNN into vision transformers. In addition to vision transformers, CNN-based large-scale foundation models have also been explored \cite{internimage}.

Compared to the large model size in the CV field, the foundation models in the RS field are usually smaller, focusing more on the pre-training strategy on RS images \cite{rsp}. Due to the high cost of annotating RS datasets and the abundance of unlabeled images, pre-training via self-supervised learning is more prevalent. Among them, contrastive learning \cite{mocov2} leverages RS-specific characteristics such as seasonal \cite{seco}, temporal \cite{caco}, and geographic variations \cite{geoclip}. Besides, many models \cite{ringmo, rvsa, satmae, smlfr, bfm, gfm, spectralgpt} employ masked image modeling (MIM) \cite{simmim, mae}. For instance, SatMAE \cite{satmae} extends MAE \cite{mae} to multispectral and multitemporal data. SpectralGPT \cite{spectralgpt} partitions multispectral images into 3-D cubes to create visual embeddings for MAE. While SMLFR \cite{smlfr} develops pioneering generative CNN foundation models for RS scenes by pre-training on approximately 9 million images, achieving superior performance on multiple RS downstream tasks. In addition to pre-training from scratch, some RS foundation models are constructed based on existing vision models. GFM \cite{gfm} utilizes a powerful universal foundation model to guide the learning of RS features, while MTP \cite{mtp} performs continuous multi-task supervised pre-training on existing RS or CV foundation models. In contrast, UPetu \cite{upetu} introduces the first unified RS parameter-efficient fine-tuning framework, achieving performance comparable to full fine-tuning while optimizing just 0.73\% of the parameters. 

Compared to existing approaches that primarily focus on RGB and multispectral images, HyperSIGMA is the first billion-level foundation model specifically for HSI interpretation. Leveraging pre-training on the large-scale HyperGlobal-450K dataset, HyperSIGMA employs a sparse sampling mechanism to tackle spectral and spatial redundancy in HSIs, offering a unified approach to both high-level and low-level hyperspectral vision tasks. Further detailed comparison between HyperSIGMA and existing representative CV and RS foundation models can be found in the appendix, where more discussions about the related works of HSI processing techniques, MHSA-based vision transformer networks, and large-scale RS datasets are included.

\section{The HyperGlobal-450K Dataset}

\subsection{Data Source}
In view of a series of significant criteria such as global coverage and free access, we select both the Earth Observing One (EO-1) and Gaofen5 (GF-5) satellites as the data source of HyperGlobal-450K. The details of the motivation and workflow in selecting sensors and images, and the basic information of these satellites are presented in the appendix.

\subsection{Data Acquisition and Processing}

We collect all EO-1 images acquired during 2011-2017 and additionally compensate the GF-5 images from China. Then, we design a set of standards to achieve image selection from cloud contents, locations, and bands. After clipping, we obtain a large-scale hyperspectral dataset with global coverage  (see Fig. \ref{fig: Pretrainingdataset}), containing 447,072 HSI patches in size of 64$\times$64, i.e., HyperGlobal-450K, including 247,072 EO-1 patches and 200,000 GF-5 patches. Detailed data acquisition and processing pipelines can be found in the appendix.

\section{Methodology}

The construction of HyperSIGMA involves three main steps: initializing model weights through pre-training, enhancing model structure with SSA, and fusing spatial-spectral features. Following typical HSI processing strategies \cite{ssun}, we use two parallel subnetworks to extract spatial and spectral features. First, inspired by \cite{vitdet, rvsa}, we obtain the model weights using MAE~\cite{mae} pre-training on HyperGlobal-450K. Importantly, the spatial and spectral networks are pre-trained separately. Next, SSA is integrated into the model to enhance its structure. Lastly, we fuse the spatial and spectral information to enhance the representation of the extracted features, resulting in the final HyperSIGMA model. Technical details will be presented in the following sections.

\subsection{Model Pre-training}
\label{subsec:pretraining}
\subsubsection{Masked Image Modeling}

MAE~\cite{mae} is a widely used MIM-based self-supervised learning technique. In MAE, an image is divided into non-overlapping patches, some of which are masked. The network then predicts and reconstructs these masked patches using the visible ones. The loss function is determined by comparing the network's predictions to the ground truth of the masked areas. MAE is particularly effective for pre-training ViTs~\cite{vit} on large-scale unlabeled datasets. In this study, we employ MAE to pre-train both spatial and spectral subnetworks on HyperGlobal-450K.

\subsubsection{Pre-training of Spatial Subnetwork\label{section_4.1.2}}

Similar to many multispectral foundation models \cite{satmae, spectralgpt}, we use ViT~\cite{vit} as the backbone for the spatial subnetwork. Following successful practices \cite{vitdet, rvsa, satmae, spectralgpt}, we employ MAE~\cite{mae} for pre-training. The only modification from the original implementation on natural images is adjusting the input channel of the patch embedding layers to match the number of channels in the input HSIs.

\subsubsection{Pre-training of Spectral SubNetwork\label{section_4.1.3}}
 
For the spectral subnetwork, similar to \cite{ss-mae}, we extend the concept of spatial tokenization in ViTs to the spectral domain, generating spectral tokens by embedding channels. To implement spectral MAE, we adapt the channel tokenization process. Specifically, for a 3-D HSI cube \( \mathbf{X}_0 \in \mathbb{R}^{H \times W \times C} \), we first aggregate adjacent channels through average clustering along the channel dimension, resulting in \( \mathbf{X}' \in \mathbb{R}^{H \times W \times N_{spec}} \), where \( N_{spec} \) is the desired token number. Next, \( \mathbf{X}' \) is reshaped into a 2-D matrix of shape \( \mathbb{R}^{N_{spec} \times (H \cdot W)} \) by dimensional permutation and flattening. It is then projected into \( D \)-dimension embedding space via linear mapping, resulting in \( \mathbf{X} \in \mathbb{R}^{N_{spec} \times D} \). \( \mathbf{X} \) serves as the spectral channel embedding, analogous to the spatial patch embedding in standard ViT. After spectral tokenization, \( \mathbf{X} \) is processed by the subsequent ViT blocks, and the remaining pre-training steps follow those of the MAE. Here, it should be noted that, although we use the HSIs with DN or radiance value for pre-training, the spectral contexts of real-world data that adopts reflectance value still can be captured, because our spectral subnetwork perceives the relationships between channels regardless of the data type.

\subsubsection{Implementation\label{section_4.1.4}}

\noindent\textbf{Data Pre-processing} Given the high-dimensional nature of HSIs, we perform dimensionality reduction by randomly selecting continuous channels to preserve the original spectral order and enhance the diversity of pre-training data. Since HSIs in the HyperGlobal-450K dataset are acquired by different sensors, our network cannot simultaneously process the images with varying numbers of channels in the same batch. Therefore, the number of selected channels remains consistent across all HSIs. The specific details can be found in the appendix.

\noindent\textbf{Experimental Settings} The mask ratio is a crucial hyperparameter in MIM algorithms, as it influences the difficulty of recovering masked regions and, consequently, the effectiveness of pre-training. A high mask ratio makes the reconstruction task overly challenging, potentially hindering restoration, while a low mask ratio may result in ineffective model weights due to the ease of the task. Thus, selecting an appropriate mask ratio \(R\) is essential for ensuring pre-training quality. For spatial MAE, we adopted \(R_{spat} = 0.75\), a setting proven effective for both natural images \cite{mae} and remote sensing (RS) images \cite{rvsa}. In contrast, research on HSI channel masking and recovery is limited. For example, \cite{ss-mae} independently determined relevant values for different datasets in HSI and LiDAR joint classification. Therefore, we conducted a series of ablation studies to identify a suitable \(R_{spec}\) for spectral MAE. We provide the details of determining $R_{spec}$ (0.75 by default) in the appendix.

After determining the mask ratios, we retrained the spatial and spectral subnetworks using the default settings as previously described. Following \cite{mae, rvsa}, we extended the pre-training to 1,600 epochs for sufficient training. Given the image size of \(\mathbf{X}_0\), the patch size \(P\) for the spatial ViT is set to 8, resulting in \(N_{spat} = \frac{H \cdot W}{P^2} = 64\) tokens. The other training parameters of the spatial subnetwork, including batch size, learning rate, and weight decay, match those of the spectral subnetwork. We pre-trained different ViT versions - base, large, and huge - for both spatial and spectral networks, denoted as SpatViT and SpecViT, respectively. This allows HyperSIGMA's model size to reach the billion level when using the ViT-Huge backbone for both subnetworks. All experiments were conducted on NVIDIA V100 GPUs. The appendix provides more details about the pre-training cost.

\subsection{Model Structure}

\subsubsection{Preliminaries of Vision Transformer}

In this section, we briefly introduce ViTs~\cite{vit} as they form the foundational structures of HyperSIGMA. Inspired by the transformer network used in NLP, where words are mapped into vectors, ViT operates by first splitting an image into non-overlapping patches. Each patch is then mapped to a 1-D token vector. The details of mapping spatial patches and spectral channels of HSI are detailed in Sec.~\ref{section_4.1.2}, Sec.~\ref{section_4.1.3}, and Sec.~\ref{section_4.1.4}. These tokens, combined with learnable positional embeddings $\mathbf{E}$, are processed by a series of transformer blocks $f=\left\{f_1, f_2, \cdots, f_d\right\}$, where $d$ is the network depth. This process can be formulated as:
\begin{equation}
  \begin{aligned}
    \mathbf{U}_0 &= \mathbf{X} + \mathbf{E},  \quad \mathbf{X}\in\mathbb{R}^{N \times D}, \mathbf{E}\in\mathbb{R}^{N \times D}, \\
    \mathbf{U}_{i} &= f_i(\mathbf{U}_{i-1}), \quad  \mathbf{U}_{i}\in\mathbb{R}^{N \times D}, i=1,\cdots,d, \\
    \mathbf{Z} &= \text{LN}(\mathbf{U}_{n}), \quad  \mathbf{Z} \in\mathbb{R}^{N \times D}. \\
  \end{aligned}
\end{equation}
Here, $\mathbf{U}_{i}$ is the output of the $i$th block $f_i$, $\mathbf{Z}$ is the final output feature of the ViT, while LN represents the layer normalization \cite{layernorm}.  The computation process of $f_i$ is formulated as:
\begin{equation}
  \begin{split}
    \mathbf{U}_{i-1}'& = \text{MHSA}(LN(\mathbf{U}_{i-1})) + \mathbf{U}_{i-1},\\
    \mathbf{U}_{i} & = \text{FFN}(\text{LN}(\mathbf{U}_{i-1}')) + \mathbf{U}_{i-1}'.\\
  \end{split}
\end{equation}
Here, FFN denotes the feed-forward networks containing two linear layers. MHSA is the core module of transformer blocks, containing multiple parallel self-attentions (SAs). Each SA can be formulated as:
\begin{equation}
   \text{SA}(\mathbf{U})=softmax(\frac{\mathbf{Q} \mathbf{K}^T}{\sqrt{D'}}) \mathbf{V},
\end{equation}
where $\mathbf{Q}\in\mathbb{R}^{N \times D'},\mathbf{K}\in\mathbb{R}^{N \times D'},\mathbf{V}\in\mathbb{R}^{N \times D'}$ are the query, key and value generated from $\mathbf{U} \in \mathbb{R}^{N \times D}$ by three linear layers, respectively. The output feature of MHSA is obtained by concatenating all the outputs of SAs along the channel dimension:
\begin{equation}
  \text{MHSA}(\mathbf{U})=\left[Concat(\text{SA}_1(\mathbf{U}), \cdots, \text{SA}_h(\mathbf{U}))\mathbf{W}\right]^T.
\end{equation}
Here, $\text{SA}_h(\cdot)$ denotes the $h$-th SA head. $\mathbf{W} \in \mathbb{R}^{hD' \times D}$ denotes the weight matrix of a linear layer to recover the original embedding dimension $D$. In practice, $D=h \cdot D'$ for convenience.

\subsubsection{Sparse Sampling Attention}
\label{subsubsec:ssa}

In this section, we introduce the proposed SSA, designed to efficiently learn diverse contextual features by addressing the spatial and spectral redundancy of HSIs. Given $\mathbf{Q}=\{q_1, \cdots, q_N\}$, $\mathbf{K}=\{k_1, \cdots, k_N\}$, and $\mathbf{V}=\{v_1, \cdots, v_N\}$, we predict $N_p$ offsets $[(\Delta x_1, \Delta y_1), \cdots, (\Delta x_{N_p}, \Delta y_{N_p})]$ using a linear layer $\bm{W}_p\in \mathbb{R}^{D' \times 2N_p}$ for each query vector $q$ at coordinates $(c_x, c_y)$. Next, we sample new key $k'$ and value $v'$ from the original key and value matrices $\mathbf{K}$ and $\mathbf{V}$ at these positions using bilinear interpolation. Note that $\mathbf{K}$ and $\mathbf{V}$ have been reshaped into 2-D feature maps of shape $\mathbb{R}^{H' \times W' \times D'}$, where $N=H'W'$. This process can be formulated as:
\begin{equation}
  \begin{aligned}
    k'_j=  \sum_{(o_x,o_y)} & \max(0,1-|o_x-(c_x+\Delta x_j)|) \\
    & \max(0, 1-|o_y-(c_y+\Delta y_j)|) \mathbf{K}[o_x,o_y,:], \\
    v'_j=  \sum_{(o_x,o_y)} & \max(0,1-|o_x-(c_x+\Delta x_j)|) \\
    & \max(0, 1-|o_y-(c_y+\Delta y_j)|) \mathbf{V}[o_x,o_y,:].
  \end{aligned}
\end{equation}
Here, $j=1,...,N_p$, $\mathbf{K}[o_x,o_y,:]$ is a vector extracted at $(o_x,o_y)$ from $\mathbf{K}$, where $(o_x,o_y)$ represents all coordinates. We totally sample $N \cdot N_p$ points and obtain $\mathbf{K}', \mathbf{V}'\in \mathbb{R}^{N \times N_p\times D'}$. Consequently, SSA can be formulated as:
\begin{equation}
  \text{SSA}(\mathbf{U}) = softmax(\frac{\mathbf{Q} \otimes \mathbf{K}'}{\sqrt{D'}}) \otimes \mathbf{V}'.
\end{equation}
Here, $\otimes$ contains a series of operations including matrix broadcasting, Hadamard product, and removing dimensions through accumulation.

\begin{figure}[t]
  \centering
  \includegraphics[width=0.8\linewidth]{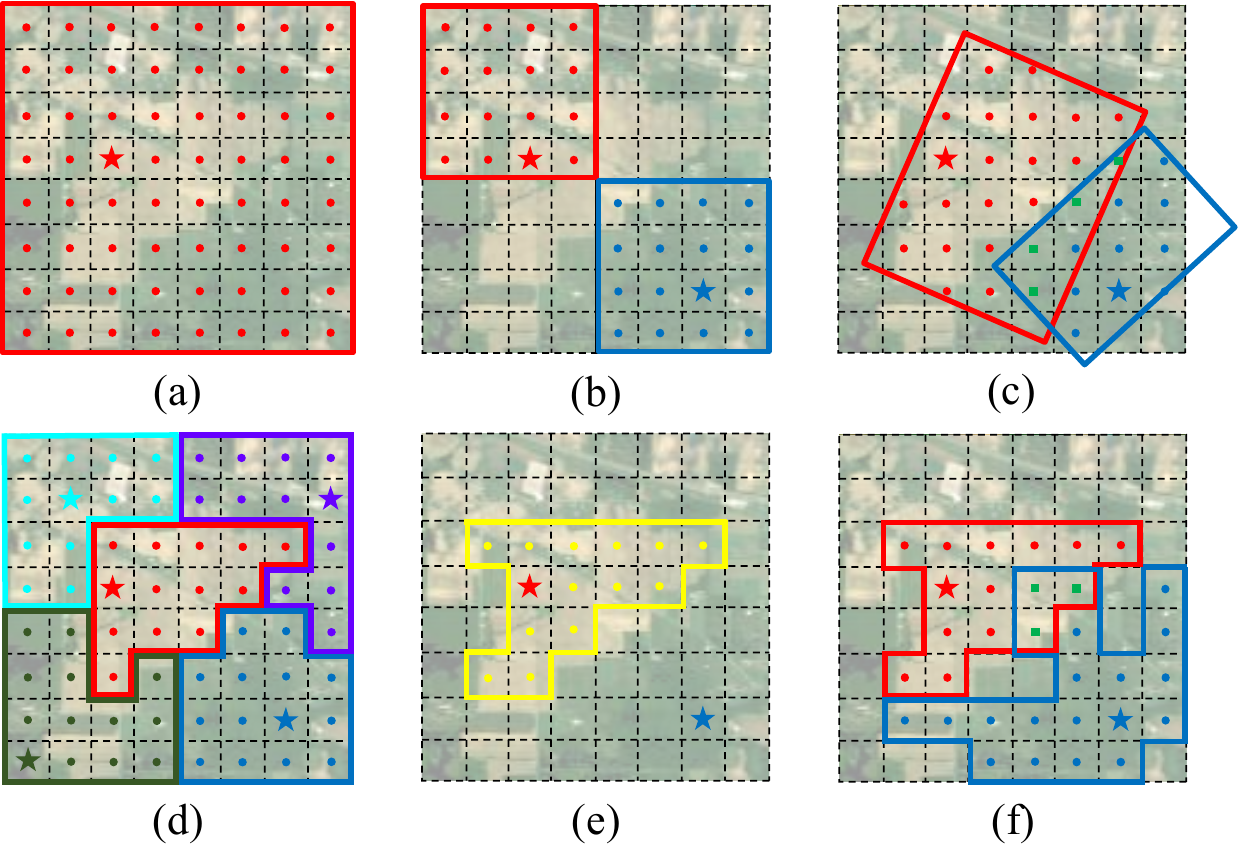}
  \caption{
  Comparison of different attention mechanisms: (a) Full SA~\cite{vit}, (b) WMHSA~\cite{swint}, (c) RVSA~\cite{rvsa}, (d) NLSA~\cite{nlsa}, (e) DMHA~\cite{dat}, (f) SSA. Stars represent queries, with dots surrounded by corresponding colored lines indicating the attention regions of captured contexts. Green rectangles in (c) and (f) denote common areas shared by both queries. In DMHA, all queries share the same keys in the yellow region.
  }
\label{attn_compre}
\end{figure}

\begin{figure*}[t]
  \centering
  \includegraphics[width=0.75\linewidth]{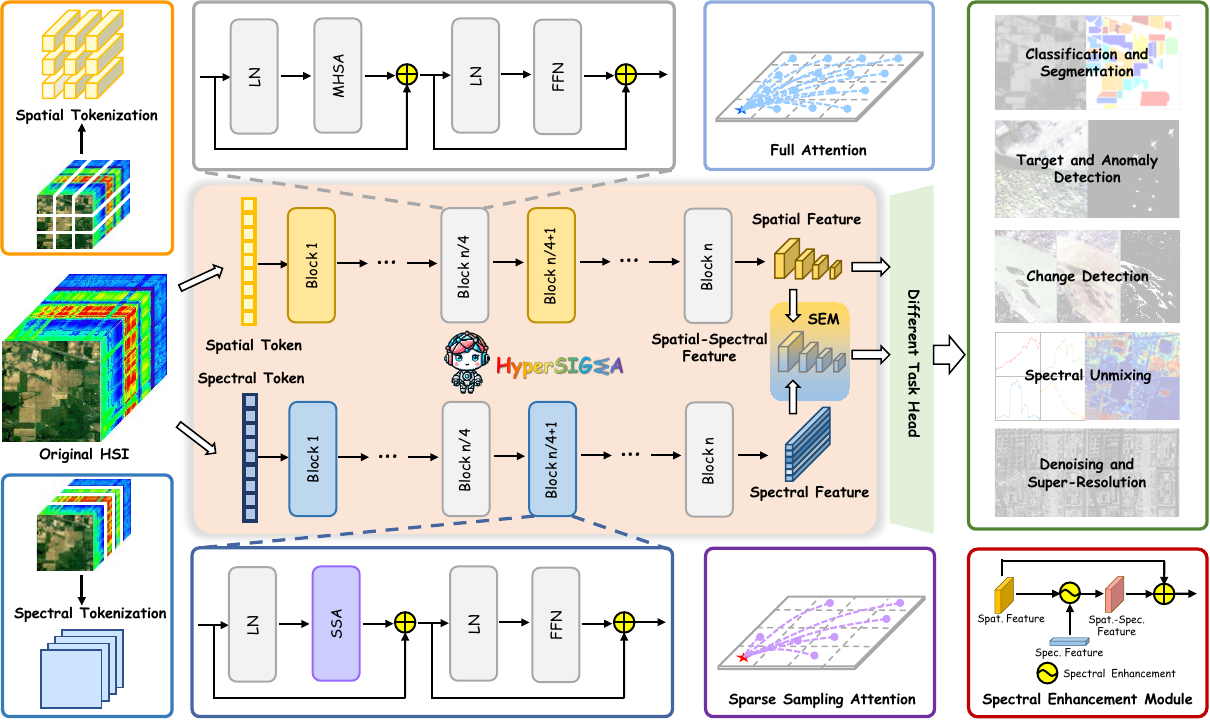}
  \caption{
  The HyperSIGMA model consists of two subnetworks tailored for spatial and spectral feature extraction (Sec.~\ref{subsec:pretraining}). Initially, spatial patches or spectral channels are tokenized and processed through multiple transformer blocks, where some SAs are substituted with the proposed Sparse Sampling Attention (SSA) (Sec.~\ref{subsubsec:ssa}). Spatial and spectral features are then generated and fused by the Spatial-Spectral Fusion Module (SEM) (Sec.~\ref{subsubsec:hypersigma}). These fusion features (or spatial features) are fed into task-specific heads for various high-level and low-level HSI tasks.
  }
\label{fig: whole_framework}
\end{figure*}

\noindent\textbf{Comparison to Other Attention Methods} Fig. \ref{attn_compre} illustrates the differences between the proposed SSA and various attention methods. Full SA (Fig.~\ref{attn_compre} (a)) captures global context with quadratic complexity relative to token sequence length, whereas window-based attention methods \cite{swint, rvsa} (Fig.~\ref{attn_compre} (b)-(c)) enhance efficiency by focusing on local rectangular regions. NLSA\cite{nlsa} partitions the feature map into non-overlapped attention buckets by spherical locality-sensitive hashing, i.e., each token is assigned to a specific group, where the number of buckets is pre-defined. For efficient parallel computing, it requires the buckets to be further divided into fixed-size chunks.  Deformable-DETR \cite{deformable_detr} introduces deformable attention across features, enhancing context diversity by using adaptive sampling points for each query, primarily for object detection. Inspired by this, DMHA \cite{dat} (Fig.~\ref{attn_compre} (e)) integrates deformable attention into vision transformer backbones, but all queries capture context from the same regions, even when different queries might need different contexts. In contrast, our SSA (Fig.~\ref{attn_compre} (f)) learns a unique set of sampling positions for each query, further enriching the context. Additionally, unlike deformable attention which uses linear layers to predict attention weights, SSA calculates attention weights through token interaction, enhancing the ability to capture inter-region relationships.

\subsubsection{Structure of HyperSIGMA}
\label{subsubsec:hypersigma}

After obtaining the pre-trained weights of SpatViT and SpecViT (Sec.~\ref{subsec:pretraining}), we construct the HyperSIGMA model by replacing the original full SA with the proposed SSA (Sec.~\ref{subsubsec:ssa}) and fusing spatial-spectral features. The overall structure of HyperSIGMA is shown in Fig.~\ref{fig: whole_framework}.

\noindent\textbf{Attention Replacement} Following the approach in \cite{vitdet, rvsa}, we retain only the full SA in the $i \cdot (n/4)$th ($i=1,2,3,4$) layers, while replacing the full SA in other layers with the proposed SSA. This approach allows the network to effectively extract both local and global contexts.

\noindent\textbf{Feature Fusion} In traditional HSI dual-branch processing pipelines \cite{ssun}, it is common practice to combine spatial and spectral features in the late stage. Following this practice, we fuse the features extracted from spatial and spectral subnetworks. It is noteworthy that, in the spectral ViT, each channel is transformed into a 1-D token vector, disrupting the original spatial structures. Our preliminary experiments have shown that attempting to reproject these tokens into 2-D features is ineffective. Therefore, we retain the 1-D shape features for the spectral subnetwork. In our implementation, we fuse the spatial and spectral features using a specially designed approach, the Spectral Enhancement Module (SEM), which enhances spatial features with spectral information.

Specifically, given a spatial feature $\mathbf{F}_{spat} \in \mathbb{R}^{H' \times W' \times D}$ and a spectral feature $\mathbf{F}_{spec} \in \mathbb{R}^{N_{spec} \times D}$, we first compress their dimensions using a linear layer to obtain $\mathbf{F}_{spat} \in \mathbb{R}^{H' \times W' \times D_1}$ and $\mathbf{F}_{spec} \in \mathbb{R}^{N_{spec} \times D_1}$. Next, we reduce the spatial dimension of $\mathbf{F}_{spec}$ through average aggregation to create a 1D vector $\mathbf{V} \in \mathbb{R}^{N_{spec}}$. Finally, we apply another linear layer to align the dimension of $\mathbf{V}$ with the channel number of $\mathbf{F}_{spat}$. The spectral enhancement process is formulated as follows:
\begin{equation}
  \mathbf{F}^{\star} = (1+\mathbf{V})*\mathbf{F}_{spat}, 
\end{equation}
Here, we reuse the symbol $\mathbf{V}$. $\ast$ denotes channel-wise product. Note that a skip-connection is employed to retain the original spatial information. In this way, the extracted spatial features are calibrated using spectral information.

\noindent\textbf{SpatSIGMA} Since both $\mathbf{F}_{spat}$ and $\mathbf{F}^{\star}$ offer potential for interpreting HSIs, alongside HyperSIGMA, the spatial subnetwork termed SpatSIGMA also forms a foundational model for solving various high-level and low-level HSI tasks.

\noindent\textbf{Remarks} The proposed method introduces three key novelties:  
(1) \textit{Unsupervised Pre-training}: By eliminating the reliance on labeled data, unsupervised pre-training significantly reduces annotation costs. A random channel selection strategy is employed during preprocessing, effectively mitigating the high dimensionality of HSI and enabling adaptation to data from mixed sensors.  
(2) \textit{Spatial-Spectral Fusion}: The SEM is designed and integrate spectral information into spatial representations through channel-wise corrections, offering a robust spatial-spectral feature representation.  
(3) \textit{Adaptive Attention}: The SSA mechanism adaptively determines optimal positions for sampling keys and values, addressing spatial and spectral redundancy in HSIs. By replacing the full attention mechanism in both SpatViT and SpecViT, SSA improves contextual diversity, enhances representational capability, and lowers computational complexity.

\section{Experiments}
In this section, we thoroughly evaluate HyperSIGMA's performance on representative HSI high-level interpretation tasks: image classification (Sec.~\ref{subsec:exp_classification}), target detection (Sec.~\ref{subsec:exp_detection}), anomaly detection (Sec.~\ref{subsec:exp_detection}), and change detection (Sec.~\ref{subsec:exp_change}), and low-level tasks: spectral unmixing (Sec.~\ref{subsec:exp_unmixing}), image denoising (Sec.~\ref{subsec:exp_denoising}) and image super-resolution (Sec.~\ref{subsec:exp_sr}). Additionally, we conduct experiments to examine HyperSIGMA's scalability (Sec.~\ref{subsubsec:exp_more_scalability}), robustness (Sec.~\ref{subsubsec:exp_more_robustness}), cross-modal transferring capability (Sec.~\ref{subsubsec:exp_more_crossmodal}), real-world applicability (Sec.~\ref{subsubsec:exp_more_applications}) and computational efficiency (Sec.~\ref{subsubsec:exp_complexity}).

\noindent\textbf{HyperSIGMA in Fine-tuning} Considering the variability in HSIs, where the number of channels differs across datasets, pre-trained weights for the spatial patch embedding layer may not be usable during fine-tuning if the channel numbers do not match. In such cases, we randomly initialize the embedding layer and retrain it during fine-tuning. Notably, the number of channels remains consistent during fine-tuning and inference, utilizing all channels to fully leverage the spectral information. Similarly, pre-trained weights for the spectral tokenization layer cannot be used if the spatial size of inputs for the spectral subnetwork is not consistent during pre-training and fine-tuning. Besides, pre-trained positional embeddings in spatial (spectral) subnetwork will be adjusted through interpolation to fit the spatial size (spectral channel) of the fine-tuning data.

\subsection{Hyperspectral Image Classification}
\label{subsec:exp_classification}
Image classification is a fundamental task in HSI interpretation. In this section, we begin with a series of ablation studies (Sec.~\ref{subsubsec:exp_classification_ablation}), followed by fine-tuning pre-trained models on common HSI classification datasets (Sec.~\ref{subsubsec:exp_classification_mainresults}).

\subsubsection{Experiment Settings}
\noindent\textbf{Dataset} In addition to the widely used Indian Pines (IP) \cite{indian_pines} and Pavia University (PU)\footnote{\url{https://www.ehu.eus/ccwintco/index.php/Hyperspectral\_Remote\_Sensing\_Scenes}} datasets, we also utilize three challenging datasets: HanChuan (HC) \cite{whu_hi1}, HongHu (HH) \cite{whu_hi1}, and Houston (HU) \cite{houston}. Furthermore, we introduce a new dataset from the Yellow River Delta, captured by the ZY1-02D satellite (ZY) \cite{huanghe}.

\noindent\textbf{Implementation} The classification task focuses on classifying all pixels in a hyperspectral image, similar to the semantic segmentation tasks in natural or aerial RGB images. This can be achieved through either patch-level or image-level classification \cite{fullycontnet}. Patch-level classification selects a patch centered around each pixel and classifies the center pixel based on this patch, while image-level classification corresponds directly to semantic segmentation. For computational efficiency, we primarily employ semantic segmentation unless otherwise specified. 

Classification performance is assessed using the common overall accuracy (OA). We compare HyperSIGMA against several classical and advanced classification and segmentation methods, including MSDN\cite{MSDN}, SSFCN\cite{ssfcn}, FullyContNet\cite{fullycontnet}, SpectralFormer\cite{spectralformer}, HSIC-FM\cite{HSIC-FM}, SSGRN \cite{ssgrn}, CSIL\cite{yang_csil}, IDCN\cite{idcn}, and CLOLN\cite{CLOLN}. The details related to network structure and experimental configurations are presented in the appendix.

\subsubsection{Ablation Studies}
\label{subsubsec:exp_classification_ablation}

To determine effective hyperparameter settings, we conduct a series of ablation experiments on the IP and PU datasets. For simplicity, we use only the spatial subnetwork. According to the ablation study (see the appendix), the number of sampling points is set to 8. Additionally, we prove the effectiveness of the proposed SSA and the necessity for pre-training on the constructed large-scale hyperspectral dataset, i.e., HyperGlobal-450K.

\begin{table}[t]
\scriptsize
    \centering
    \caption{OA (\%) of various methods across HSI classification datasets. \textbf{\color{red}{Best}} and \textbf{\color{blue}{2nd-best}} results are highlighted.}
 \resizebox{\linewidth}{!}{
    \begin{tabular}{lcccccccccccc}
    \hline
        Method & IP & PU & HC & HH & HU & ZY \\ 
        \hline
        MSDN \cite{MSDN} & 57.54  & 76.48  & 73.40  & 78.55 & 72.18  &  88.57 \\
        SSFCN \cite{ssfcn} & 41.93  & 78.88  & 63.35  & 71.62 & 72.39 & 82.46 \\ 
        FullyContNet \cite{fullycontnet} & 71.11  & 80.31  & 78.80  & 67.12 & 51.07 & 84.47 \\ 
        SpectralFormer\cite{spectralformer} & 50.02  & 75.37  & 82.60  & 85.33 & 77.21 &  72.03 \\ 
        HSIC-FM\cite{HSIC-FM} & 36.02  & 77.28  & 66.21  & 70.47  &  54.43 & 76.98 \\
        SSGRN\cite{ssgrn} & 69.58  & 81.45  & 90.43  & 82.19 & 68.62  & 77.46 \\
        CSIL\cite{yang_csil} & 66.53  & 88.23  & 88.55  & 91.86 & 66.11  & 92.49 \\ 
        IDCN \cite{idcn} & 71.12 & 91.64 & 84.15 & 89.19 & 85.34 & 92.29 \\
        CLOLN\cite{CLOLN} & 72.75  & 93.11  & 86.73  & 87.89  & 85.95   & 80.43 \\ 
        \hline
        SpatSIGMA & \textbf{\color{blue}{85.08}}  & \textbf{\color{blue}{93.36}}  & \textbf{\color{blue}{94.03}}  & \textbf{\color{blue}{94.35}} & \textbf{\color{red}{87.33}} & \textbf{\color{blue}{94.72}}  \\ 
        HyperSIGMA & \textbf{\color{red}{85.54}} & \textbf{\color{red}{93.52}} & \textbf{\color{red}{94.44}} & \textbf{\color{red}{94.87}} &\textbf{\color{blue}{86.80}} &\textbf{\color{red}{94.92}} \\ \hline
    \end{tabular}
    }
    \label{tab:hsi_classification}
\end{table}

\subsubsection{Results and Analyses}
\label{subsubsec:exp_classification_mainresults}

Table \ref{tab:hsi_classification} presents the classification accuracies of different methods. Our models consistently outperform state-of-the-art approaches. For example, on the HanChuan dataset, containing 16 visually similar agricultural categories, models were trained with 50 labeled samples per class (about 0.22\% of the entire image, please see the appendix). Despite these constraints, our approach still outperforms the recent graph convolutional network \cite{kipfgcn} based method SSGRN \cite{ssgrn}, which achieves an OA of 90.43\% but lags 4\% behind HyperSIGMA. Apart from the Houston dataset, HyperSIGMA consistently outperforms SpatSIGMA across diverse scenes, leveraging spectral information to improve accuracy. In summary, our models exhibit significant advancements in extracting robust and universal representations for HSI classification, demonstrating notable performance improvements. The appendix provides more qualitative classification results to demonstrate the performance of HyperSIGMA more clearly.

\subsection{Hyperspectral Target and Anomaly Detection}
\label{subsec:exp_detection}

Unlike typical object detection for optical images, which identifies objects in natural or aerial images by predicting bounding boxes, hyperspectral detection tasks utilize unique spectral information from HSIs to identify target regions of interest. These tasks commonly include hyperspectral target detection (HTD) and hyperspectral anomaly detection (HAD). HTD aims to locate areas with spectral characteristics similar to known target spectra, whereas HAD identifies anomalies based on spectral differences from the surrounding environment, without requiring prior target spectra. Both tasks share the goal of detecting specific regions of interest. In this section, to demonstrate HyperSIGMA's transferability, we conduct fine-tuning experiments on both HTD and HAD tasks.

\subsubsection{Experiment Settings}
\noindent\textbf{Dataset} We adopt Mosaic \cite{htd_mosaic}, AVIRIS \cite{htd_aviris}, and Renourban \cite{htd_renourban} datasets for the HTD tasks, and Pavia \cite{had_pavia}, Cri \cite{had_cri}, and Viareggio \cite{had_viareggio} for the HAD tasks.

\noindent\textbf{Implementation} Following HTD-ViT \cite{htd_vit}, we tackle the hyperspectral detection task using coarse detection labels. Unlike HTD-ViT, which employs per-pixel classification, we draw inspiration from promptable segmentation \cite{sam} to design a segmentation network utilizing HyperSIGMA as a feature extractor. By reframing hyperspectral detection as a segmentation problem, we evaluate HyperSIGMA's performance in both HTD and HAD tasks.

We compare our models, SpatSIGMA and HyperSIGMA, with existing classical and advanced HTD methods: CEM \cite{cem}, STD \cite{htd_std}, CSCR \cite{cscr}, SRBBH \cite{srbbh}, HTD-Net \cite{htdnet}, HTD-IRN \cite{htdirn}, and CGSAL \cite{cgsal}, as well as HAD methods including RX \cite{rx}, KIFD \cite{kifd}, CRD \cite{crd}, GTVLRR \cite{gtvlrr}, Auto-AD \cite{autoad}, RGAE \cite{rgae}, and GT-HAD \cite{gthad}. The area under the ROC curve (AUC) is used as the evaluation metric. Please refer to the appendix for detailed network structure and experimental configurations.

\subsubsection{Results and Analyses}

\begin{table}[t]
  \caption{AUC (\%) of various methods for the HTD and HAD tasks. \textbf{\color{red}{Best}} and \textbf{\color{blue}{2nd-best}} results are highlighted.}
  \centering
  \resizebox{\linewidth}{!}{
  \begin{tabular}{lccc|lccc}
\hline
\multicolumn{4}{c|}{Target Detection} & \multicolumn{4}{c}{Anomaly Detection} \\
\hline
 Method & Mosaic & AVIRIS & Renourban & Method & Pavia &  Cri & Viareggio \\
 \hline
 CEM\cite{cem} & 97.92 & \textbf{\color{red}{99.83}} & 95.52 & RX\cite{rx} & 99.82  &  96.75 & 95.25 \\
 STD \cite{htd_std} &99.30  & 93.98 & 82.30 & KIFD\cite{kifd}  & 93.33 & \textbf{\color{blue}{99.28}} & 96.79  \\
 CSCR\cite{cscr} & 98.95 & 99.35 & 87.60 & CRD \cite{crd} & 99.33  &  91.64 &95.87 \\
 SRBBH\cite{srbbh} & 93.33 & 84.43 & 92.89 & GTVLRR \cite{gtvlrr} & 99.49 & 88.25 & 94.54 \\
 HTD-Net\cite{htdnet} &89.70  & 98.80 & 94.43 & Auto-AD \cite{autoad} & 99.79 & 97.86 & 97.28 \\
 HTD-IRN\cite{htdirn} & 99.30 & 98.98 & 97.65 &RGAE\cite{rgae} &\textbf{\color{blue}{99.94}} &  97.09 & 79.68 \\
 CGSAL \cite{cgsal} & 99.44 & 98.98 & 99.34 & GT-HAD \cite{gthad}& 99.87 & 94.39 & 93.05  \\
 \hline
SpatSIGMA &\textbf{\color{red}{99.75}} & \textbf{\color{blue}{99.78}} & \textbf{\color{blue}{99.79}} & SpatSIGMA &\textbf{\color{blue}{99.94}} & 98.14 & \textbf{\color{red}{98.44}} \\
HyperSIGMA & \textbf{\color{blue}{99.56}}  &  \textbf{\color{red}{99.83}} & \textbf{\color{red}{99.86}} & HyperSIGMA & \textbf{\color{red}{99.96}} & \textbf{\color{red}{99.32}} &\textbf{\color{blue}{98.32}} \\
 \hline
\end{tabular}
 }
 \label{htd_had_result}
\end{table}

 The results, shown in Table \ref{htd_had_result}, demonstrate that our models outperform all existing methods in both HTD and HAD tasks. HyperSIGMA, with enhanced spectral information, shows further accuracy improvements over SpatSIGMA in most cases. In summary, our models achieve state-of-the-art performance in hyperspectral detection tasks, demonstrating their effectiveness of transferability. The appendix offers further analyses of the detection results.

\subsection{Hyperspectral Change Detection}
\label{subsec:exp_change}
In this section, we fine-tune pre-trained models for hyperspectral change detection, which parallels conventional remote sensing tasks using aerial RGB images \cite{hanet, c2fnet}. Our focus here is primarily on classical bi-temporal scenarios.

\subsubsection{Experiment Settings}
\noindent\textbf{Dataset}
We tested on four common benchmark datasets: Hermiston, Farmland, Bay Area (BA), Santa Barbara (SB), which are publicly available\footnote{\url{https://citius.usc.es/investigacion/datasets/hyperspectral-change-detection-dataset}}\footnote{\url{https://rslab.ut.ac.ir/data}}.

\noindent\textbf{Implementation} Following the successful practice of SST-Former \cite{hsi_cd_sstformer}, a transformer-based method for hyperspectral change detection, we adopt a classification network. We compared our models against several established methods: CVA \cite{hsi_cd_cva}, ISFA \cite{hsi_cd_isfa}, GETNET \cite{hsi_cd_getnet}, ML-EDAN \cite{hsi_cd_mledan}, BIT \cite{hsi_cd_bit}, EMS-Net \cite{hsi_cd_emsnet}, CSA-Net \cite{hsi_cd_csanet}, SST-Former \cite{hsi_cd_sstformer}, and GlobalMind \cite{hsi_cd_globalmind}. Model performance was evaluated using F1 score. Please refer to the appendix for detailed network structure and hyperparameter settings.

\begin{table}[t]
\scriptsize
    \centering
    \caption{F1 (\%) of various methods on hyperspectral change detection tasks. \textbf{\color{red}{Best}} and \textbf{\color{blue}{2nd-best}} results are highlighted.}
    \begin{tabular}{lcccc}
    \hline
        Method & Hermiston & Farmland & BA & SB \\ \hline
        CVA \cite{hsi_cd_cva} &81.03&92.49&87.09&83.76\\ 
        ISFA \cite{hsi_cd_isfa} &72.62&93.01&89.05&85.35\\
        GETNET \cite{hsi_cd_getnet} &90.54&94.39&96.07&96.59\\
        ML-EDAN \cite{hsi_cd_mledan} &89.03&93.93&97.41&97.80\\
        BIT \cite{hsi_cd_bit} &71.28&84.56&94.69&96.19\\ 
        EMS-Net \cite{hsi_cd_emsnet} &85.67&92.97&97.67&97.32\\
        CSA-Net \cite{hsi_cd_csanet} &85.29&93.27&97.80&98.24\\
        SST-Former\cite{hsi_cd_sstformer} &88.80&94.34&97.23&97.58\\
        GlobalMind\cite{hsi_cd_globalmind}  &90.71&94.98&98.26&98.28\\
        \hline
        SpatSIGMA &\textbf{\color{red}{92.08}} &\textbf{\color{blue}{95.31}} &\textbf{\color{blue}{98.89}} &\textbf{\color{blue}{98.95}}\\
        HyperSIGMA &\textbf{\color{blue}{91.74}} &\textbf{\color{red}{95.49}}&\textbf{\color{red}{98.93}} &\textbf{\color{red}{99.04}}\\ \hline
    \end{tabular}
    \label{table:change_detection}
\end{table}

\subsubsection{Results and Analyses}
As shown in Table \ref{table:change_detection}, our models achieve the highest F1 scores across all datasets, demonstrating their clear superiority over other methods. Compared to SpatSIGMA, HyperSIGMA further improves its performance on three datasets. Overall, our models deliver finer and more complete detection results, demonstrating HyperSIGMA's strong feature representation for hyperspectral change detection tasks. Please refer to the appendix for more results.

\subsection{Hyperspectral Unmixing}
\label{subsec:exp_unmixing}

After applying HyperSIGMA to high-level downstream tasks, we extended its use to low-level tasks, starting with hyperspectral unmixing \cite{bayesian_unmixing}. This task aims to address the complex spectral mixtures in hyperspectral data by decomposing each pixel's spectral signature into pure spectral signatures (endmembers) and their corresponding proportions (abundances) \cite{spectral_mixing, hyperspectral_unmixing, ghamisi2017advances}. This facilitates the identification and quantification of various components in each pixel.

\begin{table}[t]
\scriptsize
\caption{
Quantitative comparison of endmember and abundance prediction performance across various methods on the Urban dataset. \textbf{\color{red}{Best}} and \textbf{\color{blue}{2nd-best}} results are highlighted.
}
  \centering
  \begin{tabular}{lcc}
\hline
 Method & Abundance & Endmember \\
 \hline
FCLS \cite{fcls} & 0.1406 & 0.4373 \\
ELMM \cite{elmm} & 0.0334  & - \\
SUnSAL \cite{sunsal} & 0.1859 & - \\
CNNAEU \cite{cnnaeu} & 0.0216 & 0.0865 \\
CyCU \cite{cycu} & 0.0821 & 0.2952 \\
DeepTrans \cite{deeptrans} & 0.0488 &  0.1512 \\
EGUnet\cite{egunet}  & 0.1732 & 0.5796 \\
\hline
SpatSIGMA & \textbf{\color{blue}{0.0176}} & \textbf{\color{blue}{0.0598}} \\
HyperSIGMA &\textbf{\color{red}{0.0157}}  & \textbf{\color{red}{0.0584}} \\
 \hline
\end{tabular}
\label{hsi_unmixing}
\end{table}

\subsubsection{Experiment Settings}
\noindent\textbf{Dataset} We adopt the Urban dataset\footnote{\url{https://rslab.ut.ac.ir/data}} to evaluate the performance of HyperSIGMA for hyperspectral unmixing.

\noindent\textbf{Implementation} Following \cite{cnnaeu}, we employ a simple encoder-decoder structure for hyperspectral unmixing. Seven advanced linear and nonlinear unmixing approaches were chosen for comparison. These include three classical methods: FCLS \cite{fcls}, ELMM \cite{elmm}, and SUnSAL \cite{sunsal}, as well as recent deep learning-based methods: CNNAEU \cite{cnnaeu}, CyCU-Net \cite{cycu}, DeepTrans \cite{deeptrans}, and EGU-Net \cite{egunet}. We used two common quantitative evaluation indices: the mean spectral angle distance (mSAD) to compare the similarity between the learned endmembers and references, and the mean squared error (MSE) to measure the quality of the obtained abundance map. Note that lower values indicate better performance. More details about the task setting, network structure, and hyperparameter settings are presented in the appendix.

\subsubsection{Results and Analyses}

The evaluation results are shown in Table \ref{hsi_unmixing}, where we only list the average value across all endmembers. It can be seen that HyperSIGMA achieves the best performance for both endmembers and abundances, demonstrating its superior feature representation capability. HyperSIGMA consistently outperforms SpatSIGMA with the help of spectral information, which is significant for the unmixing task. These findings highlight the potential of our models to enhance hyperspectral low-level tasks. The appendix provides more results and analyses for each endmember.

\subsection{Hyperspectral Image Denoising}
\label{subsec:exp_denoising}

Denoising, a fundamental low-level vision task, is widely discussed across various data forms, such as natural \cite{DBLP:journals/tip/ChenWGWYL23, DBLP:journals/corr/abs-2306-13653}, raw \cite{DBLP:journals/nn/MaWZZ23, DBLP:conf/mm/0002YZW022} and thermal images \cite{DBLP:journals/tgrs/KorosovDMFP22, DBLP:journals/lgrs/MiaoZZ22}. In this section, we consider a classic hyperspectral image restoration task - HSI denoising, which aims at recovering a clean HSI from its noisy observation.

\begin{table}[t]
\scriptsize
\centering
\caption{
Quantitative comparison of different HSI denoising methods on the WDC Mall dataset with non-i.i.d Gaussian, impulse, stripe, and deadline noises. \textbf{\color{red}{Best}} and \textbf{\color{blue}{2nd-best}} results are highlighted.
}
\label{tab:denoising}
\begin{tabular}{lccc}
\hline
Method & PSNR & SSIM & SAM \\ \hline
Noisy & 10.984 & 0.351 & 0.719 \\
LLRT \cite{DBLP:conf/cvpr/ChangYZ17}  & 21.933 & 0.824 & 0.221 \\
NGMeet \cite{DBLP:conf/cvpr/0003YLYZ19}  & 23.452 & 0.872 & 0.260 \\
LRTFL0 \cite{xiong2019hyperspectral}  & 25.558 & 0.907 & 0.198 \\
E-3DTV \cite{peng2020enhanced}  & 25.966 & 0.919 & 0.117 \\
DS2DP \cite{miao2022hyperspectral} & 26.741 & 0.931 & 0.152 \\
QRNN3D \cite{DBLP:journals/tnn/WeiFH21} & 28.097 & 0.945 & 0.132 \\
SST \cite{DBLP:conf/aaai/LiFZ23}  & 28.061 & 0.957 & 0.111 \\ \hline
SpatSIGMA &  \textbf{\color{blue} 28.103} & \textbf{\color{blue} 0.959} & \textbf{\color{blue} 0.109} \\
HyperSIGMA  & \textbf{\color{red} 28.503} & \textbf{\color{red} 0.961} & \textbf{\color{red} 0.106} \\
\hline
\end{tabular}
\end{table}

\subsubsection{Experiment Settings}
\noindent\textbf{Dataset}
For the HSI denoising experiment, we utilize the widely used Washington DC Mall (WDC Mall) dataset\footnote{\url{http://lesun.weebly.com/hyperspectral-data-set.html}}, which contains images with 191 spectral bands and a spatial resolution of 1,208$\times$307.

\noindent\textbf{Implementation} The comparison methods include classical model-based approaches (LLRT \cite{DBLP:conf/cvpr/ChangYZ17}, NGMeet~\cite{DBLP:conf/cvpr/0003YLYZ19}, LRTFL0~\cite{xiong2019hyperspectral}, and E-3DTV~\cite{peng2020enhanced}), and recent deep learning-based methods (DS2DP~\cite{miao2022hyperspectral}, QRNN3D \cite{DBLP:journals/tnn/WeiFH21}, and SST \cite{DBLP:conf/aaai/LiFZ23}). We use three common evaluation metrics: Peak Signal-to-Noise Ratio (PSNR), Structure SIMilarity (SSIM), and Spectral Angle Mapper (SAM). Higher PSNR and SSIM values and lower SAM values indicate better denoising performance. The appendix provides more details about network structure and experimental settings.

To demonstrate the effectiveness of our model, we consider a challenging mixed noise case that contains four types of noises, incorporating Gaussian noise, impulse noise, stripes, and deadlines. Details can be found in the appendix.

\subsubsection{Results and Analyses}

As shown in Table \ref{tab:denoising}, our proposed methods greatly outperform other approaches. For instance, HyperSIGMA achieves a higher PSNR than the second-best method, SST~\cite{DBLP:conf/aaai/LiFZ23}.
Even SpatSIGMA, the simplified version using only spatial information from the pre-trained RS model, performs comparably to SST. We attribute these results to the superior feature representations by pre-training models, especially the utilization of spectral information. For additional results in various noisy cases, please see the appendix.

\subsection{Hyperspectral Image Super-Resolution}
\label{subsec:exp_sr}
In addition to HSI denoising, we investigate another fundamental low-level task: HSI super-resolution, which aims to produce a clearer HSI from its low-resolution observation.

\begin{table}[t]
    \centering
\caption{Quantitative comparison of different methods on the Houston dataset at  8$\times$ scale factors. \textbf{\color{red}{Best}} and \textbf{\color{blue}{2nd-best}} results are highlighted.}
\label{tab:super}
\resizebox{\linewidth}{!}{
    \begin{tabular}{lcccccc}
    \hline
        Method & PSNR & SSIM & SAM & CC & RMSE & ERGAS \\ 
        \hline
        Bicubic & 38.108 & 0.899 & 4.671 & 0.918 & 0.015 & 5.123 \\ 
        GDRRN \cite{li2018single} & 38.259 & 0.909 & 4.905 & 0.914 & 0.014 & 4.914 \\ 
        SwinIR \cite{liang2021swinir} & 39.401 & 0.920 & 4.059 & 0.937 & \textbf{\color{blue}{0.013}} & 4.333 \\ 
        SSPSR \cite{jiang2020learning} & 39.284 & 0.916 & 4.267 & 0.935 & \textbf{\color{blue}{0.013}} & 4.421 \\ 
        RFSR \cite{wang2021hyperspectral} & 39.490 & 0.921 & 3.840 & 0.938 & \textbf{\color{blue}{0.013}} & 4.297 \\ 
        Gelin\cite{wang2022group}  & 39.639 & 0.921 & 3.923 & 0.940 & \textbf{\color{blue}{0.013}} & 4.245 \\ 
        MSDFormer \cite{msdformer} & 39.745 & 0.923 & 3.661 & 0.942 & \textbf{\color{red}{0.012}} & 4.211 \\
        \hline
        SpatSIGMA & \textbf{\color{blue}{39.895}} & \textbf{\color{blue}{0.926}} & \textbf{\color{blue}{3.628}} & 
        \textbf{\color{blue}{0.943}} & \textbf{\color{red}{0.012}} & \textbf{\color{blue}{4.121}} \\
        HyperSIGMA & \textbf{\color{red}{39.940}} & \textbf{\color{red}{0.927}} & \textbf{\color{red}{3.552}} & \textbf{\color{red}{0.944}} & \textbf{\color{red}{0.012}} & \textbf{\color{red}{4.101}} \\\hline
    \end{tabular}
    }
\end{table}

\subsubsection{Experiment Settings}
\noindent\textbf{Dataset}
We employ the Houston dataset\footnote{\url{https://hyperspectral.ee.uh.edu/?page id=1075}} that contains 48 channels with a spatial resolution of 4,172$\times$1,202 for the super-resolution task.

\noindent\textbf{Implementation} In our experiments, we first acquire training patches. These patches are uniformly downsampled using bicubic interpolation. We then upsample the downsampled patches back to the original size using various super-resolution methods. Specifically, we compare our SpatSIGMA and HyperSIGMA with several advanced methods, including Bicubic, SwinIR~\cite{liang2021swinir}, GDRRN~\cite{li2018single}, SSPSR~\cite{jiang2020learning}, RFSR~\cite{wang2021hyperspectral}, GELIN~\cite{wang2022group}, and MSDFormer~\cite{msdformer}. Here, we adopt a difficult case: 8$\times$ super-resolution to better test the model performance.

We evaluate the models using six popular metrics to comprehensively assess performance in both spatial and spectral dimensions: Peak Signal-to-Noise Ratio (PSNR), Structural Similarity Index (SSIM), Spectral Angle Mapper (SAM), Cross Correlation (CC), Root-Mean-Squared Error (RMSE), and Erreur Relative Globale Adimensionnelle de Synthèse (ERGAS). While PSNR, SSIM, and RMSE are standard metrics for natural image restoration, CC, SAM, and ERGAS are common in hyperspectral image fusion tasks. Superior super-resolution performance is indicated by higher PSNR, SSIM, and CC values, and lower SAM, RMSE, and ERGAS values. The appendix offers details of network structure and experimental settings.

\subsubsection{Results and Analyses}

The quantitative results of HSI super-resolution are shown in Table \ref{tab:super}. We can find that the proposed HyperSIGMA significantly outperforms other methods. Notably, it surpasses the recent advanced approach, MSDFormer \cite{msdformer}, across all metrics. These findings further demonstrate the capability of our models in HSI low-level tasks. The appendix showcases the super-resolution results, and more results under other upsampling ratios.

\subsection{Further Investigations and Analyses}
\label{subsec:exp_more}
\subsubsection{Model Scalability}
\label{subsubsec:exp_more_scalability}

\begin{table}[t]
  \caption{
  Fine-tuning accuracies of SpatSIGMA and HyperSIGMA with different ViT backbones on Indian Pines and Xiongan datasets. Base: ViT-Base. Large: ViT-Large. Huge: ViT-Huge.}
  \resizebox{\linewidth}{!}{
    \begin{tabular}{l|llccc}
    \hline
    Method & SpatViT & SpecViT & OA & AA & Kappa \\
    \hline
   \bfseries\textit{Segmentation on Indian Pines}  &  \multicolumn{4}{c}{} \\
    \hline
    SpatSIGMA &  Base   & - & 85.08  & 78.30 & 83.04 \\
    SpatSIGMA &  Large  & - & 86.55 & 81.70 & 84.63 \\
    SpatSIGMA & Huge & - & 86.74 & 75.02 & 84.89 \\
    \hline
    HyperSIGMA  &  Base  & Base & 85.54 & 76.68  & 83.58  \\
    HyperSIGMA  &  Base  & Large & 86.81 & 79.49 & 85.00 \\
    HyperSIGMA  &  Base  & Huge  & 87.38 & 83.24 & 85.63 \\
    HyperSIGMA  &  Huge  & Huge & 87.70  & 79.19  & 86.01 \\
   \hline
   \bfseries\textit{Classification on Xiongan}  &  \multicolumn{4}{c}{} \\
    \hline
    1DCNN \cite{3dcnn}  &  -   & - & 55.89  & 71.19 & 52.26 \\
    2DCNN \cite{3dcnn}  &  -   & - & 27.32  &  38.79 &  23.61 \\
    HybridSN \cite{hybridsn} &  -   & - &  89.83 &  74.72 & 88.39  \\
    A$^2$S$^2$KResNet \cite{a2s2kresnet} &  -   & - &  90.32 &  76.85 & 88.93  \\
    \hline
    SpatSIGMA &  Base   & - &  93.68 & 79.80 & 92.75 \\
    SpatSIGMA &  Large  & - & 94.59  &  82.95 & 93.78 \\
    SpatSIGMA & Huge & - & 94.73 & 82.43 & 93.94 \\
    \hline
    HyperSIGMA  &  Base  & Base & 95.06 & 82.43 & 94.32 \\
    HyperSIGMA  &  Base  & Large & 95.30 & 82.77 & 94.58 \\ 
    HyperSIGMA  &  Base  & Huge  &95.65  & 83.85 & 94.98 \\
    HyperSIGMA  &  Huge  & Huge  & 95.78 & 84.19  & 95.13 \\
    \hline
    \end{tabular}
    }
    \label{table:model_scalable}
\end{table}

In previous sections, we have shown that HyperSIGMA consistently outperforms state-of-the-art methods, even when using the ViT-Base backbone. To evaluate the scalability of HyperSIGMA, we fine-tuned SpatSIGMA and HyperSIGMA using different ViT backbones on the Indian Pines dataset and the more challenging Xiongan dataset \cite{xiongan}, which is 1,580$\times$3,750 pixels with 256 channels and 19 categories, mainly comprising cash crops. We also implemented some classical methods for comparison, including 1DCNN and 2DCNN \cite{3dcnn}, HybridSN \cite{hybridsn}, and A$^2$S$^2$KResNet \cite{a2s2kresnet}. Table \ref{table:model_scalable} presents the results. It is clear that using larger ViT backbones leads to better performance for both SpatSIGMA and HyperSIGMA, underscoring their good scalability. Notably, we observed that ViT-Huge in SpatSIGMA did not perform as well as lighter versions used in HyperSIGMA, highlighting the importance of utilizing spectral information. The largest HyperSIGMA, with over 1 billion parameters, achieved 87.70\% OA and 95.78\% OA on Indian Pines and Xiongan, respectively, surpassing other methods significantly. The experimental settings and more analyses related to the model scalability can be found in the appendix.

\subsubsection{Model Robustness}
\label{subsubsec:exp_more_robustness}

\begin{figure}[t]
\centering
\subfigure[]{\includegraphics[width=0.5\linewidth \hspace{-0.15em}]{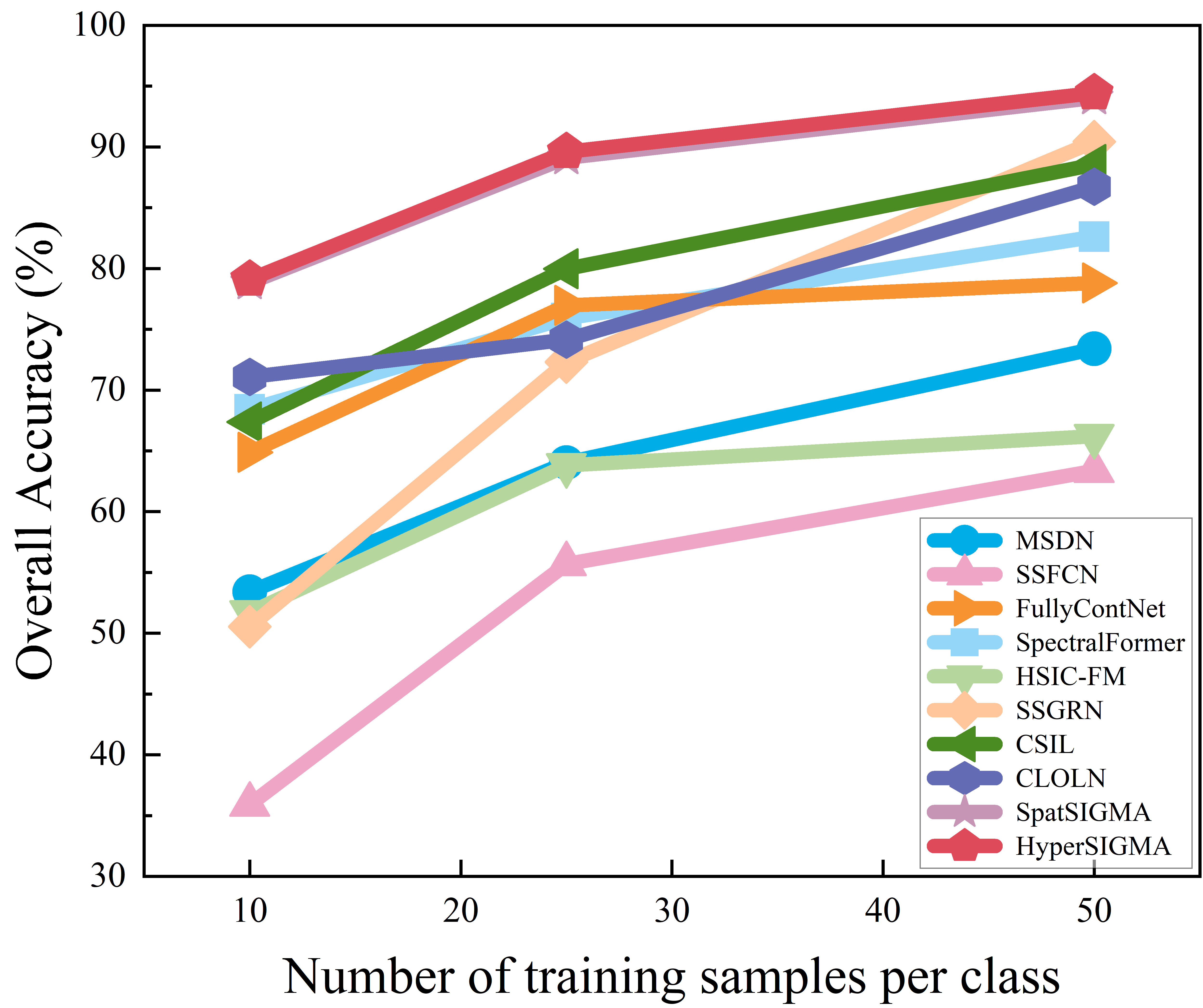}}
\subfigure[]{\includegraphics[width=0.5\linewidth \hspace{-0.15em}]{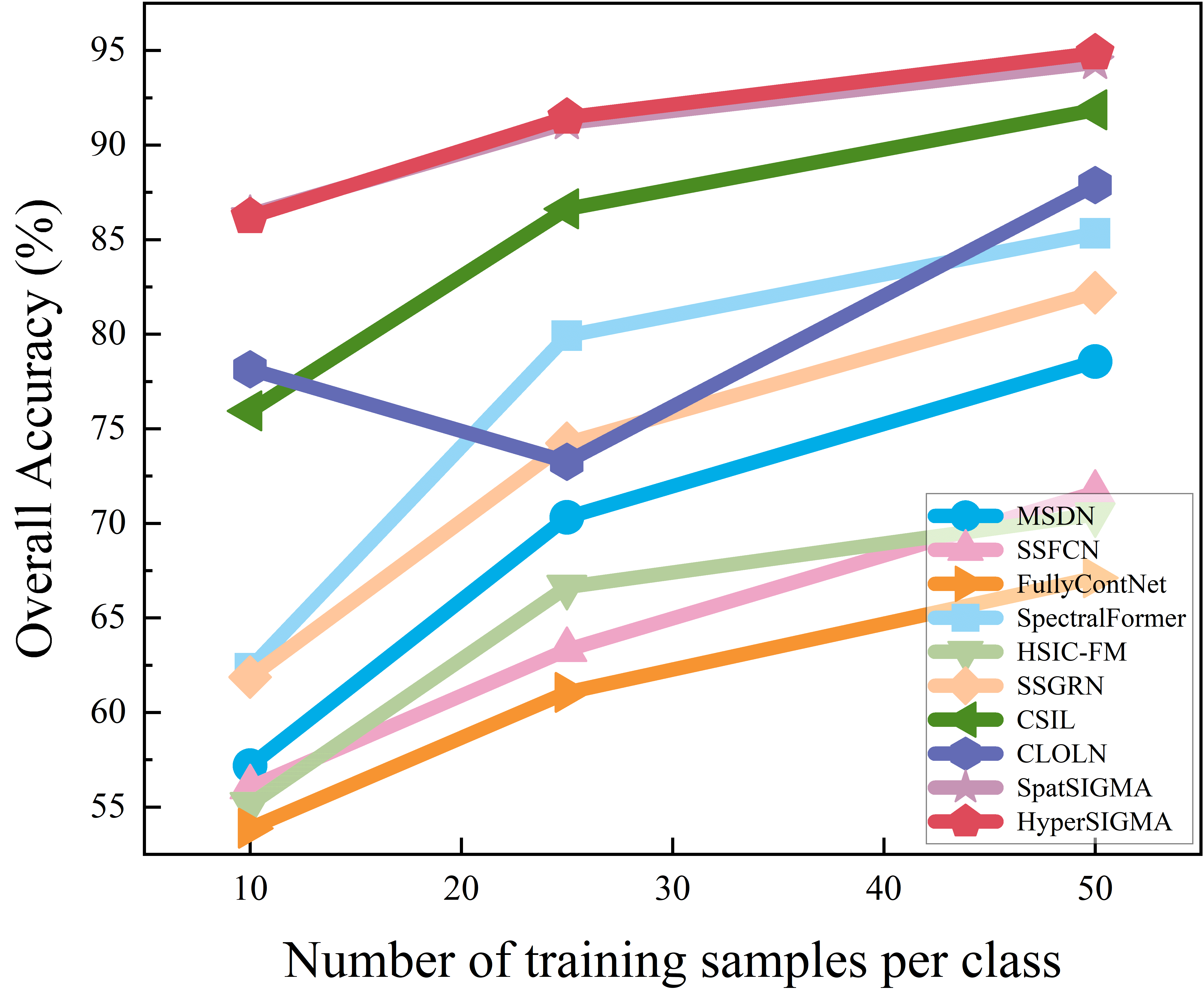}}
\caption{Overall accuracies of different methods with limited training samples: (a) HanChuan dataset. (b) HongHu dataset.}
\label{fig:FewerTrainingSamples}
\end{figure}

\noindent\textbf{Fine-Tuning with Limited Training Samples} To evaluate model robustness, we conducted experiments using limited samples for training. Annotating HSI is challenging and costly due to varying ambient light, spectral nonlinearities, and diverse object types, leading to a scarcity of labeled samples \cite{ITER}. Therefore, it is crucial to assess model performance with limited training samples. We used the HanChuan and HongHu datasets, with 10, 25, and 50 samples per class for training.

Figure~\ref{fig:FewerTrainingSamples} presents the accuracies of different methods on these datasets with limited training samples. Our models consistently achieved the highest accuracy regardless of the number of training samples. For example, on the HanChuan dataset, when only 10 samples per class were used (0.04\% of the total samples), HyperSIGMA achieved 79.10\% accuracy, which is 8\% higher than the second-best approach, CLOLN~\cite{CLOLN}. Moreover, our methods present minimal accuracy decreases under the reduction of labels. In summary, our model demonstrates strong robustness even with limited training samples.

\noindent\textbf{Robustness to Adversarial Attack} Previous research has shown that HSI interpretation methods using deep neural networks are vulnerable to adversarial attack \cite{SACNet,hsi_aa}. This motivates us to evaluate the stability of our proposed models against such attacks.

\begin{table}[t]
\scriptsize
 \centering
{ 
\caption{
OA (\%) of different methods under various attacks on the Indian Pines dataset. CTS: Clean Test Set. \textbf{\color{red}{Best}} and \textbf{\color{blue}{2nd-best}} results are highlighted.
}\label{table:Clean test}
\begin{tabular}{lccc}
\hline
 Method       &  CTS   &  FGSM     &  PGD       \\
\hline
 SSFCN   \cite{ssfcn}    &   82.57  &   60.26   &   1.48   \\
 SACNet  \cite{SACNet}    &   88.63 &   80.19   &    30.18  \\
 RCCA    \cite{RCCA}     &  94.51  &  91.33    &    82.96  \\
 S$^{3}$ANet \cite{S3ANet}&   96.23  &    95.48  &   42.01   \\
\hline
 SpatSIGMA &   \textbf{\color{blue}{96.28}}   &  \textbf{\color{blue}{96.19}}   &  \textbf{\color{blue}{95.23}}   \\
 HyperSIGMA &  \textbf{\color{red}{96.76}}   &   \textbf{\color{red}{96.62}}    &   \textbf{\color{red}{95.92}}   \\
\hline                 
\end{tabular}
}
\end{table}

\begin{figure}[t]
\centering
\includegraphics[width=0.7\linewidth]{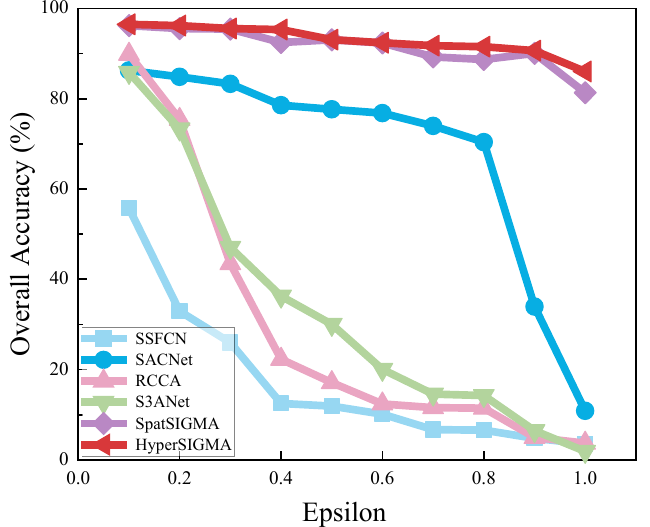}
\caption{Classification performance of various methods under different FGSM attack perturbation values on the Indian Pines dataset.}
\label{fig:Attack_epilson}
\end{figure}

We use the Indian Pines dataset, commonly utilized in adversarial attack studies, for our experiment. For training, we select 50 samples per class to better highlight accuracy variations, with the remainder used as the clean testing set. We employ two classical attack methods: FGSM \cite{FGSM} and PGD \cite{PGD}, with perturbation budgets ($\epsilon$) uniformly set to 0.1. Our models are compared with common classification methods as well as those specially designed to defend against adversarial attacks, including SSFCN \cite{ssfcn}, SACNet \cite{SACNet}, RCCA \cite{RCCA}, and S$^{3}$ANet \cite{S3ANet}. We fine-tune using the AdamW optimizer, with an initial learning rate of 0.0005 and a weight decay of 0.00005, for 1,000 epochs.

Table~\ref{table:Clean test} presents the classification accuracy on the original clean testing set and under various attacks. Our models show minimal accuracy decreases and consistently outperform others, regardless of the attack type. Compared to SpatSIGMA, HyperSIGMA, which leverages spectral information, achieves even higher accuracy. To further assess model stability, we increase the FGSM perturbation budget $\epsilon$ and plot the accuracy changes in Figure~\ref{fig:Attack_epilson}. Most comparison methods' accuracies decline rapidly when $\epsilon$ reaches 0.4. In contrast, our models maintain stability, with only slight accuracy changes even at $\epsilon$ of 1.0. These results clearly demonstrate the robustness of our models against adversarial attacks.

\noindent\textbf{Robustness to Image Degradation} We further evaluate the stability of HyperSIGMA by testing its performance on degraded images. Specifically, we use two methods to degrade hyperspectral images. First, we apply image compression for saving storage space and transmission bandwidth. Evaluating model performance on compressed HSIs is of practical significance, as current HSI approaches under image compression remain under-explored. We conduct experiments using JPEG compression at a bit rate of 0.33901, a widely-used standard for image compression\cite{JPEG}. Second, we consider complex imaging conditions such as sunlight, atmosphere, and terrain, which often result in noisy HSIs. To simulate real-world scenarios, we manually add i.i.d zero-mean Gaussian noise with a variance of 70. Image compression may eliminate valuable data, whereas noise can cause interference, thereby reducing image quality.

\begin{figure}[t]
\centering
\includegraphics[width=0.7\linewidth]{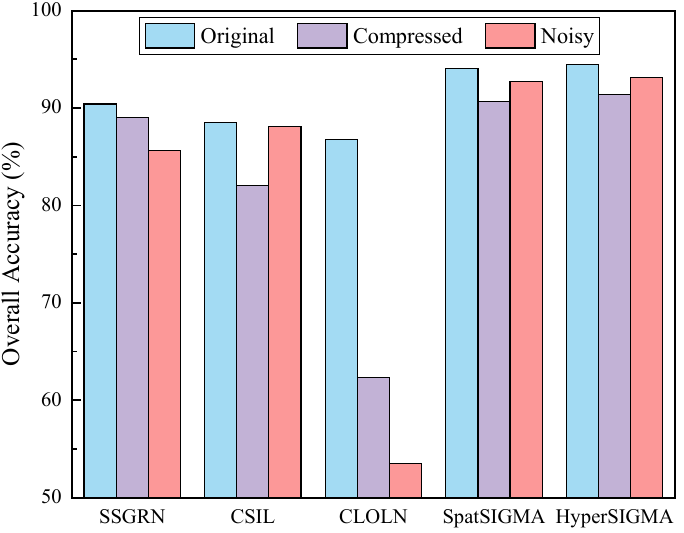}
\caption{Classification performance of various methods under different image degradation strategies on the HanChuan dataset.}
\label{fig:data_degradation}
\end{figure}

We selected competitive approaches, SSGRN \cite{ssgrn}, CSIL \cite{yang_csil} and CLOLN \cite{CLOLN} for comparison in the classification task on the HanChuan dataset. The results in Fig. \ref{fig:data_degradation} show a noticeable decline in performance for comparison methods when subjected to image compression or noise addition. In contrast, the OAs of SpatSIGMA and HyperSIGMA remain stable. Even under degraded image conditions, our models consistently outperform other state-of-the-art methods, highlighting the robustness of HyperSIGMA in practical settings. It should be noted that unlike previous adversarial attacks, which typically affect only the testing set, the data degradation in this study alters the entire HSI, demonstrating the comprehensive robustness of our approach.

\subsubsection{Cross-modal Transferability}
\label{subsubsec:exp_more_crossmodal}

Given the similarity between HSIs and multispectral images (MSIs), we hypothesized that HyperSIGMA could be effectively applied to multispectral tasks. To test this, we conducted experiments on two related tasks: Hyperspectral-Multispectral Collaborative Classification and Multispectral Change Detection.

\noindent\textbf{Hyperspectral-Multispectral Collaborative Classification} We investigate the integration of MSIs with HSIs for classification tasks. Specifically, we utilize both types of imagery collaboratively and compares its performance against several state-of-the-art methods, including CGCDL \cite{cgcdl}, CDLS \cite{cdls}, Cospace1 \cite{cospace1}, Cospace2 \cite{cospace2}, J-C \cite{jc} and SPDDA \cite{spdda}, using the YRE\_2 dataset \cite{spdda} (see Table \ref{table:yre}). It can be seen that both SpatSIGMA and HyperSIGMA achieve competitive performances, demonstrating the superior abilities of our models in adapting the out-of-modal data.

\noindent\textbf{Multispectral Change Detection} In this task, we utilize the OSCD dataset \cite{oscd}, which consists of Sentinel-2 images. We opted for a simple approach \cite{Chen2023Exchange} to highlight HyperSIGMA's feature extraction capabilities, avoiding the influence of a complex change decoder. Table \ref{table:OSCD} compares HyperSIGMA's performance with leading methods, including FC-EF \cite{CayeDaudt2018Change}, FC-Siam-Diff \cite{CayeDaudt2018Change}, FC-Siam-Conc \cite{CayeDaudt2018Change}, SiamCRNN \cite{Chen2019Change}, SNUNet \cite{Fang2022SNUNet}, DSIFN \cite{Zhang2020Change}, BIT \cite{hsi_cd_bit}, ChangeFormer \cite{Bandara2022Transformer}, and ChangeMamba \cite{chen2024changemamba}. We also compared our method with SpectralGPT \cite{spectralgpt}, a recent model tailored for multispectral data. HyperSIGMA surpassed the state-of-the-art ChangeMamba, achieving an F1 score of 58.53\% versus 57.20\%. Further enhancement using spectral information boosted HyperSIGMA's performance to an F1 score of 59.28\%, marking a 4.99\% improvement over SpectralGPT. The implementation details of the cross-modal transferability experiment can be found in the appendix.

\begin{table}[t]
\scriptsize
    \centering
    \caption{
    Collaborative classification results of different methods on the YRE\_2 dataset. \textbf{\color{red}{Best}} and \textbf{\color{blue}{2nd-best}} results are highlighted.
    }
    
    \begin{tabular}{lccc}
    \hline
        Method & OA  & AA & Kappa  \\ \hline
        CGCDL \cite{cgcdl} & 69.01 & 60.10 & 58.42  \\ 
        CDLS \cite{cdls} &50.42  & 47.76 & 36.95 \\ 
        Cospace1 \cite{cospace1} & 70.42 & 63.14  & 60.62 \\ 
        Cospace2 \cite{cospace2} & 71.81 & 63.49 & 62.16 \\
        J-C \cite{jc} &72.82  & 63.91  & 63.15  \\
        SPDDA \cite{spdda} & 75.29  & 66.02 & 66.46 \\
        \hline
        SpatSIGMA & \textbf{\color{blue}{75.51}}  & \textbf{\color{red}{78.66}}  &  \textbf{\color{blue}{67.69}} \\ 
        HyperSIGMA & \textbf{\color{red}{75.67}} & \textbf{\color{blue}{78.49}}  & \textbf{\color{red}{67.83}}  \\ \hline
    \end{tabular}
    
    \label{table:yre}
\end{table}

\begin{table}[t]
\scriptsize
    \centering
    \caption{
    Change detection results of different methods on the OSCD dataset. SpectralGPT accuracy values are sourced from \cite{spectralgpt}. \textbf{\color{red}{Best}} and \textbf{\color{blue}{2nd-best}} results are highlighted.
    }
    \begin{tabular}{lccccc}
    \hline
        Method & OA & Kappa & F1 & Precision & Recall  \\ \hline
        
        FC-EF \cite{CayeDaudt2018Change} & 94.80 & 45.10 & 47.83 & 49.67 & 46.12 \\ 
        FC-Siam-Diff \cite{CayeDaudt2018Change} & 94.06 & 45.26& 48.38 & 43.91 & 53.85\\ 
        FC-Siam-Conc \cite{CayeDaudt2018Change} & 94.55 & 47.57& 50.43& 47.60 & 53.62\\ 
        SiamCRNN \cite{Chen2019Change} & 95.53 & 52.08 & 54.42 & 57.58 & 51.60\\
        SNUNet \cite{Fang2022SNUNet} & 93.68 & 39.70 & 43.03 & 40.29 & 46.16 \\
        I3PE \cite{Chen2023Exchange} & 95.57 & 50.27 & 52.57 & 58.79 & 47.54 \\
        DSIFN \cite{Zhang2020Change} & \textbf{\color{blue}{96.04}} & 52.22 & 54.21 & \textbf{\color{red}{67.32}} & 45.38 \\
        BIT \cite{hsi_cd_bit} & 94.88 & 43.82 & 41.19 & 50.56 & 38.66 \\ 
        ChangeFormer \cite{Bandara2022Transformer} & 95.40 & 52.15 & 54.57 & 55.66 & 53.51 \\ 
        ChangeMamba  \cite{chen2024changemamba} & \textbf{\color{blue}{96.04}} & 54.82 & 57.20 & 56.08 & \textbf{\color{blue}{58.36}} \\ 
        SpectralGPT\cite{spectralgpt}  & - & - & 54.29 & 52.39 & 57.20 \\ 
        \hline
        SpatSIGMA & \textbf{\color{red}{96.08}} & \textbf{\color{blue}{56.49}} & \textbf{\color{blue}{58.53}} & \textbf{\color{blue}{64.59}} & 53.50   \\ 
        HyperSIGMA & 95.78 & \textbf{\color{red}{57.06}} & \textbf{\color{red}{59.28}} & 59.12 & \textbf{\color{red}{59.45}} \\ \hline
    \end{tabular}
    \label{table:OSCD}
\end{table}

\subsubsection{Real-world Applicability}
\label{subsubsec:exp_more_applications}

\begin{figure}[t]
\centering
\includegraphics[width=0.9\linewidth]{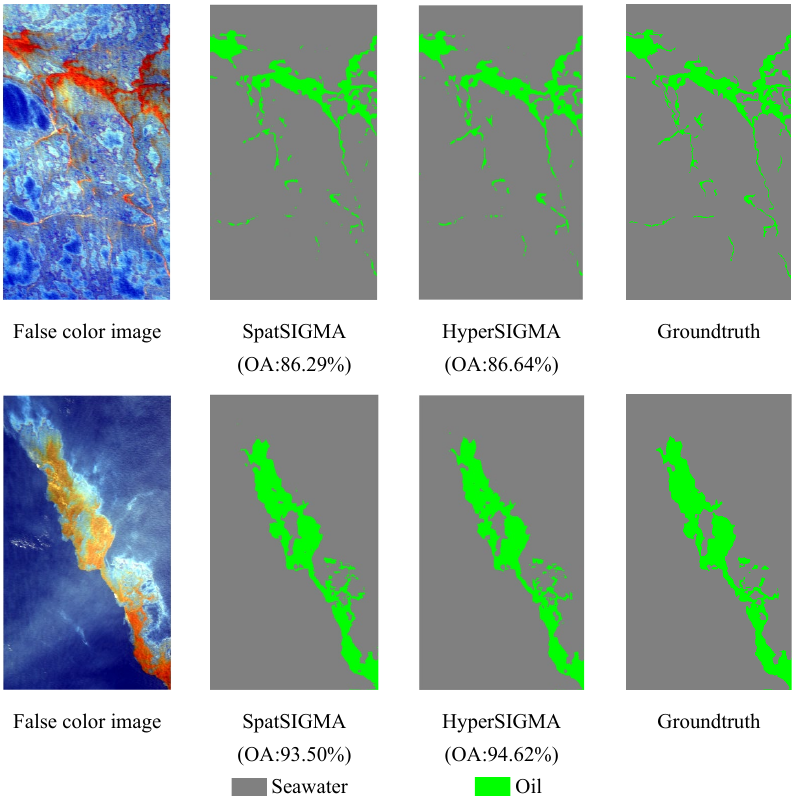}
\caption{Visualization of detected oil leakage in various Gulf of Mexico regions by our models. The first row: GM13. The second row: GM18.
}
\label{fig:Oil}
\end{figure}

We further evaluated the model's performance in practical applications, specifically focusing on offshore oil leak detection. As a case study, we examined the Gulf of Mexico oil spill on April 20, 2010, where over 200 million gallons of crude oil were released, marking the largest offshore oil disaster in U.S. history. HSIs are advantageous for monitoring such oil spills due to their rich spectral information. We utilized a hyperspectral dataset related to this event \cite{mexico_gulf_oil}, covering the Gulf of Mexico (GM) area. Following the same implementation procedure as previous segmentation tasks, we used 50 pixel-level samples each for both the oil spill area and the background seawater as training data. Fig. \ref{fig:Oil} presents the results, including pseudo-color images of the study area, our detection results, and ground truth. The results demonstrate that SpatSIGMA and HyperSIGMA effectively distinguish between seawater and oil film, even in areas with minimal oil leakage, enabling precise detection. Notably, HyperSIGMA outperformed SpatSIGMA due to its use of spectral information. These findings strongly support the universality and practical potential of our models.

\subsubsection{Model Computational Efficiency}
\label{subsubsec:exp_complexity}

\begin{figure}[t]
\centering
\subfigure[]{\includegraphics[width=0.49\linewidth]{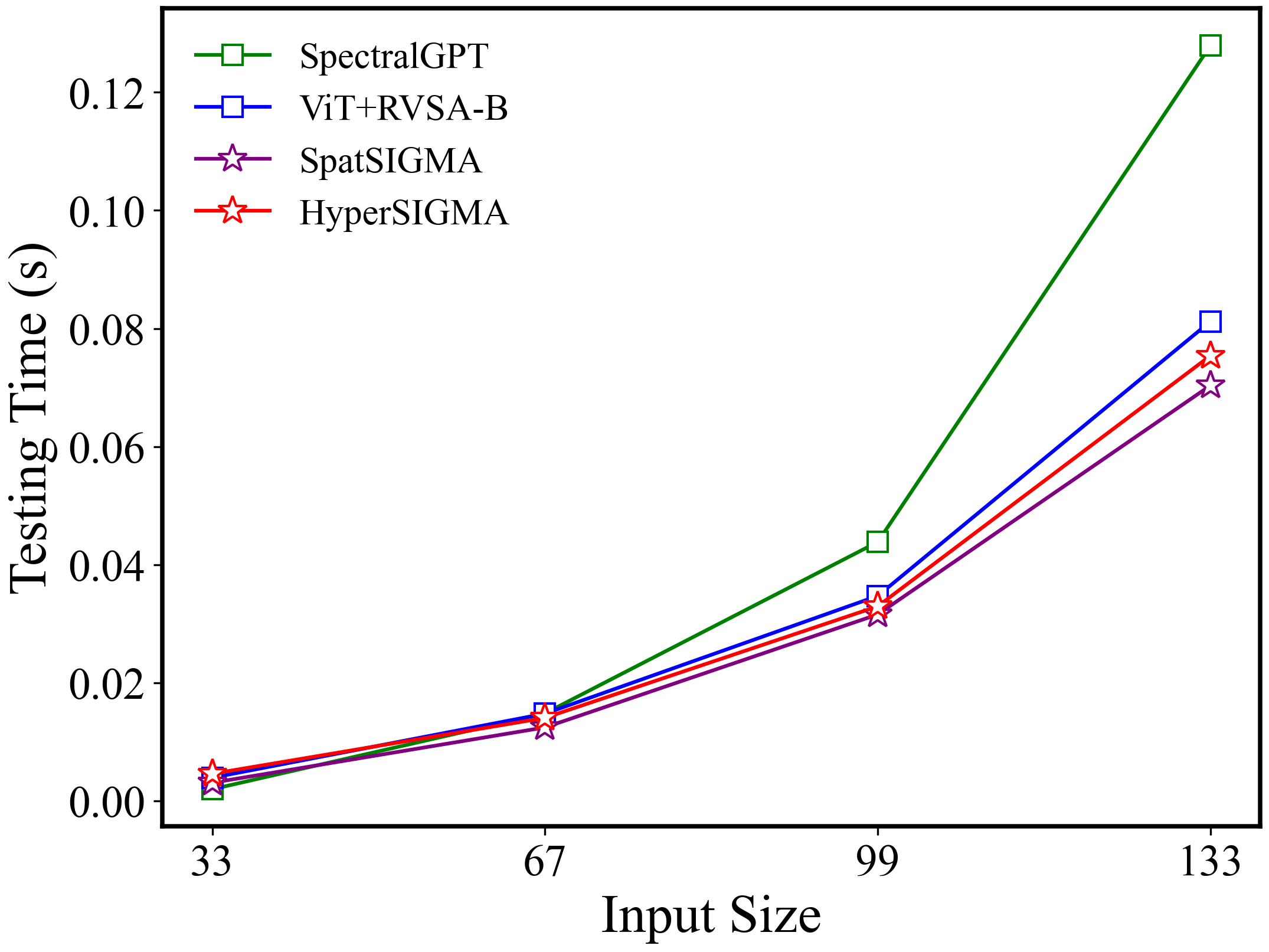}\hspace{-0.15em}}
\subfigure[]{\includegraphics[width=0.5\linewidth]{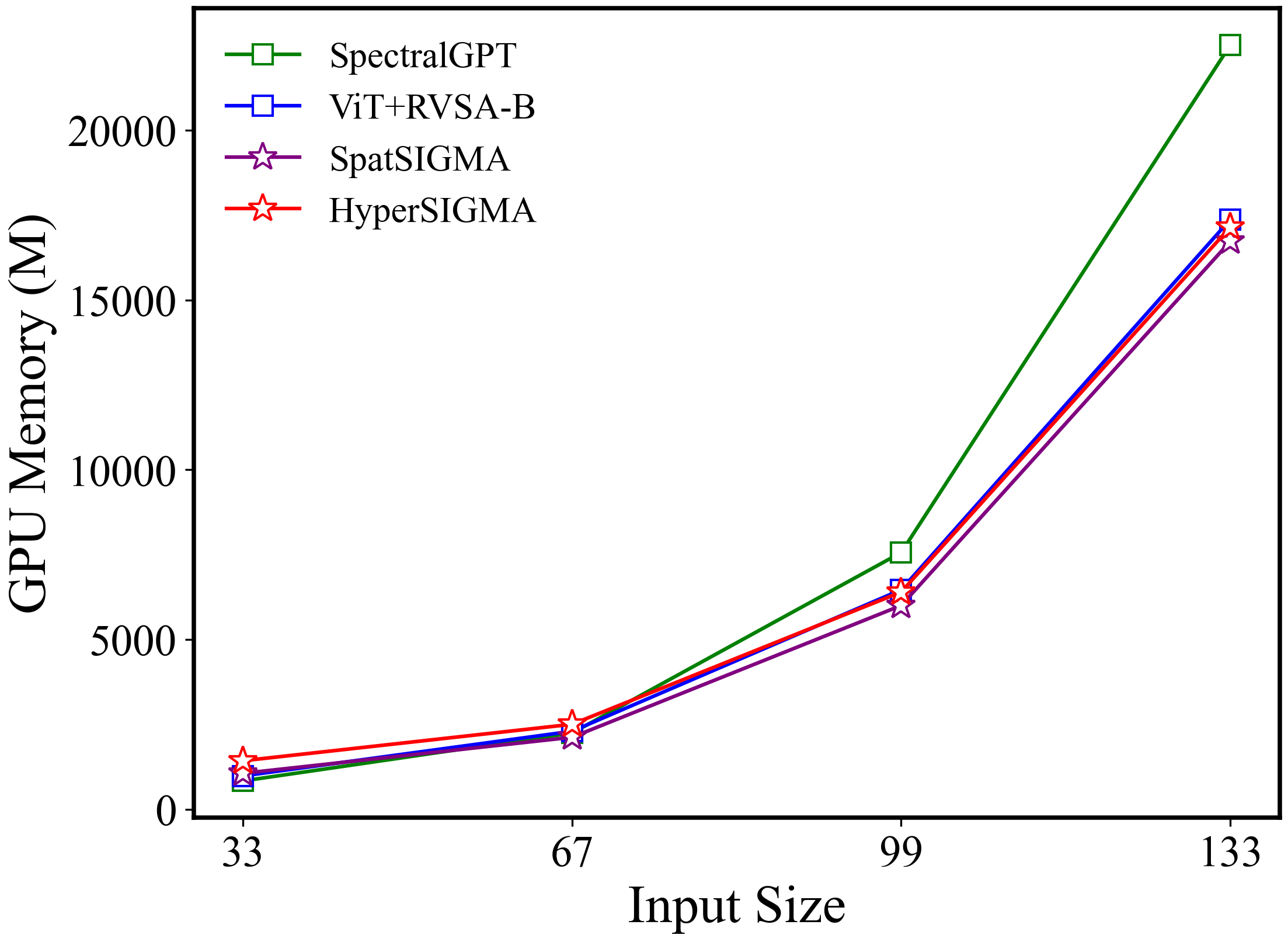}\hspace{-0.15em}}
\caption{Inference time and memory footprint of different models with varying input patch sizes, where inference time is measured for a single patch. All methods use a ViT-B backbone, and experiments are conducted on an NVIDIA V100 GPU. (a) Inference time. (b) GPU memory.}
\label{fig:model_complexity}
\end{figure}

Finally, we analyze the computational cost of our models. As discussed in Section \ref{sec:related_work}, vision transformer has become a primary choice for RS foundation models, with many methods directly adopting or advancing the original ViT structure \cite{vit_g, scale_mae, spectralgpt, dofa, vit} through modifications \cite{vsa, rvsa}. Since HyperSIGMA is also built on ViT, the main distinction between our models and existing methods lies in the attention mechanism. Theoretically, as outlined in Section \ref{subsubsec:ssa}, the proposed SSA has a complexity of $O(3N^2D'+12NN_pD'+NN_p)$ (ignoring the number of heads), which is comparable to window attention \cite{swint}, VSA \cite{vsa} and RVSA \cite{rvsa}. To validate this, we measure the inference time and GPU memory footprint of our models and typical RS foundation models, including SpectralGPT \cite{spectralgpt} and RVSA \cite{rvsa}, as shown in Fig. \ref{fig:model_complexity}. The results show that our models exhibit similar inference time and memory footprint, with a slight advantage for larger patches, even for the two-branch HyperSIGMA, when compared to ViT-B+RVSA. These advantages are particularly evident when compared to SpectralGPT, which employs full attention, especially with larger patches. These findings underscore the efficiency of our models. Further SSA complexity analysis and experimental details are provided in the appendix.

\section{Conclusion}

In this paper, we present HyperSIGMA, the first hyperspectral foundation model with over 1 billion parameters for hyperspectral image (HSI) interpretation. Furthermore, we curate HyperGlobal-450K, the largest hyperspectral dataset to date, composed of HSIs from around the world, establishing a solid basis for self-supervised pre-training research. To address HSI redundancy, we propose a novel sparse sampling attention mechanism, enabling adaptive perception of relevant contextual regions with few learnable sampling points. We also design a spectral enhancement module to achieve effective spatial-spectral feature fusion. Comprehensive evaluations on both high-level and low-level hyperspectral tasks demonstrate HyperSIGMA's superior performance. Further analysis reveals its excellent scalability, robustness, cross-modal transferability, and computational efficiency, making it suitable for various real-world applications.

We find that HyperSIGMA offers only limited improvements over SpatSIGMA in some cases. This may be due to the challenges of recovering complete channels for pre-training the spectral subnetwork. Further investigation will delve into refining both the pre-training pretext task and network architecture to enhance the spectral foundation model.
Despite this, we believe our findings provide valuable guidance to the hyperspectral community on advancing foundation models. We hope that HyperSIGMA, with its superior performance and excellent properties, will be widely adopted in various applications.

\section*{acknowledgement}
The model pre-training was supported by The Dawning Information Industry Co., Ltd., with fine-tuning conducted on the supercomputing system at the Supercomputing Center of Wuhan University. Thanks Wentao Jiang from the School of Computer Science at Wuhan University for implementing the accelerated version of HyperSIGMA (Sec. \ref{subsubsec:exp_complexity}).

\bibliographystyle{IEEEtran}
\bibliography{HyperSIGMA}

\begin{thebibliography}{100}
\providecommand{\url}[1]{#1}
\csname url@samestyle\endcsname
\providecommand{\newblock}{\relax}
\providecommand{\bibinfo}[2]{#2}
\providecommand{\BIBentrySTDinterwordspacing}{\spaceskip=0pt\relax}
\providecommand{\BIBentryALTinterwordstretchfactor}{4}
\providecommand{\BIBentryALTinterwordspacing}{\spaceskip=\fontdimen2\font plus
\BIBentryALTinterwordstretchfactor\fontdimen3\font minus \fontdimen4\font\relax}
\providecommand{\BIBforeignlanguage}[2]{{%
\expandafter\ifx\csname l@#1\endcsname\relax
\typeout{** WARNING: IEEEtran.bst: No hyphenation pattern has been}%
\typeout{** loaded for the language `#1'. Using the pattern for}%
\typeout{** the default language instead.}%
\else
\language=\csname l@#1\endcsname
\fi
#2}}
\providecommand{\BIBdecl}{\relax}
\BIBdecl

\bibitem{imaging_spectrometry}
A.~F. Goetz, G.~Vane, J.~E. Solomon, and B.~N. Rock, ``Imaging spectrometry for earth remote sensing,'' \emph{Science}, vol. 228, no. 4704, pp. 1147--1153, 1985.

\bibitem{earth_data_sys_sci}
M.~Reichstein, G.~Camps-Valls, B.~Stevens, M.~Jung, J.~Denzler, N.~Carvalhais, and f.~Prabhat, ``Deep learning and process understanding for data-driven earth system science,'' \emph{Nature}, vol. 566, no. 7743, pp. 195--204, 2019.

\bibitem{spectralgpt}
D.~Hong, B.~Zhang, X.~Li, Y.~Li, C.~Li, J.~Yao, N.~Yokoya, H.~Li, P.~Ghamisi, X.~Jia, A.~Plaza, P.~Gamba, J.~A. Benediktsson, and J.~Chanussot, ``{SpectralGPT}: Spectral remote sensing foundation model,'' \emph{IEEE Transactions on Pattern Analysis and Machine Intelligence}, pp. 1--18, 2024.

\bibitem{rs_hsi_cls}
B.~Manifold, S.~Men, R.~Hu, and D.~Fu, ``A versatile deep learning architecture for classification and label-free prediction of hyperspectral images,'' \emph{Nature Machine Intelligence}, vol.~3, no.~4, pp. 306--315, 2021.

\bibitem{rs_globalwaste}
X.~Sun, D.~Yin, F.~Qin, H.~Yu, W.~Lu, F.~Yao, Q.~He, X.~Huang, Z.~Yan, P.~Wang \emph{et~al.}, ``Revealing influencing factors on global waste distribution via deep-learning based dumpsite detection from satellite imagery,'' \emph{Nature Communications}, vol.~14, no.~1, p. 1444, 2023.

\bibitem{hsi_forest_manage}
A.~M. Lechner, G.~M. Foody, and D.~S. Boyd, ``Applications in remote sensing to forest ecology and management,'' \emph{One Earth}, vol.~2, no.~5, pp. 405--412, 2020.

\bibitem{hsi_monitor_biodiversity}
D.~M. Griffith, K.~B. Byrd, L.~D. Anderegg, E.~Allan, D.~Gatziolis, D.~Roberts, R.~Yacoub, and R.~R. Nemani, ``Capturing patterns of evolutionary relatedness with reflectance spectra to model and monitor biodiversity,'' \emph{Proceedings of the National Academy of Sciences}, vol. 120, no.~24, p. e2215533120, 2023.

\bibitem{hsi_urban_plan}
S.~{van der Linden} and P.~Hostert, ``The influence of urban structures on impervious surface maps from airborne hyperspectral data,'' \emph{Remote Sensing of Environment}, vol. 113, no.~11, pp. 2298--2305, 2009.

\bibitem{agriculture_1}
X.~{Zhang}, Y.~{Sun}, K.~{Shang}, L.~{Zhang}, and S.~{Wang}, ``Crop classification based on feature band set construction and object-oriented approach using hyperspectral images,'' \emph{IEEE Journal of Selected Topics in Applied Earth Observations and Remote Sensing}, vol.~9, no.~9, pp. 4117--4128, 2016.

\bibitem{hsi_env_monitor}
W.~Obermeier, L.~Lehnert, M.~Pohl, S.~{Makowski Gianonni}, B.~Silva, R.~Seibert, H.~Laser, G.~Moser, C.~Müller, J.~Luterbacher, and J.~Bendix, ``Grassland ecosystem services in a changing environment: The potential of hyperspectral monitoring,'' \emph{Remote Sensing of Environment}, vol. 232, p. 111273, 2019.

\bibitem{ica_rgf}
J.~Xia, L.~Bombrun, T.~Adalı, Y.~Berthoumieu, and C.~Germain, ``Spectral–spatial classification of hyperspectral images using ica and edge-preserving filter via an ensemble strategy,'' \emph{IEEE Transactions on Geoscience and Remote Sensing}, vol.~54, no.~8, pp. 4971--4982, 2016.

\bibitem{hughes}
G.~Hughes, ``On the mean accuracy of statistical pattern recognizers,'' \emph{IEEE Transactions on Information Theory}, vol.~14, no.~1, pp. 55--63, 1968.

\bibitem{dr_fs}
C.-I. Chang, Q.~Du, T.-L. Sun, and M.~Althouse, ``A joint band prioritization and band-decorrelation approach to band selection for hyperspectral image classification,'' \emph{IEEE Transactions on Geoscience and Remote Sensing}, vol.~37, no.~6, pp. 2631--2641, 1999.

\bibitem{dr_fe}
L.~Bruce, C.~Koger, and J.~Li, ``Dimensionality reduction of hyperspectral data using discrete wavelet transform feature extraction,'' \emph{IEEE Transactions on Geoscience and Remote Sensing}, vol.~40, no.~10, pp. 2331--2338, 2002.

\bibitem{hsi_mp}
M.~Fauvel, J.~A. Benediktsson, J.~Chanussot, and J.~R. Sveinsson, ``Spectral and spatial classification of hyperspectral data using svms and morphological profiles,'' \emph{IEEE Transactions on Geoscience and Remote Sensing}, vol.~46, no.~11, pp. 3804--3814, 2008.

\bibitem{hsi_sp}
L.~Fang, S.~Li, X.~Kang, and J.~A. Benediktsson, ``Spectral–spatial classification of hyperspectral images with a superpixel-based discriminative sparse model,'' \emph{IEEE Transactions on Geoscience and Remote Sensing}, vol.~53, no.~8, pp. 4186--4201, 2015.

\bibitem{uniadrs}
J.~Li, X.~Wang, H.~Zhao, and Y.~Zhong, ``Learning a cross-modality anomaly detector for remote sensing imagery,'' \emph{IEEE Transactions on Image Processing}, vol.~33, pp. 6607--6621, 2024.

\bibitem{3dcnn}
Y.~Chen, H.~Jiang, C.~Li, X.~Jia, and P.~Ghamisi, ``Deep feature extraction and classification of hyperspectral images based on convolutional neural networks,'' \emph{IEEE Transactions on Geoscience and Remote Sensing}, vol.~54, no.~10, pp. 6232--6251, 2016.

\bibitem{ssun}
Y.~Xu, L.~Zhang, B.~Du, and F.~Zhang, ``Spectral-spatial unified networks for hyperspectral image classification,'' \emph{IEEE Transactions on Geoscience and Remote Sensing}, vol.~56, no.~10, pp. 5893--5909, 2018.

\bibitem{spectralformer}
D.~Hong, Z.~Han, J.~Yao, L.~Gao, B.~Zhang, A.~Plaza, and J.~Chanussot, ``{SpectralFormer}: Rethinking hyperspectral image classification with transformers,'' \emph{IEEE Transactions on Geoscience and Remote Sensing}, vol.~60, pp. 1--15, 2022.

\bibitem{hsidmamba}
Y.~Liu, J.~Xiao, Y.~Guo, P.~Jiang, H.~Yang, and F.~Wang, ``{HSIDMamba}: Exploring bidirectional state-space models for hyperspectral denoising,'' \emph{arXiv preprint arXiv:2404.09697}, 2024.

\bibitem{fm_agi}
N.~Fei, Z.~Lu, Y.~Gao, G.~Yang, Y.~Huo, J.~Wen, H.~Lu, R.~Song, X.~Gao, T.~Xiang \emph{et~al.}, ``Towards artificial general intelligence via a multimodal foundation model,'' \emph{Nature Communications}, vol.~13, no.~1, p. 3094, 2022.

\bibitem{fm_definition}
R.~Bommasani, D.~A. Hudson, E.~Adeli, R.~Altman, S.~Arora, S.~von Arx, M.~S. Bernstein, J.~Bohg, A.~Bosselut, E.~Brunskill \emph{et~al.}, ``On the opportunities and risks of foundation models,'' \emph{arXiv preprint arXiv:2108.07258}, 2021.

\bibitem{selfattention}
A.~Vaswani, N.~Shazeer, N.~Parmar, J.~Uszkoreit, L.~Jones, A.~N. Gomez, L.~u. Kaiser, and I.~Polosukhin, ``Attention is all you need,'' in \emph{NeurIPS}, vol.~30, 2017.

\bibitem{xu2021vitae}
Y.~Xu, Q.~Zhang, J.~Zhang, and D.~Tao, ``{ViTAE}: Vision transformer advanced by exploring intrinsic inductive bias,'' in \emph{NeurIPS}, vol.~34, 2021, pp. 28\,522--28\,535.

\bibitem{vitae_v2}
Q.~Zhang, Y.~Xu, J.~Zhang, and D.~Tao, ``{ViTAEv2}: Vision transformer advanced by exploring inductive bias for image recognition and beyond,'' \emph{International Journal of Computer Vision}, pp. 1--22, 2023.

\bibitem{vit_g}
X.~Zhai, A.~Kolesnikov, N.~Houlsby, and L.~Beyer, ``Scaling vision transformers,'' in \emph{CVPR}, June 2022, pp. 12\,104--12\,113.

\bibitem{swint}
Z.~Liu, Y.~Lin, Y.~Cao, H.~Hu, Y.~Wei, Z.~Zhang, S.~Lin, and B.~Guo, ``Swin transformer: Hierarchical vision transformer using shifted windows,'' in \emph{ICCV}, 2021, pp. 10\,012--10\,022.

\bibitem{vit}
A.~Dosovitskiy, L.~Beyer, A.~Kolesnikov, D.~Weissenborn, X.~Zhai, T.~Unterthiner, M.~Dehghani, M.~Minderer, G.~Heigold, S.~Gelly, J.~Uszkoreit, and N.~Houlsby, ``An image is worth 16x16 words: Transformers for image recognition at scale,'' in \emph{ICLR}, 2021.

\bibitem{ringmo}
X.~Sun, P.~Wang, W.~Lu, Z.~Zhu, X.~Lu, Q.~He, J.~Li, X.~Rong, Z.~Yang, H.~Chang, Q.~He, G.~Yang, R.~Wang, J.~Lu, and K.~Fu, ``{RingMo}: A remote sensing foundation model with masked image modeling,'' \emph{IEEE Transactions on Geoscience and Remote Sensing}, vol.~61, pp. 1--22, 2023.

\bibitem{rvsa}
D.~Wang, Q.~Zhang, Y.~Xu, J.~Zhang, B.~Du, D.~Tao, and L.~Zhang, ``Advancing plain vision transformer toward remote sensing foundation model,'' \emph{IEEE Transactions on Geoscience and Remote Sensing}, vol.~61, pp. 1--15, 2023.

\bibitem{scaling_law}
J.~Kaplan, S.~McCandlish, T.~Henighan, T.~B. Brown, B.~Chess, R.~Child, S.~Gray, A.~Radford, J.~Wu, and D.~Amodei, ``Scaling laws for neural language models,'' \emph{arXiv preprint arXiv:2001.08361}, 2020.

\bibitem{bfm}
K.~Cha, J.~Seo, and T.~Lee, ``A billion-scale foundation model for remote sensing images,'' \emph{IEEE Journal of Selected Topics in Applied Earth Observations and Remote Sensing}, pp. 1--17, 2024.

\bibitem{skysense}
X.~Guo, J.~Lao, B.~Dang, Y.~Zhang, L.~Yu, L.~Ru, L.~Zhong, Z.~Huang, K.~Wu, D.~Hu, H.~He, J.~Wang, J.~Chen, M.~Yang, Y.~Zhang, and Y.~Li, ``Skysense: A multi-modal remote sensing foundation model towards universal interpretation for earth observation imagery,'' in \emph{CVPR}, 2024, pp. 27\,672--27\,683.

\bibitem{swin_v2}
Z.~Liu, H.~Hu, Y.~Lin, Z.~Yao, Z.~Xie, Y.~Wei, J.~Ning, Y.~Cao, Z.~Zhang, L.~Dong, F.~Wei, and B.~Guo, ``{Swin Transformer V2}: Scaling up capacity and resolution,'' in \emph{CVPR}, June 2022, pp. 12\,009--12\,019.

\bibitem{internimage}
W.~Wang, J.~Dai, Z.~Chen, Z.~Huang, Z.~Li, X.~Zhu, X.~Hu, T.~Lu, L.~Lu, H.~Li \emph{et~al.}, ``{Internimage}: Exploring large-scale vision foundation models with deformable convolutions,'' in \emph{CVPR}, 2023, pp. 14\,408--14\,419.

\bibitem{rsp}
D.~Wang, J.~Zhang, B.~Du, G.-S. Xia, and D.~Tao, ``An empirical study of remote sensing pretraining,'' \emph{IEEE Transactions on Geoscience and Remote Sensing}, vol.~61, pp. 1--20, 2023.

\bibitem{mocov2}
X.~Chen, H.~Fan, R.~Girshick, and K.~He, ``Improved baselines with momentum contrastive learning,'' \emph{arXiv preprint arXiv:2003.04297}, 2020.

\bibitem{seco}
O.~Ma{\~n}as, A.~Lacoste, X.~Giro-i Nieto, D.~Vazquez, and P.~Rodriguez, ``Seasonal contrast: Unsupervised pre-training from uncurated remote sensing data,'' in \emph{ICCV}, 2021, pp. 9414--9423.

\bibitem{caco}
U.~Mall, B.~Hariharan, and K.~Bala, ``Change-aware sampling and contrastive learning for satellite images,'' in \emph{CVPR}, 2023, pp. 5261--5270.

\bibitem{geoclip}
V.~Vivanco~Cepeda, G.~K. Nayak, and M.~Shah, ``{GeoCLIP}: Clip-inspired alignment between locations and images for effective worldwide geo-localization,'' in \emph{NeurIPS}, vol.~36, 2023, pp. 8690--8701.

\bibitem{satmae}
Y.~Cong, S.~Khanna, C.~Meng, P.~Liu, E.~Rozi, Y.~He, M.~Burke, D.~Lobell, and S.~Ermon, ``{SatMAE}: Pre-training transformers for temporal and multi-spectral satellite imagery,'' in \emph{NeurIPS}, vol.~35, 2022, pp. 197--211.

\bibitem{smlfr}
Z.~Dong, Y.~Gu, and T.~Liu, ``Generative convnet foundation model with sparse modeling and low-frequency reconstruction for remote sensing image interpretation,'' \emph{IEEE Transactions on Geoscience and Remote Sensing}, vol.~62, pp. 1--16, 2024.

\bibitem{gfm}
M.~Mendieta, B.~Han, X.~Shi, Y.~Zhu, and C.~Chen, ``Towards geospatial foundation models via continual pretraining,'' in \emph{ICCV}, 2023, pp. 16\,806--16\,816.

\bibitem{simmim}
Z.~Xie, Z.~Zhang, Y.~Cao, Y.~Lin, J.~Bao, Z.~Yao, Q.~Dai, and H.~Hu, ``{SimMIM}: A simple framework for masked image modeling,'' in \emph{CVPR}, 2022, pp. 9653--9663.

\bibitem{mae}
K.~He, X.~Chen, S.~Xie, Y.~Li, P.~Doll\'ar, and R.~Girshick, ``Masked autoencoders are scalable vision learners,'' in \emph{CVPR}, 2022, pp. 16\,000--16\,009.

\bibitem{mtp}
D.~Wang, J.~Zhang, M.~Xu, L.~Liu, D.~Wang, E.~Gao, C.~Han, H.~Guo, B.~Du, D.~Tao, and L.~Zhang, ``{MTP}: Advancing remote sensing foundation model via multi-task pretraining,'' \emph{IEEE Journal of Selected Topics in Applied Earth Observations and Remote Sensing}, pp. 1--24, 2024.

\bibitem{upetu}
Z.~Dong, Y.~Gu, and T.~Liu, ``Upetu: A unified parameter-efficient fine-tuning framework for remote sensing foundation model,'' \emph{IEEE Transactions on Geoscience and Remote Sensing}, vol.~62, pp. 1--13, 2024.

\bibitem{vitdet}
Y.~Li, H.~Mao, R.~Girshick, and K.~He, ``Exploring plain vision transformer backbones for object detection,'' in \emph{ECCV}, 2022, pp. 280--296.

\bibitem{ss-mae}
J.~Lin, F.~Gao, X.~Shi, J.~Dong, and Q.~Du, ``{SS-MAE}: Spatial–spectral masked autoencoder for multisource remote sensing image classification,'' \emph{IEEE Transactions on Geoscience and Remote Sensing}, vol.~61, pp. 1--14, 2023.

\bibitem{layernorm}
J.~{Lei Ba}, J.~R. {Kiros}, and G.~E. {Hinton}, ``Layer normalization,'' \emph{arXiv e-prints}, p. arXiv:1607.06450, 2016.

\bibitem{nlsa}
Y.~Mei, Y.~Fan, and Y.~Zhou, ``Image super-resolution with non-local sparse attention,'' in \emph{CVPR}, 2021, pp. 3517--3526.

\bibitem{dat}
Z.~Xia, X.~Pan, S.~Song, L.~E. Li, and G.~Huang, ``Vision transformer with deformable attention,'' in \emph{CVPR}, 2022, pp. 4794--4803.

\bibitem{deformable_detr}
X.~Zhu, W.~Su, L.~Lu, B.~Li, X.~Wang, and J.~Dai, ``{Deformable DETR}: Deformable transformers for end-to-end object detection,'' in \emph{ICLR}, 2021.

\bibitem{indian_pines}
M.~F. Baumgardner, L.~L. Biehl, and D.~A. Landgrebe, ``220 band aviris hyperspectral image data set: June 12, 1992 indian pine test site 3,'' \emph{Purdue University Research Repository}, vol.~10, no.~7, p. 991, 2015.

\bibitem{whu_hi1}
Y.~Zhong, X.~Hu, C.~Luo, X.~Wang, J.~Zhao, and L.~Zhang, ``{WHU-Hi}: Uav-borne hyperspectral with high spatial resolution (h2) benchmark datasets and classifier for precise crop identification based on deep convolutional neural network with crf,'' \emph{Remote Sensing of Environment}, vol. 250, p. 112012, 2020.

\bibitem{houston}
C.~Debes, A.~Merentitis, R.~Heremans, J.~Hahn, N.~Frangiadakis, T.~van Kasteren, W.~Liao, R.~Bellens, A.~Pižurica, S.~Gautama, W.~Philips, S.~Prasad, Q.~Du, and F.~Pacifici, ``Hyperspectral and lidar data fusion: Outcome of the 2013 grss data fusion contest,'' \emph{IEEE Journal of Selected Topics in Applied Earth Observations and Remote Sensing}, vol.~7, no.~6, pp. 2405--2418, 2014.

\bibitem{huanghe}
W.~Sun, K.~Liu, G.~Ren, W.~Liu, G.~Yang, X.~Meng, and J.~Peng, ``A simple and effective spectral-spatial method for mapping large-scale coastal wetlands using china zy1-02d satellite hyperspectral images,'' \emph{International Journal of Applied Earth Observation and Geoinformation}, vol. 104, p. 102572, 2021.

\bibitem{fullycontnet}
D.~Wang, B.~Du, and L.~Zhang, ``Fully contextual network for hyperspectral scene parsing,'' \emph{IEEE Transactions on Geoscience and Remote Sensing}, vol.~60, pp. 1--16, 2022.

\bibitem{MSDN}
C.~Zhang, G.~Li, and S.~Du, ``Multi-scale dense networks for hyperspectral remote sensing image classification,'' \emph{IEEE Transactions on Geoscience and Remote Sensing}, vol.~57, no.~11, pp. 9201--9222, 2019.

\bibitem{ssfcn}
Y.~{Xu}, B.~{Du}, and L.~{Zhang}, ``Beyond the patchwise classification: Spectral-spatial fully convolutional networks for hyperpsectral image classificaiton,'' \emph{IEEE Transactions on Big Data}, pp. 1--1, 2019.

\bibitem{HSIC-FM}
J.~Yang, B.~Du, and L.~Zhang, ``Overcoming the barrier of incompleteness: A hyperspectral image classification full model,'' \emph{IEEE Transactions on Neural Networks and Learning Systems}, 2023.

\bibitem{ssgrn}
D.~Wang, B.~Du, and L.~Zhang, ``Spectral-spatial global graph reasoning for hyperspectral image classification,'' \emph{IEEE Transactions on Neural Networks and Learning Systems}, pp. 1--14, 2023.

\bibitem{yang_csil}
J.~Yang, B.~Du, and L.~Zhang, ``From center to surrounding: An interactive learning framework for hyperspectral image classification,'' \emph{ISPRS Journal of Photogrammetry and Remote Sensing}, vol. 197, pp. 145--166, 2023.

\bibitem{idcn}
X.~Li, M.~Ding, Y.~Gu, and A.~Pi{\v{z}}urica, ``An end-to-end framework for joint denoising and classification of hyperspectral images,'' \emph{IEEE Transactions on Neural Networks and Learning Systems}, vol.~34, no.~7, pp. 3269--3283, 2023.

\bibitem{CLOLN}
C.~Li, B.~Rasti, X.~Tang, P.~Duan, J.~Li, and Y.~Peng, ``Channel-layer-oriented lightweight spectral-spatial network for hyperspectral image classification,'' \emph{IEEE Transactions on Geoscience and Remote Sensing}, 2024.

\bibitem{kipfgcn}
T.~N. Kipf and M.~Welling, ``Semi-supervised classification with graph convolutional networks,'' in \emph{ICLR}, 2017.

\bibitem{htd_mosaic}
A.~Giannandrea, N.~Raqueno, D.~W. Messinger, J.~Faulring, J.~P. Kerekes, J.~Van~Aardt, K.~Canham, S.~Hagstrom, E.~Ontiveros, A.~Gerace \emph{et~al.}, ``The share 2012 data campaign,'' in \emph{Algorithms and Technologies for Multispectral, Hyperspectral, and Ultraspectral Imagery XIX}, vol. 8743, 2013, pp. 94--108.

\bibitem{htd_aviris}
L.~Zhang, L.~Zhang, D.~Tao, and X.~Huang, ``Sparse transfer manifold embedding for hyperspectral target detection,'' \emph{IEEE Transactions on Geoscience and Remote Sensing}, vol.~52, no.~2, pp. 1030--1043, 2013.

\bibitem{htd_renourban}
D.~Zhu, P.~Zhong, B.~Du, and L.~Zhang, ``Attention-based sparse and collaborative spectral abundance learning for hyperspectral subpixel target detection,'' \emph{Neural Networks}, vol. 178, p. 106416, 2024.

\bibitem{had_pavia}
N.~Billor, A.~S. Hadi, and P.~F. Velleman, ``Bacon: blocked adaptive computationally efficient outlier nominators,'' \emph{Computational statistics \& data analysis}, vol.~34, no.~3, pp. 279--298, 2000.

\bibitem{had_cri}
T.~Zhou, D.~Tao, and X.~Wu, ``Manifold elastic net: a unified framework for sparse dimension reduction,'' \emph{Data Mining and Knowledge Discovery}, vol.~22, pp. 340--371, 2011.

\bibitem{had_viareggio}
S.~Sun, J.~Liu, and S.~Sun, ``Hyperspectral subpixel target detection based on interaction subspace model,'' \emph{Pattern Recognition}, vol. 139, p. 109464, 2023.

\bibitem{htd_vit}
H.~Qin, W.~Xie, Y.~Li, and Q.~Du, ``Htd-vit: Spectral-spatial joint hyperspectral target detection with vision transformer,'' in \emph{IGARSS}, 2022, pp. 1967--1970.

\bibitem{sam}
A.~Kirillov, E.~Mintun, N.~Ravi, H.~Mao, C.~Rolland, L.~Gustafson, T.~Xiao, S.~Whitehead, A.~C. Berg, W.-Y. Lo, P.~Dollar, and R.~Girshick, ``Segment anything,'' in \emph{ICCV}, 2023, pp. 4015--4026.

\bibitem{cem}
C.-I. Chang, J.~Liu, B.~Chieu, H.~Ren, C.-M. Wang, C.~Lo, P.-C. Chung, C.-W. Yang, and D.~Ma, ``Generalized constrained energy minimization approach to subpixel target detection for multispectral imagery,'' \emph{Optical Engineering}, vol.~39, no.~5, pp. 1275--1281, 2000.

\bibitem{htd_std}
Y.~Chen, N.~M. Nasrabadi, and T.~D. Tran, ``Sparse representation for target detection in hyperspectral imagery,'' \emph{IEEE Journal of Selected Topics in Signal Processing}, vol.~5, no.~3, pp. 629--640, 2011.

\bibitem{cscr}
W.~Li and Q.~Du, ``Collaborative representation for hyperspectral anomaly detection,'' \emph{IEEE Transactions on geoscience and remote sensing}, vol.~53, no.~3, pp. 1463--1474, 2014.

\bibitem{srbbh}
Y.~Zhang, B.~Du, and L.~Zhang, ``A sparse representation-based binary hypothesis model for target detection in hyperspectral images,'' \emph{IEEE Transactions on Geoscience and Remote Sensing}, vol.~53, no.~3, pp. 1346--1354, 2014.

\bibitem{htdnet}
G.~Zhang, S.~Zhao, W.~Li, Q.~Du, Q.~Ran, and R.~Tao, ``{HTD-Net}: A deep convolutional neural network for target detection in hyperspectral imagery,'' \emph{Remote Sensing}, vol.~12, no.~9, p. 1489, 2020.

\bibitem{htdirn}
D.~Shen, X.~Ma, W.~Kong, J.~Liu, J.~Wang, and H.~Wang, ``Hyperspectral target detection based on interpretable representation network,'' \emph{IEEE Transactions on Geoscience and Remote Sensing}, vol.~61, pp. 1--16, 2023.

\bibitem{cgsal}
D.~Zhu, B.~Du, M.~Hu, Y.~Dong, and L.~Zhang, ``Collaborative-guided spectral abundance learning with bilinear mixing model for hyperspectral subpixel target detection,'' \emph{Neural Networks}, vol. 163, pp. 205--218, 2023.

\bibitem{rx}
I.~S. Reed and X.~Yu, ``Adaptive multiple-band cfar detection of an optical pattern with unknown spectral distribution,'' \emph{IEEE transactions on acoustics, speech, and signal processing}, vol.~38, no.~10, pp. 1760--1770, 1990.

\bibitem{kifd}
S.~Li, K.~Zhang, P.~Duan, and X.~Kang, ``Hyperspectral anomaly detection with kernel isolation forest,'' \emph{IEEE Transactions on Geoscience and Remote Sensing}, vol.~58, no.~1, pp. 319--329, 2019.

\bibitem{crd}
W.~Li and Q.~Du, ``Collaborative representation for hyperspectral anomaly detection,'' \emph{IEEE Transactions on geoscience and remote sensing}, vol.~53, no.~3, pp. 1463--1474, 2014.

\bibitem{gtvlrr}
T.~Cheng and B.~Wang, ``Graph and total variation regularized low-rank representation for hyperspectral anomaly detection,'' \emph{IEEE Transactions on Geoscience and Remote Sensing}, vol.~58, no.~1, pp. 391--406, 2019.

\bibitem{autoad}
S.~Wang, X.~Wang, L.~Zhang, and Y.~Zhong, ``{Auto-AD}: Autonomous hyperspectral anomaly detection network based on fully convolutional autoencoder,'' \emph{IEEE Transactions on Geoscience and Remote Sensing}, vol.~60, pp. 1--14, 2021.

\bibitem{rgae}
G.~Fan, Y.~Ma, X.~Mei, F.~Fan, J.~Huang, and J.~Ma, ``Hyperspectral anomaly detection with robust graph autoencoders,'' \emph{IEEE Transactions on Geoscience and Remote Sensing}, vol.~60, pp. 1--14, 2021.

\bibitem{gthad}
J.~Lian, L.~Wang, H.~Sun, and H.~Huang, ``{GT-HAD}: Gated transformer for hyperspectral anomaly detection,'' \emph{IEEE Transactions on Neural Networks and Learning Systems}, pp. 1--15, 2024.

\bibitem{hanet}
C.~Han, C.~Wu, H.~Guo, M.~Hu, and H.~Chen, ``{HANet}: A hierarchical attention network for change detection with bitemporal very-high-resolution remote sensing images,'' \emph{IEEE Journal of Selected Topics in Applied Earth Observations and Remote Sensing}, vol.~16, pp. 3867--3878, 2023.

\bibitem{c2fnet}
C.~Han, C.~Wu, M.~Hu, J.~Li, and H.~Chen, ``{C2F-SemiCD}: A coarse-to-fine semi-supervised change detection method based on consistency regularization in high-resolution remote-sensing images,'' \emph{IEEE Transactions on Geoscience and Remote Sensing}, pp. 1--1, 2024.

\bibitem{hsi_cd_sstformer}
Y.~Wang, D.~Hong, J.~Sha, L.~Gao, L.~Liu, Y.~Zhang, and X.~Rong, ``Spectral–spatial–temporal transformers for hyperspectral image change detection,'' \emph{IEEE Transactions on Geoscience and Remote Sensing}, vol.~60, pp. 1--14, 2022.

\bibitem{hsi_cd_cva}
R.~D. Johnson and E.~S. Kasischke, ``Change vector analysis: A technique for the multispectral monitoring of land cover and condition,'' \emph{International Journal of Remote Sensing}, vol.~19, no.~3, pp. 411--426, 1998.

\bibitem{hsi_cd_isfa}
C.~Wu, B.~Du, and L.~Zhang, ``Slow feature analysis for change detection in multispectral imagery,'' \emph{IEEE Transactions on Geoscience and Remote Sensing}, vol.~52, no.~5, pp. 2858--2874, 2014.

\bibitem{hsi_cd_getnet}
Q.~Wang, Z.~Yuan, Q.~Du, and X.~Li, ``{GETNET}: A general end-to-end 2-d cnn framework for hyperspectral image change detection,'' \emph{IEEE Transactions on Geoscience and Remote Sensing}, vol.~57, no.~1, pp. 3--13, 2019.

\bibitem{hsi_cd_mledan}
J.~Qu, S.~Hou, W.~Dong, Y.~Li, and W.~Xie, ``A multilevel encoder–decoder attention network for change detection in hyperspectral images,'' \emph{IEEE Transactions on Geoscience and Remote Sensing}, vol.~60, pp. 1--13, 2022.

\bibitem{hsi_cd_bit}
H.~Chen, Z.~Qi, and Z.~Shi, ``Remote sensing image change detection with transformers,'' \emph{IEEE Transactions on Geoscience and Remote Sensing}, vol.~60, pp. 1--14, 2022.

\bibitem{hsi_cd_emsnet}
M.~Hu, C.~Wu, and B.~Du, ``{EMS-NET}: Efficient multi-temporal self-attention for hyperspectral change detection,'' in \emph{IGARSS}, 2023, pp. 6664--6667.

\bibitem{hsi_cd_csanet}
R.~Song, W.~Ni, W.~Cheng, and X.~Wang, ``{CSANet}: Cross-temporal interaction symmetric attention network for hyperspectral image change detection,'' \emph{IEEE Geoscience and Remote Sensing Letters}, vol.~19, pp. 1--5, 2022.

\bibitem{hsi_cd_globalmind}
M.~Hu, C.~Wu, and L.~Zhang, ``Globalmind: Global multi-head interactive self-attention network for hyperspectral change detection,'' \emph{ISPRS Journal of Photogrammetry and Remote Sensing}, vol. 211, pp. 465--483, 2024.

\bibitem{bayesian_unmixing}
A.~Lagrange, M.~Fauvel, S.~May, and N.~Dobigeon, ``A bayesian model for joint unmixing and robust classification of hyperspectral images,'' in \emph{ICASSP}, 2018, pp. 3399--3403.

\bibitem{spectral_mixing}
N.~Keshava and J.~Mustard, ``Spectral unmixing,'' \emph{IEEE Signal Processing Magazine}, vol.~19, no.~1, pp. 44--57, 2002.

\bibitem{hyperspectral_unmixing}
J.~M. Bioucas-Dias, A.~Plaza, N.~Dobigeon, M.~Parente, Q.~Du, P.~Gader, and J.~Chanussot, ``Hyperspectral unmixing overview: Geometrical, statistical, and sparse regression-based approaches,'' \emph{IEEE journal of selected topics in applied earth observations and remote sensing}, vol.~5, no.~2, pp. 354--379, 2012.

\bibitem{ghamisi2017advances}
P.~Ghamisi, N.~Yokoya, J.~Li, W.~Liao, S.~Liu, J.~Plaza, B.~Rasti, and A.~Plaza, ``Advances in hyperspectral image and signal processing: A comprehensive overview of the state of the art,'' \emph{IEEE Geoscience and Remote Sensing Magazine}, vol.~5, no.~4, pp. 37--78, 2017.

\bibitem{fcls}
D.~C. Heinz \emph{et~al.}, ``Fully constrained least squares linear spectral mixture analysis method for material quantification in hyperspectral imagery,'' \emph{IEEE transactions on geoscience and remote sensing}, vol.~39, no.~3, pp. 529--545, 2001.

\bibitem{elmm}
M.~A. Veganzones, L.~Drumetz, G.~Tochon, M.~Dalla~Mura, A.~Plaza, J.~Bioucas-Dias, and J.~Chanussot, ``A new extended linear mixing model to address spectral variability,'' in \emph{WHISPERS}, 2014, pp. 1--4.

\bibitem{sunsal}
U.~Kumar, C.~Milesi, R.~R. Nemani, S.~Kumar~Raja, S.~Ganguly, and W.~Wang, ``Sparse unmixing via variable splitting and augmented lagrangian for vegetation and urban area classification using landsat data,'' \emph{The International Archives of the Photogrammetry, Remote Sensing and Spatial Information Sciences}, vol.~40, pp. 59--65, 2015.

\bibitem{cnnaeu}
B.~Palsson, M.~O. Ulfarsson, and J.~R. Sveinsson, ``Convolutional autoencoder for spectral--spatial hyperspectral unmixing,'' \emph{IEEE Transactions on Geoscience and Remote Sensing}, vol.~59, no.~1, pp. 535--549, 2020.

\bibitem{cycu}
L.~Gao, Z.~Han, D.~Hong, B.~Zhang, and J.~Chanussot, ``{CyCU-Net}: Cycle-consistency unmixing network by learning cascaded autoencoders,'' \emph{IEEE Transactions on Geoscience and Remote Sensing}, vol.~60, pp. 1--14, 2021.

\bibitem{deeptrans}
P.~Ghosh, S.~K. Roy, B.~Koirala, B.~Rasti, and P.~Scheunders, ``Hyperspectral unmixing using transformer network,'' \emph{IEEE Transactions on Geoscience and Remote Sensing}, vol.~60, pp. 1--16, 2022.

\bibitem{egunet}
D.~Hong, L.~Gao, J.~Yao, N.~Yokoya, J.~Chanussot, U.~Heiden, and B.~Zhang, ``Endmember-guided unmixing network (egu-net): A general deep learning framework for self-supervised hyperspectral unmixing,'' \emph{IEEE Transactions on Neural Networks and Learning Systems}, vol.~33, no.~11, pp. 6518--6531, 2021.

\bibitem{DBLP:journals/tip/ChenWGWYL23}
Q.~Chen, Y.~Wang, Z.~Geng, Y.~Wang, J.~Yang, and Z.~Lin, ``Equilibrium image denoising with implicit differentiation,'' \emph{IEEE Transactions on Image Processing}, vol.~32, pp. 1868--1881, 2023.

\bibitem{DBLP:journals/corr/abs-2306-13653}
J.~Ma, T.~Cheng, G.~Wang, Q.~Zhang, X.~Wang, and L.~Zhang, ``Prores: Exploring degradation-aware visual prompt for universal image restoration,'' \emph{CoRR}, vol. abs/2306.13653, 2023.

\bibitem{DBLP:journals/nn/MaWZZ23}
J.~Ma, G.~Wang, L.~Zhang, and Q.~Zhang, ``Restoration and enhancement on low exposure raw images by joint demosaicing and denoising,'' \emph{Neural Networks}, vol. 162, pp. 557--570, 2023.

\bibitem{DBLP:conf/mm/0002YZW022}
J.~Ma, S.~Yan, L.~Zhang, G.~Wang, and Q.~Zhang, ``Elmformer: Efficient raw image restoration with a locally multiplicative transformer,'' in \emph{ACM MM}, 2022, pp. 5842--5852.

\bibitem{DBLP:journals/tgrs/KorosovDMFP22}
A.~A. Korosov, D.~Demchev, N.~Miranda, N.~Franceschi, and J.~Park, ``Thermal denoising of cross-polarized sentinel-1 data in interferometric and extra wide swath modes,'' \emph{IEEE Transactions on Geoscience and Remote Sensing}, vol.~60, pp. 1--11, 2022.

\bibitem{DBLP:journals/lgrs/MiaoZZ22}
X.~Miao, Y.~Zhang, and J.~Zhang, ``Thermal hyperspectral image denoising using total variation based on bidirectional estimation and brightness temperature smoothing,'' \emph{IEEE Geoscience and Remote Sensing Letters}, vol.~19, pp. 1--5, 2022.

\bibitem{DBLP:conf/cvpr/ChangYZ17}
Y.~Chang, L.~Yan, and S.~Zhong, ``Hyper-laplacian regularized unidirectional low-rank tensor recovery for multispectral image denoising,'' in \emph{CVPR}, 2017, pp. 5901--5909.

\bibitem{DBLP:conf/cvpr/0003YLYZ19}
W.~He, Q.~Yao, C.~Li, N.~Yokoya, and Q.~Zhao, ``Non-local meets global: An integrated paradigm for hyperspectral denoising,'' in \emph{CVPR}, 2019, pp. 6868--6877.

\bibitem{xiong2019hyperspectral}
F.~Xiong, J.~Zhou, and Y.~Qian, ``Hyperspectral restoration via $l_0$ gradient regularized low-rank tensor factorization,'' \emph{IEEE Transactions on Geoscience and Remote Sensing}, vol.~57, no.~12, pp. 10\,410--10\,425, 2019.

\bibitem{peng2020enhanced}
J.~Peng, Q.~Xie, Q.~Zhao, Y.~Wang, L.~Yee, and D.~Meng, ``Enhanced 3dtv regularization and its applications on hsi denoising and compressed sensing,'' \emph{IEEE Transactions on Image Processing}, vol.~29, pp. 7889--7903, 2020.

\bibitem{miao2022hyperspectral}
Y.-C. Miao, X.-L. Zhao, X.~Fu, J.-L. Wang, and Y.-B. Zheng, ``Hyperspectral denoising using unsupervised disentangled spatiospectral deep priors.'' \emph{IEEE Transactions on Geoscience and Remote Sensing}, vol.~60, pp. 1--16, 2022.

\bibitem{DBLP:journals/tnn/WeiFH21}
K.~Wei, Y.~Fu, and H.~Huang, ``3-d quasi-recurrent neural network for hyperspectral image denoising,'' \emph{IEEE Transactions on Neural Networks and Learning Systems}, vol.~32, no.~1, pp. 363--375, 2021.

\bibitem{DBLP:conf/aaai/LiFZ23}
M.~Li, Y.~Fu, and Y.~Zhang, ``Spatial-spectral transformer for hyperspectral image denoising,'' in \emph{AAAI}, 2023, pp. 1368--1376.

\bibitem{li2018single}
Y.~Li, L.~Zhang, C.~Dingl, W.~Wei, and Y.~Zhang, ``Single hyperspectral image super-resolution with grouped deep recursive residual network,'' in \emph{BigMM}, 2018, pp. 1--4.

\bibitem{liang2021swinir}
J.~Liang, J.~Cao, G.~Sun, K.~Zhang, L.~Van~Gool, and R.~Timofte, ``Swinir: Image restoration using swin transformer,'' in \emph{CVPR}, 2021, pp. 1833--1844.

\bibitem{jiang2020learning}
J.~Jiang, H.~Sun, X.~Liu, and J.~Ma, ``Learning spatial-spectral prior for super-resolution of hyperspectral imagery,'' \emph{IEEE Transactions on Computational Imaging}, vol.~6, pp. 1082--1096, 2020.

\bibitem{wang2021hyperspectral}
X.~Wang, J.~Ma, and J.~Jiang, ``Hyperspectral image super-resolution via recurrent feedback embedding and spatial--spectral consistency regularization,'' \emph{IEEE Transactions on Geoscience and Remote Sensing}, vol.~60, pp. 1--13, 2021.

\bibitem{wang2022group}
X.~Wang, Q.~Hu, J.~Jiang, and J.~Ma, ``A group-based embedding learning and integration network for hyperspectral image super-resolution,'' \emph{IEEE Transactions on Geoscience and Remote Sensing}, vol.~60, pp. 1--16, 2022.

\bibitem{msdformer}
S.~Chen, L.~Zhang, and L.~Zhang, ``{MSDformer}: Multiscale deformable transformer for hyperspectral image super-resolution,'' \emph{IEEE Transactions on Geoscience and Remote Sensing}, vol.~61, pp. 1--14, 2023.

\bibitem{hybridsn}
S.~K. Roy, G.~Krishna, S.~R. Dubey, and B.~B. Chaudhuri, ``{HybridSN}: Exploring 3-d–2-d cnn feature hierarchy for hyperspectral image classification,'' \emph{IEEE Geoscience and Remote Sensing Letters}, vol.~17, no.~2, pp. 277--281, 2020.

\bibitem{a2s2kresnet}
S.~K. Roy, S.~Manna, T.~Song, and L.~Bruzzone, ``Attention-based adaptive spectral–spatial kernel resnet for hyperspectral image classification,'' \emph{IEEE Transactions on Geoscience and Remote Sensing}, vol.~59, no.~9, pp. 7831--7843, 2021.

\bibitem{xiongan}
C.~Yi, L.~Zhang, X.~Zhang, W.~Yueming, Q.~Wenchao, T.~Senlin, and P.~Zhang, ``Aerial hyperspectral remote sensing classification dataset of xiongan new area (matiwan village),'' \emph{National Remote Sensing Bulletin}, vol.~24, no.~11, pp. 1299--1306, 2020.

\bibitem{ITER}
J.~Yang, B.~Du, D.~Wang, and L.~Zhang, ``{ITER}: Image-to-pixel representation for weakly supervised hsi classification,'' \emph{IEEE Transactions on Image Processing}, vol.~33, pp. 257--272, 2024.

\bibitem{SACNet}
Y.~Xu, B.~Du, and L.~Zhang, ``Self-attention context network: Addressing the threat of adversarial attacks for hyperspectral image classification,'' \emph{IEEE Transactions on Image Processing}, vol.~30, pp. 8671--8685, 2021.

\bibitem{hsi_aa}
C.~Shi, Y.~Dang, L.~Fang, Z.~Lv, and M.~Zhao, ``Hyperspectral image classification with adversarial attack,'' \emph{IEEE Geoscience and Remote Sensing Letters}, vol.~19, pp. 1--5, 2022.

\bibitem{RCCA}
B.~Tu, W.~He, Q.~Li, Y.~Peng, and A.~Plaza, ``A new context-aware framework for defending against adversarial attacks in hyperspectral image classification,'' \emph{IEEE Transactions on Geoscience and Remote Sensing}, vol.~61, pp. 1--14, 2023.

\bibitem{S3ANet}
Y.~Xu, Y.~Xu, H.~Jiao, Z.~Gao, and L.~Zhang, ``{S³ANet}: Spatial–spectral self-attention learning network for defending against adversarial attacks in hyperspectral image classification,'' \emph{IEEE Transactions on Geoscience and Remote Sensing}, vol.~62, pp. 1--13, 2024.

\bibitem{FGSM}
A.~Kurakin, I.~J. Goodfellow, and S.~Bengio, ``Adversarial machine learning at scale,'' in \emph{ICLR}, 2017.

\bibitem{PGD}
A.~Madry, A.~Makelov, L.~Schmidt, D.~Tsipras, and A.~Vladu, ``Towards deep learning models resistant to adversarial attacks,'' in \emph{ICLR}, 2018.

\bibitem{JPEG}
G.~K. Wallace, ``The jpeg still picture compression standard,'' \emph{Communications of the ACM}, vol.~34, no.~4, pp. 30--44, 1991.

\bibitem{cgcdl}
T.~Liu, Y.~Gu, and X.~Jia, ``Class-guided coupled dictionary learning for multispectral-hyperspectral remote sensing image collaborative classification,'' \emph{Science China Technological Sciences}, vol.~65, no.~4, pp. 744--758, 2022.

\bibitem{cdls}
Y.-H.~H. Tsai, Y.-R. Yeh, and Y.-C.~F. Wang, ``Learning cross-domain landmarks for heterogeneous domain adaptation,'' in \emph{CVPR}, 2016, pp. 5081--5090.

\bibitem{cospace1}
D.~Hong, N.~Yokoya, J.~Chanussot, and X.~X. Zhu, ``Cospace: Common subspace learning from hyperspectral-multispectral correspondences,'' \emph{IEEE Transactions on Geoscience and Remote Sensing}, vol.~57, no.~7, pp. 4349--4359, 2019.

\bibitem{cospace2}
D.~Hong, J.~Chanussot, N.~Yokoya, J.~Kang, and X.~X. Zhu, ``Learning-shared cross-modality representation using multispectral-lidar and hyperspectral data,'' \emph{IEEE Geoscience and Remote Sensing Letters}, vol.~17, no.~8, pp. 1470--1474, 2020.

\bibitem{jc}
B.~Guo, T.~Liu, and Y.~Gu, ``Integrating coupled dictionary learning and distance preserved probability distribution adaptation for multispectral--hyperspectral image collaborative classification,'' \emph{IEEE Transactions on Geoscience and Remote Sensing}, vol.~60, pp. 1--13, 2022.

\bibitem{spdda}
------, ``Structure preserved discriminative distribution adaptation for multi-hyperspectral image collaborative classification,'' \emph{IEEE Transactions on Geoscience and Remote Sensing}, 2023.

\bibitem{oscd}
R.~C. Daudt, B.~Le~Saux, A.~Boulch, and Y.~Gousseau, ``Urban change detection for multispectral earth observation using convolutional neural networks,'' in \emph{IGARSS}, 2018, pp. 2115--2118.

\bibitem{Chen2023Exchange}
H.~Chen, J.~Song, C.~Wu, B.~Du, and N.~Yokoya, ``Exchange means change: An unsupervised single-temporal change detection framework based on intra- and inter-image patch exchange,'' \emph{ISPRS Journal of Photogrammetry and Remote Sensing}, vol. 206, pp. 87--105, 2023.

\bibitem{CayeDaudt2018Change}
R.~{Caye Daudt}, B.~{Le Saux}, and A.~Boulch, ``{Fully convolutional siamese networks for change detection},'' in \emph{ICIP}, 2018, pp. 4063--4067.

\bibitem{Chen2019Change}
H.~Chen, C.~Wu, B.~Du, L.~Zhang, and L.~Wang, ``{Change Detection in Multisource VHR Images via Deep Siamese Convolutional Multiple-Layers Recurrent Neural Network},'' \emph{IEEE Transactions on Geoscience and Remote Sensing}, vol.~58, no.~4, pp. 2848--2864, 2020.

\bibitem{Fang2022SNUNet}
S.~{Fang}, K.~{Li}, J.~{Shao}, and Z.~{Li}, ``{SNUNet-CD}: A densely connected siamese network for change detection of vhr images,'' \emph{IEEE Geoscience and Remote Sensing Letters}, vol.~19, p. 3056416, 2022.

\bibitem{Zhang2020Change}
C.~Zhang, P.~Yue, D.~Tapete, L.~Jiang, B.~Shangguan, L.~Huang, and G.~Liu, ``{A deeply supervised image fusion network for change detection in high resolution bi-temporal remote sensing images},'' \emph{ISPRS Journal of Photogrammetry and Remote Sensing}, vol. 166, pp. 183--200, 2020.

\bibitem{Bandara2022Transformer}
W.~G.~C. Bandara and V.~M. Patel, ``A transformer-based siamese network for change detection,'' in \emph{IGARSS}, 2022, pp. 207--210.

\bibitem{chen2024changemamba}
H.~Chen, J.~Song, C.~Han, J.~Xia, and N.~Yokoya, ``Changemamba: Remote sensing change detection with spatio-temporal state space model,'' \emph{arXiv preprint arXiv:2404.03425}, 2024.

\bibitem{mexico_gulf_oil}
P.~Duan, X.~Kang, P.~Ghamisi, and S.~Li, ``Hyperspectral remote sensing benchmark database for oil spill detection with an isolation forest-guided unsupervised detector,'' \emph{IEEE Transactions on Geoscience and Remote Sensing}, vol.~61, pp. 1--11, 2023.

\bibitem{scale_mae}
C.~J. Reed, R.~Gupta, S.~Li, S.~Brockman, C.~Funk, B.~Clipp, K.~Keutzer, S.~Candido, M.~Uyttendaele, and T.~Darrell, ``{Scale-MAE}: A scale-aware masked autoencoder for multiscale geospatial representation learning,'' in \emph{ICCV}, 2023, pp. 4088--4099.

\bibitem{dofa}
Z.~Xiong, Y.~Wang, F.~Zhang, A.~J. Stewart, J.~Hanna, D.~Borth, I.~Papoutsis, B.~L. Saux, G.~Camps-Valls, and X.~X. Zhu, ``Neural plasticity-inspired foundation model for observing the earth crossing modalities,'' \emph{arXiv preprint arXiv:2403.15356}, 2024.

\bibitem{vsa}
Q.~Zhang, Y.~Xu, J.~Zhang, and D.~Tao, ``{VSA}: Learning varied-size window attention in vision transformers,'' in \emph{ECCV}, 2022, pp. 466--483.

\bibitem{imagenet}
J.~Deng, W.~Dong, R.~Socher, L.-J. Li, K.~Li, and L.~Fei-Fei, ``{ImageNet}: A large-scale hierarchical image database,'' in \emph{CVPR}, 2009, pp. 248--255.

\bibitem{resnet}
K.~{He}, X.~{Zhang}, S.~{Ren}, and J.~{Sun}, ``Deep residual learning for image recognition,'' in \emph{CVPR}, 2016, pp. 770--778.

\bibitem{millionaid}
Y.~Long, G.-S. Xia, S.~Li, W.~Yang, M.~Y. Yang, X.~X. Zhu, L.~Zhang, and D.~Li, ``On creating benchmark dataset for aerial image interpretation: Reviews, guidances and million-aid,'' \emph{IEEE Journal of Selected Topics in Applied Earth Observations and Remote Sensing}, vol.~14, pp. 4205--4230, 2021.

\bibitem{gassl}
K.~Ayush, B.~Uzkent, C.~Meng, K.~Tanmay, M.~Burke, D.~Lobell, and S.~Ermon, ``Geography-aware self-supervised learning,'' in \emph{ICCV}, 2021, pp. 10\,181--10\,190.

\bibitem{fmow}
G.~Christie, N.~Fendley, J.~Wilson, and R.~Mukherjee, ``Functional map of the world,'' in \emph{CVPR}, 2018, pp. 6172--6180.

\bibitem{matter}
P.~Akiva, M.~Purri, and M.~Leotta, ``Self-supervised material and texture representation learning for remote sensing tasks,'' in \emph{CVPR}, 2022, pp. 8203--8215.

\bibitem{tov}
C.~Tao, J.~Qi, G.~Zhang, Q.~Zhu, W.~Lu, and H.~Li, ``{TOV}: The original vision model for optical remote sensing image understanding via self-supervised learning,'' \emph{IEEE Journal of Selected Topics in Applied Earth Observations and Remote Sensing}, vol.~16, pp. 4916--4930, 2023.

\bibitem{cspt}
T.~Zhang, P.~Gao, H.~Dong, Y.~Zhuang, G.~Wang, W.~Zhang, and H.~Chen, ``Consecutive pre-training: A knowledge transfer learning strategy with relevant unlabeled data for remote sensing domain,'' \emph{Remote Sensing}, vol.~14, no.~22, 2022.

\bibitem{cmid}
D.~Muhtar, X.~Zhang, P.~Xiao, Z.~Li, and F.~Gu, ``{CMID}: A unified self-supervised learning framework for remote sensing image understanding,'' \emph{IEEE Transactions on Geoscience and Remote Sensing}, vol.~61, pp. 1--17, 2023.

\bibitem{gersp}
Z.~Huang, M.~Zhang, Y.~Gong, Q.~Liu, and Y.~Wang, ``Generic knowledge boosted pretraining for remote sensing images,'' \emph{IEEE Transactions on Geoscience and Remote Sensing}, vol.~62, pp. 1--13, 2024.

\bibitem{croma}
A.~Fuller, K.~Millard, and J.~Green, ``{CROMA}: Remote sensing representations with contrastive radar-optical masked autoencoders,'' in \emph{NeurIPS}, vol.~36, 2023, pp. 5506--5538.

\bibitem{ss4leo_s12}
Y.~Wang, N.~A.~A. Braham, Z.~Xiong, C.~Liu, C.~M. Albrecht, and X.~X. Zhu, ``{SSL4EO-S12}: A large-scale multimodal, multitemporal dataset for self-supervised learning in earth observation [software and data sets],'' \emph{IEEE Geoscience and Remote Sensing Magazine}, vol.~11, no.~3, pp. 98--106, 2023.

\bibitem{cross-scale_mae}
M.~Tang, A.~Cozma, K.~Georgiou, and H.~Qi, ``{Cross-Scale MAE}: A tale of multiscale exploitation in remote sensing,'' in \emph{NeurIPS}, vol.~36, 2023, pp. 20\,054--20\,066.

\bibitem{ctxmim}
M.~Zhang, Q.~Liu, and Y.~Wang, ``{CtxMIM}: Context-enhanced masked image modeling for remote sensing image understanding,'' \emph{arXiv preprint arXiv:2310.00022}, 2023.

\bibitem{bigearthnet}
G.~Sumbul, M.~Charfuelan, B.~Demir, and V.~Markl, ``Bigearthnet: A large-scale benchmark archive for remote sensing image understanding,'' in \emph{IGARSS}, 2019, pp. 5901--5904.

\bibitem{fgmae}
Y.~Wang, H.~H. Hern{\'a}ndez, C.~M. Albrecht, and X.~X. Zhu, ``Feature guided masked autoencoder for self-supervised learning in remote sensing,'' \emph{arXiv preprint arXiv:2310.18653}, 2023.

\bibitem{msgfm}
B.~Han, S.~Zhang, X.~Shi, and M.~Reichstein, ``Bridging remote sensors with multisensor geospatial foundation models,'' in \emph{CVPR}, 2024, pp. 27\,852--27\,862.

\bibitem{lemevit}
W.~Jiang, J.~Zhang, D.~Wang, Q.~Zhang, Z.~Wang, and B.~Du, ``{LeMeViT}: Efficient vision transformer with learnable meta tokens for remote sensing image interpretation,'' in \emph{IJCAI}, 2024.

\bibitem{convnext}
Z.~Liu, H.~Mao, C.-Y. Wu, C.~Feichtenhofer, T.~Darrell, and S.~Xie, ``A convnet for the 2020s,'' in \emph{CVPR}, 2022, pp. 11\,976--11\,986.

\bibitem{samrs}
D.~Wang, J.~Zhang, B.~Du, M.~Xu, L.~Liu, D.~Tao, and L.~Zhang, ``{SAMRS}: Scaling-up remote sensing segmentation dataset with segment anything model,'' in \emph{NeurIPS}, vol.~36, 2023, pp. 8815--8827.

\bibitem{decur}
Y.~Wang, C.~M. Albrecht, N.~A.~A. Braham, C.~Liu, Z.~Xiong, and X.~X. Zhu, ``{DeCUR}: decoupling common \& unique representations for multimodal self-supervision,'' \emph{arXiv preprint arXiv:2309.05300}, 2023.

\bibitem{geonrw}
\BIBentryALTinterwordspacing
G.~Baier, A.~Deschemps, M.~Schmitt, and N.~Yokoya, ``Geonrw,'' 2020. [Online]. Available: \url{https://dx.doi.org/10.21227/s5xq-b822}
\BIBentrySTDinterwordspacing

\bibitem{sun-rgbd}
S.~Song, S.~P. Lichtenberg, and J.~Xiao, ``Sun rgb-d: A rgb-d scene understanding benchmark suite,'' in \emph{CVPR}, 2015, pp. 567--576.

\bibitem{softcon}
Y.~Wang, C.~M. Albrecht, and X.~X. Zhu, ``Multilabel-guided soft contrastive learning for efficient earth observation pretraining,'' \emph{IEEE Transactions on Geoscience and Remote Sensing}, vol.~62, pp. 1--16, 2024.

\bibitem{ma3e}
Z.~Li, B.~Hou, S.~Ma, Z.~Wu, X.~Guo, B.~Ren, and L.~Jiao, ``Masked angle-aware autoencoder for remote sensing images,'' in \emph{European Conference on Computer Vision}, 2025, pp. 260--278.

\bibitem{selectivemae}
F.~Wang, H.~Wang, D.~Wang, Z.~Guo, Z.~Zhong, L.~Lan, J.~Zhang, Z.~Liu, and M.~Sun, ``Scaling efficient masked image modeling on large remote sensing dataset,'' \emph{arXiv preprint arXiv:2406.11933}, 2024.

\bibitem{PCAinHSIclassify}
G.~{Licciardi}, P.~R. {Marpu}, J.~{Chanussot}, and J.~A. {Benediktsson}, ``Linear versus nonlinear pca for the classification of hyperspectral data based on the extended morphological profiles,'' \emph{IEEE Geoscience and Remote Sensing Letters}, vol.~9, no.~3, pp. 447--451, 2012.

\bibitem{ICAinHSIclassify}
A.~{Villa}, J.~A. {Benediktsson}, J.~{Chanussot}, and C.~{Jutten}, ``Hyperspectral image classification with independent component discriminant analysis,'' \emph{IEEE Transactions on Geoscience and Remote Sensing}, vol.~49, no.~12, pp. 4865--4876, 2011.

\bibitem{LDAinHSIclassify}
T.~V. {Bandos}, L.~{Bruzzone}, and G.~{Camps-Valls}, ``Classification of hyperspectral images with regularized linear discriminant analysis,'' \emph{IEEE Transactions on Geoscience and Remote Sensing}, vol.~47, no.~3, pp. 862--873, 2009.

\bibitem{KNNinHSIclassify}
B.~{Tu}, J.~{Wang}, X.~{Kang}, G.~{Zhang}, X.~{Ou}, and L.~{Guo}, ``{KNN}-based representation of superpixels for hyperspectral image classification,'' \emph{IEEE Journal of Selected Topics in Applied Earth Observations and Remote Sensing}, vol.~11, no.~11, pp. 4032--4047, 2018.

\bibitem{MLRinhsi}
J.~Li, J.~M. Bioucas-Dias, and A.~Plaza, ``Semisupervised hyperspectral image segmentation using multinomial logistic regression with active learning,'' \emph{IEEE Transactions on Geoscience and Remote Sensing}, vol.~48, no.~11, pp. 4085--4098, 2010.

\bibitem{rfinhsi}
J.~Ham, Y.~Chen, M.~Crawford, and J.~Ghosh, ``Investigation of the random forest framework for classification of hyperspectral data,'' \emph{IEEE Transactions on Geoscience and Remote Sensing}, vol.~43, no.~3, pp. 492--501, 2005.

\bibitem{svmforhsiclassify}
F.~{Melgani} and L.~{Bruzzone}, ``Classification of hyperspectral remote sensing images with support vector machines,'' \emph{IEEE Transactions on Geoscience and Remote Sensing}, vol.~42, no.~8, pp. 1778--1790, 2004.

\bibitem{MP_1}
J.~A. {Benediktsson}, J.~A. {Palmason}, and J.~R. {Sveinsson}, ``Classification of hyperspectral data from urban areas based on extended morphological profiles,'' \emph{IEEE Transactions on Geoscience and Remote Sensing}, vol.~43, no.~3, pp. 480--491, 2005.

\bibitem{Fang2015SS}
L.~{Fang}, S.~{Li}, X.~{Kang}, and J.~A. {Benediktsson}, ``Spectral-spatial classification of hyperspectral images with a superpixel-based discriminative sparse model,'' \emph{IEEE Transactions on Geoscience and Remote Sensing}, vol.~53, no.~8, pp. 4186--4201, 2015.

\bibitem{hsi_glcm2}
F.~Tsai, C.-K. Chang, J.-Y. Rau, T.-H. Lin, and G.-R. Liu, ``3d computation of gray level co-occurrence in hyperspectral image cubes,'' in \emph{EMMCVPR}, 2007, pp. 429--440.

\bibitem{mugnet}
B.~Pan, Z.~Shi, and X.~Xu, ``{MugNet}: Deep learning for hyperspectral image classification using limited samples,'' \emph{ISPRS Journal of Photogrammetry and Remote Sensing}, vol. 145, pp. 108--119, 2018.

\bibitem{representlearning}
Y.~{Bengio}, A.~{Courville}, and P.~{Vincent}, ``Representation learning: A review and new perspectives,'' \emph{IEEE Transactions on Pattern Analysis and Machine Intelligence}, vol.~35, no.~8, pp. 1798--1828, 2013.

\bibitem{1dcnn}
W.~Hu, Y.~Huang, L.~Wei, F.~Zhang, and H.~Li, ``Deep convolutional neural networks for hyperspectral image classification,'' \emph{Journal of Sensors}, vol. 2015, pp. 1--12, 2015.

\bibitem{deeplab_hsi}
Z.~{Niu}, W.~{Liu}, J.~{Zhao}, and G.~{Jiang}, ``Deeplab-based spatial feature extraction for hyperspectral image classification,'' \emph{IEEE Geoscience and Remote Sensing Letters}, vol.~16, no.~2, pp. 251--255, 2019.

\bibitem{ssrn}
Z.~Zhong, J.~Li, Z.~Luo, and M.~Chapman, ``Spectral–spatial residual network for hyperspectral image classification: A 3-d deep learning framework,'' \emph{IEEE Transactions on Geoscience and Remote Sensing}, vol.~56, no.~2, pp. 847--858, 2018.

\bibitem{fu2024resc}
C.~Fu, B.~Du, and L.~Zhang, ``{ReSC-net}: Hyperspectral image classification based on attention-enhanced residual module and spatial-channel attention,'' \emph{IEEE Transactions on Geoscience and Remote Sensing}, 2024.

\bibitem{miao2023dds2m}
Y.~Miao, L.~Zhang, L.~Zhang, and D.~Tao, ``{DDS2M}: Self-supervised denoising diffusion spatio-spectral model for hyperspectral image restoration,'' in \emph{ICCV}, 2023, pp. 12\,086--12\,096.

\bibitem{sae}
Y.~{Chen}, Z.~{Lin}, X.~{Zhao}, G.~{Wang}, and Y.~{Gu}, ``Deep learning-based classification of hyperspectral data,'' \emph{IEEE Journal of Selected Topics in Applied Earth Observations and Remote Sensing}, vol.~7, no.~6, pp. 2094--2107, 2014.

\bibitem{dbn}
T.~{Li}, J.~{Zhang}, and Y.~{Zhang}, ``Classification of hyperspectral image based on deep belief networks,'' in \emph{ICIP}, 2014, pp. 5132--5136.

\bibitem{ssatt}
R.~{Hang}, Z.~{Li}, Q.~{Liu}, P.~{Ghamisi}, and S.~S. {Bhattacharyya}, ``Hyperspectral image classification with attention-aided cnns,'' \emph{IEEE Transactions on Geoscience and Remote Sensing}, pp. 1--13, 2020.

\bibitem{rssan}
M.~{Zhu}, L.~{Jiao}, F.~{Liu}, S.~{Yang}, and J.~{Wang}, ``Residual spectral-spatial attention network for hyperspectral image classification,'' \emph{IEEE Transactions on Geoscience and Remote Sensing}, pp. 1--14, 2020.

\bibitem{ssan}
H.~{Sun}, X.~{Zheng}, X.~{Lu}, and S.~{Wu}, ``Spectral–spatial attention network for hyperspectral image classification,'' \emph{IEEE Transactions on Geoscience and Remote Sensing}, vol.~58, no.~5, pp. 3232--3245, 2020.

\bibitem{ssftt}
L.~Sun, G.~Zhao, Y.~Zheng, and Z.~Wu, ``Spectral–spatial feature tokenization transformer for hyperspectral image classification,'' \emph{IEEE Transactions on Geoscience and Remote Sensing}, vol.~60, pp. 1--14, 2022.

\bibitem{qformer}
Q.~Zhang, J.~Zhang, Y.~Xu, and D.~Tao, ``Vision transformer with quadrangle attention,'' \emph{IEEE Transactions on Pattern Analysis and Machine Intelligence}, 2024.

\bibitem{dcn}
J.~Dai, H.~Qi, Y.~Xiong, Y.~Li, G.~Zhang, H.~Hu, and Y.~Wei, ``Deformable convolutional networks,'' in \emph{CVPR}, 2017, pp. 764--773.

\bibitem{geochat}
K.~Kuckreja, M.~S. Danish, M.~Naseer, A.~Das, S.~Khan, and F.~S. Khan, ``Geochat: Grounded large vision-language model for remote sensing,'' in \emph{CVPR}, 2024, pp. 27\,831--27\,840.

\bibitem{dotav2}
J.~Ding, N.~Xue, G.-S. Xia, X.~Bai, W.~Yang, M.~Y. Yang, S.~Belongie, J.~Luo, M.~Datcu, M.~Pelillo, and L.~Zhang, ``Object detection in aerial images: A large-scale benchmark and challenges,'' \emph{IEEE Transactions on Pattern Analysis and Machine Intelligence}, vol.~44, no.~11, pp. 7778--7796, 2022.

\bibitem{dior}
K.~Li, G.~Wan, G.~Cheng, L.~Meng, and J.~Han, ``Object detection in optical remote sensing images: A survey and a new benchmark,'' \emph{ISPRS journal of photogrammetry and remote sensing}, vol. 159, pp. 296--307, 2020.

\bibitem{fair1m}
X.~Sun, P.~Wang, Z.~Yan, F.~Xu, R.~Wang, W.~Diao, J.~Chen, J.~Li, Y.~Feng, T.~Xu \emph{et~al.}, ``{FAIR1M}: A benchmark dataset for fine-grained object recognition in high-resolution remote sensing imagery,'' \emph{ISPRS Journal of Photogrammetry and Remote Sensing}, vol. 184, pp. 116--130, 2022.

\bibitem{loveda}
J.~Wang, Z.~Zheng, A.~Ma, X.~Lu, and Y.~Zhong, ``{LoveDA}: A remote sensing land-cover dataset for domain adaptive semantic segmentation,'' in \emph{NeurIPS Track on Datasets and Benchmarks}, vol.~1, 2021.

\bibitem{isaid}
S.~Waqas~Zamir, A.~Arora, A.~Gupta, S.~Khan, G.~Sun, F.~Shahbaz~Khan, F.~Zhu, L.~Shao, G.-S. Xia, and X.~Bai, ``{iSAID}: A large-scale dataset for instance segmentation in aerial images,'' in \emph{CVPRW}, 2019, pp. 28--37.

\bibitem{satlas}
F.~Bastani, P.~Wolters, R.~Gupta, J.~Ferdinando, and A.~Kembhavi, ``{SatlasPretrain}: A large-scale dataset for remote sensing image understanding,'' in \emph{ICCV}, 2023, pp. 16\,772--16\,782.

\bibitem{adamw}
I.~Loshchilov and F.~Hutter, ``Decoupled weight decay regularization,'' in \emph{ICLR}, 2019.

\bibitem{DBLP:conf/cvpr/ShiCHTABRW16}
W.~Shi, J.~Caballero, F.~Huszar, J.~Totz, A.~P. Aitken, R.~Bishop, D.~Rueckert, and Z.~Wang, ``Real-time single image and video super-resolution using an efficient sub-pixel convolutional neural network,'' in \emph{CVPR}, 2016, pp. 1874--1883.

\end{thebibliography}

\twocolumn
\newpage

\makeatletter

\newcommand{\repeatmaketitle}{%
  \if@twocolumn
    \twocolumn[{
      \begingroup
      \centering
      \vspace*{3.25\baselineskip}
      {\@IEEEcompsoc@title@font \@title \par}
      \vspace{11pt}
      {\@IEEEcompsoc@author@font
       \def\IEEEauthorhalign##1{\centering##1}
       \@author \par}
      \vspace{2\baselineskip}
      \endgroup
    }]%
  \else
    \thispagestyle{empty}
    \begingroup
    \centering
    \vspace*{3.25\baselineskip}
    {\@IEEEcompsoc@title@font \@title \par}
    \vspace{11pt}
    {\@IEEEcompsoc@author@font \@author \par}
    \vspace{2\baselineskip}
    \endgroup
  \fi
  \captionsetup{
    font=normalfont,       
    textfont=normalfont,   
    size=normalsize,       
}
}

\newcommand{\@IEEEcompsoc@title@font}{%
  \sffamily\Huge\selectfont
  \ifCLASSOPTIONconference\large\fi
}

\newcommand{\@IEEEcompsoc@author@font}{%
  \sffamily\large\selectfont
  \ifCLASSOPTIONconference\sublargesize\fi
}

\def\IEEEauthorblockN#1{%
  \vskip0.5ex
  {\bfseries #1}\par
}

\def\IEEEauthorblockA#1{%
  \vskip0.25ex
  {\itshape #1}\par
}

\makeatother

\title{HyperSIGMA: Hyperspectral Intelligence Comprehension Foundation Model\\
---Appendix---}

\author{

Di Wang$^*$, 
Meiqi Hu$^*$, 
Yao Jin$^*$,
Yuchun Miao$^*$, 
Jiaqi Yang$^*$, 
Yichu Xu$^*$, 
Xiaolei Qin$^*$, 
Jiaqi Ma$^*$, \\
Lingyu Sun$^*$, 
Chenxing Li$^*$, 
Chuan Fu, 
Hongruixuan Chen, 
Chengxi Han$\dagger$, 
Naoto Yokoya,\\~\IEEEmembership{Member,~IEEE,}
Jing Zhang$\dagger$,~\IEEEmembership{Senior Member,~IEEE,}
Minqiang Xu, 
Lin Liu,
Lefei Zhang,\\~\IEEEmembership{Senior Member,~IEEE,}
Chen Wu$\dagger$,~\IEEEmembership{Member,~IEEE,}
Bo Du$\dagger$,~\IEEEmembership{Senior Member,~IEEE,}\\
Dacheng Tao,~\IEEEmembership{Fellow,~IEEE}
and Liangpei Zhang$\dagger$,~\IEEEmembership{Fellow,~IEEE}

\thanks{$^*$: Equal contribution; $\dagger$: Corresponding author.}

}

\repeatmaketitle

\appendices

This technical appendix provides additional experimental results of our models, as well as detailed implementations that are omitted from the main body of this paper due to the page limit.

Specifically, this appendix is organized as follows:

\begin{itemize}

   \item Sec.~\ref{related_work} reviews the recent works related to this study, which have been omitted from the main paper due to the page limit.

   \item Sec.~\ref{sec_hyperglobal_450k} introduces detailed data collection criteria, data source information, data acquisition method, and data processing pipeline in constructing the HyperGlobal-450K dataset.

    \item Sec.~\ref{sec_hyper_pretrain} outlines the implementation details of model pre-training, including the random selection of channels during data preprocessing and the method for determining the spectral mask ratio. Additionally, we discuss the costs associated with pre-training both the spatial and spectral subnetworks.

    \item Sec.~\ref{sec_hyper_cls} outlines the network structure and experimental settings for the hyperspectral image (HSI) classification task. It then presents detailed analyses and results of the ablation study, followed by additional metrics and visualized classification maps from the comparison experiment.

    \item Sec.~\ref{sec_hyper_det} presents the network structure and experimental settings for hyperspectral target and anomaly detection tasks. It also includes detection probability maps, ROC curves, and corresponding analyses.

    \item Sec.~\ref{sec_hyper_cd} outlines the network structure and detailed experimental settings for the hyperspectral change detection task. Additionally, we present further comparison results, including other evaluation metrics and visualizations of change detection maps from various methods.

    \item Sec.~\ref{sec_hyper_unmixing} introduces the task formulation, network structure for fine-tuning, and experimental settings for hyperspectral unmixing. It then presents detailed quantitative prediction results for each endmember, along with predicted abundance maps and endmember curves.

    \item Sec.~\ref{sec_hyper_denoising} outlines the network architecture, experimental settings, and reconstruction visualizations for the hyperspectral denoising task, including details of the specific noise cases discussed in the main text. Additionally, we conducted experiments with other noise cases, with implementation details and result analyses provided.

    \item Sec.~\ref{sec_hyper_super_resolution} presents the network architecture, experimental settings, and qualitative results for hyperspectral super-resolution. Additionally, we present experimental results using various scale factors.

    \item Sec.~\ref{sec_model_scalability} outlines the experimental settings of employing larger backbone networks (Section 5.7.1 in the main text), and presents more analyses for the model scalability through loading different weights.

    \item Sec.~\ref{sec_cross_modal} presents the network structure and experimental settings for the cross-modal transferability experiments (Section 5.7.3 in the main text).

    \item Sec.~\ref{sec_real_world} provides additional visualization examples from the real-world application experiment (Section 5.7.4 in the main text).

    \item Sec.~\ref{sec_model_complexity} introduces the analysis process of SSA computational complexity, and relevant experimental settings (Section 5.7.5 in the main text). 

    \item Sec.~\ref{sec_datasheet} presents the datasheet of HyperGlobal-450K.

    \item Sec.~\ref{sec:detailed_hsic} provides the per-class accuracies achieved by various methods in hyperspectral image classification tasks.
\end{itemize}

\section{Related Work \label{related_work}}

\begin{table*}[t]
   \newcommand{\tabincell}[2]{\begin{tabular}{@{}#1@{}}#2\end{tabular}}
    \caption{Detailed comparison between HyperSIGMA and existing typical CV and RS foundation models.}
    \centering
 \resizebox{\linewidth}{!}{
   \begin{threeparttable}
    \begin{tabular}{llllllll}
    \hline
        RSFM & Backbone & \#Param.(M) & Pre-training Dataset & Modal\tnote{1} & Pre-training\tnote{2}  & Fine-tuning Task\tnote{3} & Highlight \\ 
        \hline
        \multicolumn{8}{c}{\textbf{Computer Vision Foundation Model}}\\
        \hline
        ViT-G \cite{vit_g} & ViT-G/14 \cite{vit} & 1843 & JFT-3B\cite{vit_g} & RGB & SUP & SLC & \tabincell{l}{Refining the architecture and training of ViT by scaling \\  model and data}\\
        SwinV2\cite{swin_v2} & SwinV2-G\cite{swin_v2} & 3000 & ImageNet22K-ext \cite{swin_v2} & RGB & MIM+SUP & SLC,HOD,SS &  \tabincell{l}{Residual-post-norm method with log-spaced continuous\\ position encoding}\\
        ViTAE\cite{xu2021vitae} & ViTAE-H\cite{vitae_v2} & 644 & ImageNet-1K \cite{imagenet} & RGB & MIM & SLC & \tabincell{l}{Introducing intrinsic inductive biases of convolutions to \\vision transformers}\\
        InternImage\cite{internimage} & InternImage-H\cite{internimage} & 1080  & \tabincell{l}{427 million images \\from different datasets} & RGB  & MIM & SLC,HOD,SS & \tabincell{l}{Large receptive field and adaptive spatial aggregation by \\ deformable convolution} \\
        
        \hline
        \multicolumn{8}{c}{\textbf{Remote Sensing Foundation Model}}\\
        \hline
        RSP \cite{rsp} & \tabincell{l}{ResNet-50\cite{resnet}\\Swin-T\cite{swint}\\ViTAEv2-S\cite{vitae_v2}} & \tabincell{l}{24\\28\\19} & MillionAID \cite{millionaid} & RGB & SUP & SLC,ROD,SS,BCD & \tabincell{l}{Performing classification pre-training on large-scale \\scene-level labeled dataset} \\
        SeCo \cite{seco} & ResNet-50\cite{resnet} & 24 & SeCo-1M\cite{seco} & RGB* & CL &  SLC,MLC,BCD & \tabincell{l}{Learning visual representations based on seasonal changes} \\
        GASSL \cite{gassl} & ResNet-50\cite{resnet}& 24 & \tabincell{l}{fMoW-RGB \cite{fmow} \\ GeoImageNet \cite{gassl}} &  RGB & CL & SLC,TIC,HOD,SS & \tabincell{l}{Improving temporal contrastive learning by geo-location \\ classification} \\
        CACo \cite{caco} & \tabincell{l}{ResNet-18\cite{resnet}\\ResNet-50\cite{resnet}} & \tabincell{l}{11\\24} & CACo-1M\cite{caco} & RGB* & CL & SLC,MLC,SS,BCD,MCD & \tabincell{l}{Considering long-term scene changes in satellite images} \\
        MATTER \cite{matter} & ResNet-34\cite{resnet} & 22 &\tabincell{l}{Collected 14,857\\ Sentinel-2 images} & MSI & CL & MLC,SS,BCD & \tabincell{l}{Enhancing CL by texture refinement and material-based \\cluster residual encoding}\\
        TOV \cite{tov} & ResNet-50\cite{resnet} & 24 & \tabincell{l}{TOV-NI\cite{tov}+\\TOV-RS\cite{tov}} & RGB & CL & SLC, HOD,SS & \tabincell{l}{Stage-wise pre-training on  different data and model layers}\\
        CSPT \cite{cspt} & ViT-B\cite{vit} & 86 & \tabincell{l}{ImageNet \cite{imagenet}+\\MillionAID \cite{millionaid}} & RGB & MIM & SLC,HOD,SS &\tabincell{l}{Consecutive MIM pre-training on natural and RS images}\\
        CMID\cite{cmid} & \tabincell{l}{ResNet-50\cite{resnet}\\Swin-B\cite{swint}} &  \tabincell{l}{24\\88} & MillionAID \cite{millionaid} & RGB & CL+MIM & SLC,ROD,SS,BCD & \tabincell{l}{Integrating MIM, global CL and local self-distillation in a \\ unified teacher–student architecture}\\
        GeRSP \cite{gersp} & ResNet-50\cite{resnet}& 24 & \tabincell{l}{ImageNet \cite{imagenet}+\\MillionAID \cite{millionaid}} & RGB & SUP+CL & SLC,HOD,SS &  \tabincell{l}{Integrating self-supervised RS image pre-training and \\supervised natural image pre-training} \\
        RingMo \cite{ringmo} &  \tabincell{l}{ViT-B\cite{vit}\\Swin-B\cite{swint}} & \tabincell{l}{86\\88} &  \tabincell{l}{Collected 2 million\\ RS images} & RGB & MIM & SLC,HOD,SS,BCD & \tabincell{l}{Designing patch incomplete mask strategy to preserve \\small objects during MIM} \\
        RVSA \cite{rvsa} & \tabincell{l}{ViT-B+RVSA\cite{vit}\\ViTAE-B+RVSA\cite{xu2021vitae}} & \tabincell{l}{86\\89} & MillionAID \cite{millionaid} & RGB & MIM & SLC,ROD,SS & \tabincell{l}{Developing rotated varied-size attention to handle \\oriented objects in RS scenes}\\
        SatMAE \cite{satmae} &   ViT-L\cite{vit} & 307 & \tabincell{l}{fMoW-RGB \cite{fmow} \\ fMoW-Sentinel \cite{satmae}} & RGB,MSI & MIM & SLC,TIC,MLC,SS & \tabincell{l}{Extending MAE\cite{mae} to temporal and spectral dimension}\\
        Scale-MAE \cite{scale_mae} & ViT-L\cite{vit} & 307 & FMoW-RGB \cite{fmow} & RGB & MIM & SLC,SS & \tabincell{l}{Reconstructing low/high frequency images at different \\ scales with ground sample distance positional encoding}\\
        BillionFM \cite{bfm} & ViT-G12X4\cite{vit} & 2400 & MillionAID \cite{millionaid} & RGB & MIM &  HOD,SS & \tabincell{l}{Billion-scale parameter with ViT of parallel configration} \\
        GFM \cite{gfm} & Swin-B\cite{swint} & 88 & GeoPile\cite{gfm}   & RGB & MIM &  SLC,SS,BCD,SR & \tabincell{l}{Improving MIM via continual pre-training by \\the guidance of natural knowledge with}\\
        CROMA \cite{croma} & ViT-L\cite{vit} & \tabincell{l}{86\\307} & SSL4EO-S12\cite{ss4leo_s12} & MSI,SAR & CL+MIM & SLC,MLC,SS,MMC,MMS& \tabincell{l}{Cross-modal CL with multimodal fusion and MIM} \\
        Cross-Scale MAE\cite{cross-scale_mae} & ViT-L\cite{vit} & 307 & fMoW-RGB\cite{fmow} & RGB &CL+MIM & SLC,SS & \tabincell{l}{Cross-scale CL and alignment with MIM}\\
        CtxMIM \cite{ctxmim} & Swin-B\cite{swint} & 88 & \tabincell{l}{Collected 1,287,248 \\WorldView-3 images} & RGB & MIM & SLC,HOD,SS,IS & \tabincell{l}{Providing context consistency constraint for masked \\ patches during reconstruction} \\
        SpectralGPT \cite{spectralgpt} & ViT-H\cite{vit} & 632 & \tabincell{l}{fMoW-Sentinel\cite{satmae}+\\BigEarthNet\cite{bigearthnet}} & MSI & MIM & SLC,MLC,SS,BCD & \tabincell{l}{3D patch masking and progressive pre-training}\\
        FG-MAE\cite{fgmae} & ViT-H\cite{vit} & 632 & SS4LEO-S12\cite{ss4leo_s12} & MSI,SAR & MIM & SLC,MLC,SS & \tabincell{l}{Using various RS image features as MIM's reconstruction\\ targets}\\
        SkySense \cite{skysense} & \tabincell{l}{Swin-H\cite{swint}+\\ViT-L\cite{vit}} &  2060 & \tabincell{l}{21.5 million multimodal \\RS temporal sequences} & RGB,MSI,SAR & CL & \tabincell{l}{SLC,MLC,TIC,HOD,ROD,SS,\\BCD,MMS,MMTS,MMC} & \tabincell{l}{Multi-granularity and cross-modal contrasts with \\geo-context prototype learning}\\
        msGFM\cite{msgfm} & Swin-B\cite{swint} & 88 & GeoPile-2\cite{msgfm} & RGB,MSI,SAR,DSM & MIM & MLC,SS,CR,PS& \tabincell{l}{Reconstructing targets based on cross-sensor features}\\ 
        DOFA \cite{dofa} & ViT-L\cite{vit} & 307 & \tabincell{l}{8 million mixed samples \\from different modalities} & RGB,MSI,SAR,HSI & MIM & SLC,MLC,SS,MMS & \tabincell{l}{Learning visual embeddings based on input wavelengths}\\
        LeMeViT\cite{lemevit} & LeMeViT-B\cite{lemevit} & 53 & \tabincell{l}{ImageNet \cite{imagenet}\\MillionAID \cite{millionaid}} & RGB & SUP & SLC,SS,BCD & \tabincell{l}{Proposing dual cross attention based on learnable meta \\tokens to accelerate inference}\\
        SMLFR \cite{smlfr} &  ConvNext-L\cite{convnext} & 198 & GeoSense\cite{smlfr} & RGB & MIM & ROD,SS,BCD & \tabincell{l}{Sparse MIM with low-frequency reconstruction}\\
        MTP \cite{mtp} & \tabincell{l}{ViT-L+RVSA\cite{mtp}\\InternImage-XL\cite{internimage}} & \tabincell{l}{305\\335} & SAMRS\cite{samrs} & RGB & SUP & SLC,HOD,ROD,SS,BCD & \tabincell{l}{Simultaneously pre-training existing models on multitasks} \\
        DECUR\cite{decur} & ResNet-50\cite{resnet} & 24 &  \tabincell{l}{SSL4EO-S12 \cite{ss4leo_s12}\\GeoNRW\cite{geonrw}\\SUN-RGBD\cite{sun-rgbd}} & \tabincell{l}{RGB,MSI,SAR,DEM,Depth} & CL & MLC,MMC,MMS & \tabincell{l}{Performing both inter-modal and intra-modal CL based on \\ decoupling common and unique dimensions}\\
        SoftCon \cite{softcon} & \tabincell{l}{ResNet-50\cite{resnet}\\ViT-B\cite{vit}} & \tabincell{l}{24\\86} & SSL4EO-S12\cite{ss4leo_s12} & MSI,SAR & CL & SLC,MLC,SS,BCD & \tabincell{l}{Continual pre-training with multi-label guided soft CL}\\
        MA3E\cite{ma3e} & ViT-B\cite{vit} & 86 & MillionAID \cite{millionaid} & RGB& MIM & SLC,ROD,SS & \tabincell{l}{Learning rotation-invariant representations by restoring \\ the angle variation during pre-training}\\
        SelectiveMAE\cite{selectivemae} & ViT-L\cite{vit} & 307 & OpticalRS-13M\cite{selectivemae} & RGB & MIM & SLC,HOD,ROD,SS & \tabincell{l}{Partial Reconstruction with progressive semantic token \\ selection  for efficient pre-training} \\
        UPetu \cite{upetu} & ConvNext-B\cite{convnext} & 89 & GeoSense\cite{smlfr} & RGB & SUP & SLC,SS,BCD  & \tabincell{l}{Parameter-efficient fine-tuning with quantization adapter\\ and context-aware prompting}\\
        \hline
        HyperSIGMA & ViT-H+SSA\cite{vit} & 1273 & HyperGlobal-450K & HSI & MIM & \tabincell{l}{HIC,HTD,HAD,HCD,\\HIU,HID,HSR} & \tabincell{l}{Addressing redundancy by sparse sampling attention and \\fusing features by spectral enhancement}\\
        \hline
    \end{tabular}
    \begin{tablenotes}
    \item[1] MSI: Multispectral Image, HSI: Hyperspectral Image, DSM: Digital Surface Model, DEM: Digital Elevation Model. ``*'' means the RGB channels from Sentinel-2 multispectral images.
    \item[2] SUP: Supervised Pre-training. CL: Contrastive Learning. MIM: Masked Image Modeling.
    \item[3] SLC: Single-label Classification. MLC: Multi-label Classification. TIC: Temporal Image Classification. HOD: Horizontal Object Detection. ROD: Rotated Object Detection. SS: Semantic Segmentation. IS: Instance Segmentation. BCD: Bi-temporal Change Detection. MCD: Multi-temporal Change Detection. SR: Super-Resolution. CR: Cloud Removal. PS: Pan-Sharpening. MMC: Multimodal Classification. MMS: Multimodal Segmentation. MMTS: Multimodal Temporal Segmentation. HIC: Hyperspectral Image Classification. HTD: Hyperspectral Target Detection. HAD: Hyperspectral Anomaly Detection. HCD: Hyperspectral Change Detection. HIU: Hyperspectral Image Unmixing. HID: Hyperspectral Image Denoising. HSR: Hyperspectral Super-Resolution.
  \end{tablenotes}
    \end{threeparttable}
    }
    \label{tab:compre_cv_rs}
\end{table*}

Table \ref{tab:compre_cv_rs} highlights key differences between the proposed HyperSIGMA model and current representative computer vision (CV) and remote sensing (RS) foundation models across dimensions such as backbone architecture, pre-training datasets, and modal types. Additionally, this section also introduces related works in hyperspectral image (HSI) processing, multi-head self-attention mechanisms, and large-scale RS datasets.

\subsection{Hyperspectral Image Processing}

Traditional HSI dimensionality reduction approaches primarily fall into two categories: band selection and channel compression. Band selection aims to identify the most discriminative channels by employing various criteria \cite{dr_fs}. On the other hand, channel compression techniques project valuable information into a lower-dimensional space, such as principal component analysis \cite{PCAinHSIclassify}, independent component analysis \cite{ICAinHSIclassify}, and linear discriminant analysis \cite{LDAinHSIclassify}. While these methods have been pivotal in traditional HSI processing, which typically processes a single image, their application to datasets like HyperGlobal-450K poses challenges due to the massive number of samples. Utilizing these methods in such contexts not only restricts data diversity but also introduces considerable computational overhead. To address this, we adopt a strategy during pre-training where, before feeding the data into the model, we perform dimensionality reduction by randomly selecting a fixed number of channels for each batch of samples in every iteration. This approach significantly enhances data richness while ensuring training efficiency.

Early HSI feature extraction methods relied on spectral data, either in raw pixel form or after dimensionality reduction, for various machine learning algorithms like k-nearest neighbors \cite{KNNinHSIclassify}, multinomial logistic regression \cite{MLRinhsi}, random forest \cite{rfinhsi}, and support vector machine \cite{svmforhsiclassify}. Spatial information was later incorporated through handcrafted descriptors such as morphological profiles \cite{MP_1}, superpixels \cite{Fang2015SS}, gray-level co-occurrence matrices \cite{hsi_glcm2}, and filter algorithms \cite{mugnet}. However, these methods relied heavily on expert knowledge and offered limited performance \cite{representlearning}. Nowadays, deep learning-based extractors, particularly CNNs, emerge as the mainstream solution, capturing spectral~\cite{1dcnn}, spatial~\cite{deeplab_hsi}, and spectral-spatial features~\cite{ssrn} at various levels of abstraction~\cite{fu2024resc,ITER,miao2023dds2m}. CNNs gradually replaced fully connected networks~\cite{sae,dbn} due to their ability to preserve data structure and local perception. Attention mechanisms further enhanced feature extraction by assigning weights to channels~\cite{ssatt} or spatial positions~\cite{rssan}, improving long-range perception~\cite{selfattention,ssan}. Recently, transformer-based networks~\cite{spectralformer,ssftt} have become popular for their superior context modeling capability, leveraging Multi-Head Self-Attention (MHSA). Nevertheless, existing extractors lacked generalization across scenarios since the models are trained separately on each scene. In this study, we propose HyperSIGMA, which leverages pre-training on large-scale hyperspectral data for generalized feature extraction.

\subsection{Multi-Head Self-Attention Mechanism}

The MHSA module serves as the cornerstone of transformer \cite{selfattention,vit}, enabling adaptive global context capture. However, its computational complexity scales quadratically with input token length. Swin Transformer \cite{swint} addresses this issue by introducing WMHSA, which confines attention within non-overlapping windows, leading to linear complexity. Nevertheless, fixed-size windows in WMHSA struggle with diverse object sizes and distributions in real scenes. VSA \cite{vsa} addresses this by introducing adaptive scaling and offsetting of windows, later extended to quadrangles \cite{qformer}, enhancing attention flexibility. RVSA \cite{rvsa} builds upon VSA, introducing learnable rotation for objects with varied orientations, and enhancing context extraction. Besides window-based approaches, deformable mechanisms have gained traction. Deformable convolution \cite{dcn} adapts convolution kernel sampling positions, followed by attention methods like Deformable Attention \cite{deformable_detr} and DMHA \cite{dat}. In this research, we propose a novel sparse sampling attention mechanism that enhances flexibility by adaptively capturing contexts within deformable regions, each with a few sampling points to handle redundancy in HSIs.

\begin{table*}[t]
\scriptsize
\newcommand{\tabincell}[2]{\begin{tabular}{@{}#1@{}}#2\end{tabular}}
\centering
\caption{Overview of representative hyperspectral satellites.
}
\label{tab:hyperspectral_satellites}
\begin{threeparttable} 
\begin{tabular}{lcccccccccc}
\hline
\makecell[{{l}}]{Satellite \\ Name}& \makecell{Launch \\ Date} & \makecell{Launch Country/ \\ Organization} & \makecell{Hyperspectral \\ Sensor} & \makecell{Spatial\\ Resolution} & \makecell{Spectral \\ Resolution} & \makecell{Spectral \\ Range} & \makecell{Number of \\ Bands} & \makecell{Swath \\ Width} & \makecell{Free \\ Access} & \makecell{Global \\ Coverage} \\ 
\hline
EO-1\tnote{1}      & 2000 & USA                & Hyperion & 30 m       & 10 nm        & 400–2500 nm & 242       & 7.7 km   & Full   & Yes    \\ 
PROBA-1\tnote{2}  & 2001 & ESA                & CHRIS    & 34 m       & 1.25–11 nm   & 400–1050 nm & 62        & 14 km    & Limited\tnote{a} & Yes   \\ 
HuanJing-1A\tnote{3} & 2008 & China             & HSI      & 100 m      & 5 nm         & 450–950 nm  & 110-128   & 50 km    & No     & Yes   \\ 
HICO\tnote{4}     & 2009 & USA                & HICO     & 90 m       & 5 nm         & 350–1070 nm & 128       & 42 km    & Yes   & No    \\ 
Zhuhai-1\tnote{5}  & 2018 & China              & HSI      & 10 m       & 2.5 nm       & 400–1000 nm & 32        & 150 km   & No    & Yes    \\ 
GaoFen-5\tnote{6} & 2018 & China              & AHSI     & 30 m       & 5/10 nm      & 400–2500 nm & 330       & 60 km    & No    & Yes    \\ 
HysIS\tnote{7}   & 2018 & India              & HSI      & 30 m       & 10 nm        & 400–2400 nm & 316       & 30 km    & No    & Yes    \\ 
DESIS\tnote{8}    & 2018 &  \makecell{Germany\\ \&  USA}   & DESIS    & 30 m       & 2.5 nm       & 400–1000 nm & 235       & 30 km    & Partially\tnote{b} & No   \\ 
ZY-1 02D\tnote{9}   & 2019 & China              & AHSI     & 30 m       & 10/20 nm     & 400–2500 nm & 166       & 60 km    & No    & Yes   \\ 
HISUI\tnote{10}    & 2019 & Japan              & HISUI    & 20 m       & 10/12.5 nm   & 400–2500 nm & 185       & 20 km    & No    & No     \\ 
PRISMA\tnote{11}    & 2019 & Italy              & HYC      & 30 m       & 10 nm        & 400–2500 nm & 239       & 30 km    & Partially\tnote{c} & Yes  \\ 
EnMAP\tnote{12}     & 2022 & Germany            & EnMAP    & 30 m       & 5/10 nm      & 420–2450 nm & 224       & 30 km    & Limited\tnote{d} & Yes   \\ 
\hline
\end{tabular}%
    \begin{tablenotes}
    \item[1] \url{https://www.usgs.gov/centers/eros/eo-1-sensors}
    \item[2] \url{https://earth.esa.int/eogateway/catalog/proba-chris-level-1a} 
    \item[3] \url{https://www.cresda.com/zgzywxyyzx/wxzy/hj/article/20220224113103300357009.html}
    \item[4] \url{https://oceancolor.gsfc.nasa.gov/data/hico/} 
    \item[5] \url{http://114.116.226.59/chinese/satellite/chinese/zhuhai}
    \item[6] \url{http://www.sasclouds.com/chinese/satellite/chinese/gf5}
    \item[7] \url{https://www.eoportal.org/satellite-missions/hysis\#mission-capabilities}
    \item[8] \url{https://www.tbe.com/en-us/suppliers/SiteAssets/DESIS\_FAQ.pdf} 
    \item[9] \url{http://114.116.226.59/chinese/satellite/chinese/zy102d} 
    \item[10] \url{https://www.jspacesystems.or.jp/en/project/observation/hisui}
    \item[11] \url{https://earth.esa.int/eogateway/documents/20142/2798695/3.PRISMA\%2Bdata\%2Bintroduction.pdf/31642194-c6cb-3160-44c3-0ea8f87130b3}
    \item[12] \url{https://www.sciencedirect.com/science/article/pii/S0034425723001839}
    \item[a] PROBA-1: The free access to PROBA-1 data is limited due to the challenges associated with facilitating large-scale, global data downloads, which significantly restricts widespread availability.
    \item[b] DESIS: DESIS data is available for ordering and purchase via the TCloud web interface (https://teledyne.tcloudhost.com).
    \item[c] PRISMA: Each User will be allowed to
use only a portion of the system resources, through Priority and Quota.
   \item[d] EnMAP: The licensing terms for EnMAP data download strictly restrict redistribution to third parties. 
  \end{tablenotes}
    \end{threeparttable}
\label{hyperspectral_satallite}
\end{table*}

\subsection{Large-Scale Remote Sensing Dataset}

Pre-training of RS foundation models typically relies on large-scale datasets. For aerial RGB images, the fMoW-RGB~\cite{fmow} dataset supports the development of numerous classical RS models~\cite{gassl, satmae, scale_mae}. Similarly, the MillionAID dataset~\cite{millionaid}, comparable to ImageNet-1K~\cite{imagenet} with one million samples, is widely utilized~\cite{rsp, rvsa, bfm}. Recent advancements in RS multitask pre-training~\cite{mtp} and the RS Vision-Language model GeoChat~\cite{geochat} leverage the SAMRS dataset~\cite{samrs}, a large-scale RS dataset with instance and segmentation labels built upon existing RS object detection datasets \cite{dotav2, dior, fair1m}, containing 105,090 images and 1,668,241 instances, far surpassing the capacity of existing high-resolution RS segmentation datasets~\cite{loveda, isaid}. For multispectral images, Cong et al.~\cite{satmae} introduced a multispectral version of fMoW-RGB using Sentinel-2 images, called fMoW-Sentinel. Hong et al.~\cite{spectralgpt} further selected 1,067,070 Sentinel-2 images from fMoW-Sentinel and BigEarthNet-S2~\cite{bigearthnet} to pre-train the SpectralGPT model. Besides, multimodal models~\cite{fgmae, croma} that incorporate SAR images are pre-trained on the SSL4EO-S12 dataset~\cite{ss4leo_s12}. Other datasets like GeoPile \cite{gfm}, Satlas \cite{satlas}, and SeCo-1M \cite{seco} also contribute significantly to advancing RS foundation models. Notably, our HyperGlobal-450K stands out as the first large-scale hyperspectral pre-training dataset, containing about 450K HSIs. This volume equates to over 20 million trispectral images with non-overlapping channels, facilitating the development of hyperspectral foundation models.

\section{The HyperGlobal-450K Dataset \label{sec_hyperglobal_450k}}

\subsection{Data Collection Criterion}

 Before data acquisition, we conducted a thorough survey of representative hyperspectral satellites. Table \ref{tab:hyperspectral_satellites} summarizes key specifications, including launch date, spatial and spectral resolution, wavelength range, and data availability. We prioritized global coverage and free access as essential factors for creating large-scale, worldwide hyperspectral datasets. Based on these criteria, the Earth Observing-1 (EO-1) satellite emerged as a suitable choice.

 In addition to EO-1, which is commonly used in hyperspectral tasks, pre-training with data from multiple sensors can enhance model representation. To this end, we also employed GF-5 (Gaofen-5), another hyperspectral sensor. As shown in Table \ref{hyperspectral_satallite}, EO-1 and GF-5 complement each other temporally. While the GF-5 dataset was previously our in-house data, it will soon be released to foster further research within the HSI processing community.

\subsection{Data Source}

HyperGlobal-450K contains images from two types of sensors: the EO-1 Hyperion and GF-5. The EO-1 Hyperion, as part of a NASA mission launched in November 2000\footnote{\url{https://www.usgs.gov/centers/eros/science/usgs-eros-archive-earth-observing-one-eo-1-hyperion}}, operated at an orbital altitude of about 705 km. It scanned the Earth's surface across 242 spectral bands, ranging from visible to shortwave infrared wavelengths, with a spectral resolution of 10 nm and a spatial resolution of 30 meters. The GF-5, launched by China in May 2018\footnote{\url{https://grid.cpeos.org.cn}}, also operates at an altitude of about 705 km. It covers a spectral range from visible light to shortwave infrared wavelength (0.4-2.5$\mu$m), capturing data across 330 hyperspectral bands with a spectral resolution of up to 5 nm and a spatial resolution of 30 meters. The EO-1 satellite was retired in 2017, while the GF-5 remains operational.

\subsection{Data Acquisition}

For the EO-1 Hyperion sensor, we downloaded the images acquired during 2011-2017 from the websites shown in Table \ref{hyperspectral_satallite}. While for the GF-5 data, 215 images from Nei Mongol, Hebei, Henan, Jilin, and Liaoning Provinces in China are employed, respectively. These images provide coverage of vast areas of farmland, grasslands, and forest regions, thereby supplementing abundant hyperspectral vegetation observation data. Notably, since we mainly considered the high-dimensional channel characteristic rather than the wavelength information for specific channels, we directly use the L1Gst-level (radiance) and L1-level (digital number, i.e., DN) products for EO-1 and GF-5, respectively.

\subsection{Data Processing}

For EO-1 Hyperion images, the data processing process mainly incorporates four steps: cloudy image removal, image location selection, band refinement, and clipping. 

\noindent\textbf{Cloudy Image Removal} We screen the data with a cloud cover percentage less than 5\% and exclude the data with unknown cloud cover percentages. This step was crucial to ensure the image quality.

\noindent\textbf{Image Location Selection} The EO-1 images are named according to the observation year and their path/row location in the Worldwide Reference System (WRS), which divides the globe into a grid of 233 paths and 248 rows, resulting in approximately 58k path/row combinations. However, for any given year, both the specific locations covered by the images and the number of images per path/row are uncertain. To ensure comprehensive global coverage and maintain balance across locations, we implement the following location selection strategy:

For each year, we randomly selected one image from each path/row location, ensuring that every location with available data contributed at least one image to the dataset. It is important to note that the downloaded images may not cover all 58k path/row combinations. However, we only require that all available path/row locations be represented after filtering out cloud-covered images. 

Finally, we obtained a total of 1,486 EO-1 images in size of 2,000$\times$256, covering all continents across the globe.

\noindent\textbf{Band Refinement} We remove bad bands and water vapor absorption bands (both are presented in official instruction), producing the HSIs that contain 175 channels.

\noindent\textbf{Clipping} We clip the obtaining EO-1 images to patches in size of 64 $\times$ 64 without overlaps.

Compared to the EO-1 data, GF-5 images were cloudy-free and did not need to remove cloudy samples and select locations. Therefore, when processing GF-5 images, we directly used 150 spectral bands ranging from 0.4 to 1.0 $\mu$m (visible light and near-infrared), as this range is beneficial for typical RS land object recognition. The obtained HSIs are also clipped to 64 $\times$ 64 pixels for subsequent pre-training.

\section{Pre-training \label{sec_hyper_pretrain}}

\subsection{Implementation Details}

\begin{figure}[t]
  \centering
  \includegraphics[width=0.7\linewidth]{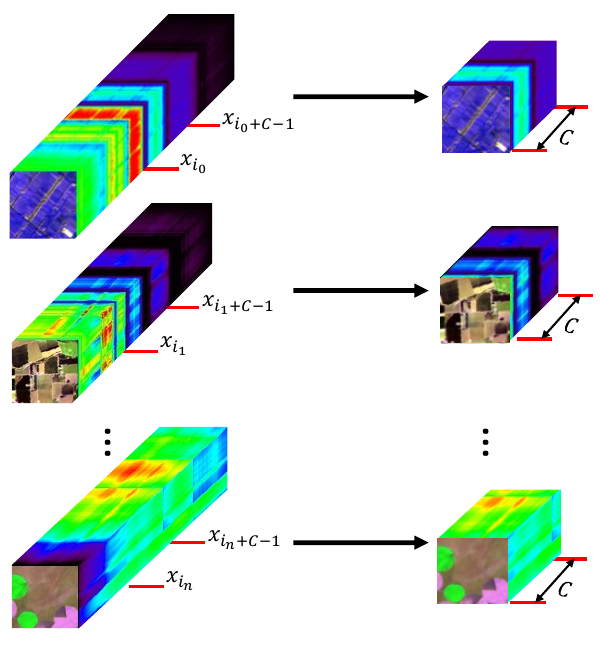}
  \caption{
  Illustration of the channel selection strategy. For images with varying channel counts, we randomly select $C$ consecutive channels.
  }
\label{fig: band_selection}
\end{figure}

\noindent\textbf{Random Channel Selection} Specifically, for any original HSI \(\mathbf{X}_{HSI} \in \mathbb{R}^{H \times W \times L} = \{x_1, x_2, \ldots, x_L\}\) with \(L\) channels, where \(x_i \in \mathbb{R}^{H \times W}\) represents a channel, we randomly select continuous channels \(\{x_i, x_{i+1}, \ldots, x_{i+C-1}\}\) by determining a value of \(i\) that satisfies \(1 \leq i \leq L+1-C\). This results in \(\mathbf{X}_0 \in \mathbb{R}^{H \times W \times C}\), which is then processed by the network as described in Sec. 4.1.2 and 4.1.3 of the main text. During pre-training, \(C\) is fixed to ensure that different HSIs have the same number of channels. This channel selection strategy is illustrated in Fig. \ref{fig: band_selection}. In our implementation, \(C\) is set to 100. Additionally, all HSIs are normalized by dividing by 4,000 based on our exploratory analysis of the data.

\begin{figure}[t]
  \centering
  \includegraphics[width=\linewidth]{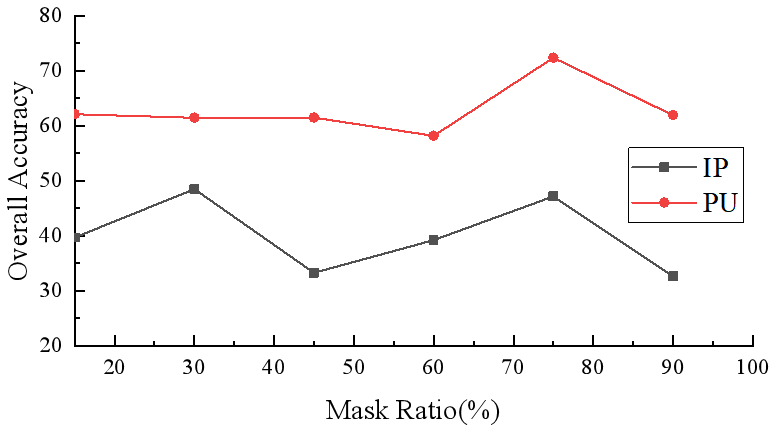}
  \caption{
  Fine-tuning accuracies on the Indian Pines (IP) and Pavia University (PU) datasets using spectral subnetworks pre-trained on the HyperGlobal-450K dataset with different mask ratios.
  }
\label{spec_mask_ratio}
\end{figure}

\begin{figure}[t]
  \centering
  \includegraphics[width=\linewidth]{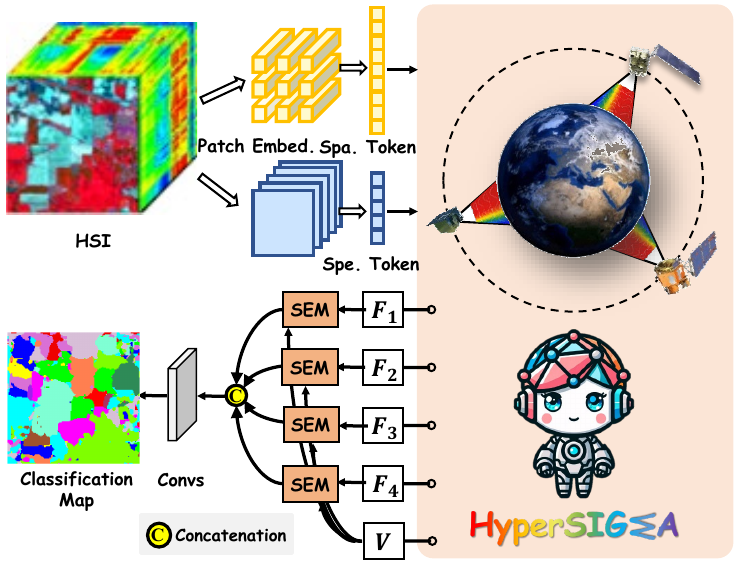}
  \caption{Diagram illustrating the application of HyperSIGMA for HSI classification.}
\label{fig: HSIClassificationFlowchart}
\end{figure}

\begin{figure*}[t]
  \centering
  \includegraphics[width=0.9\linewidth]{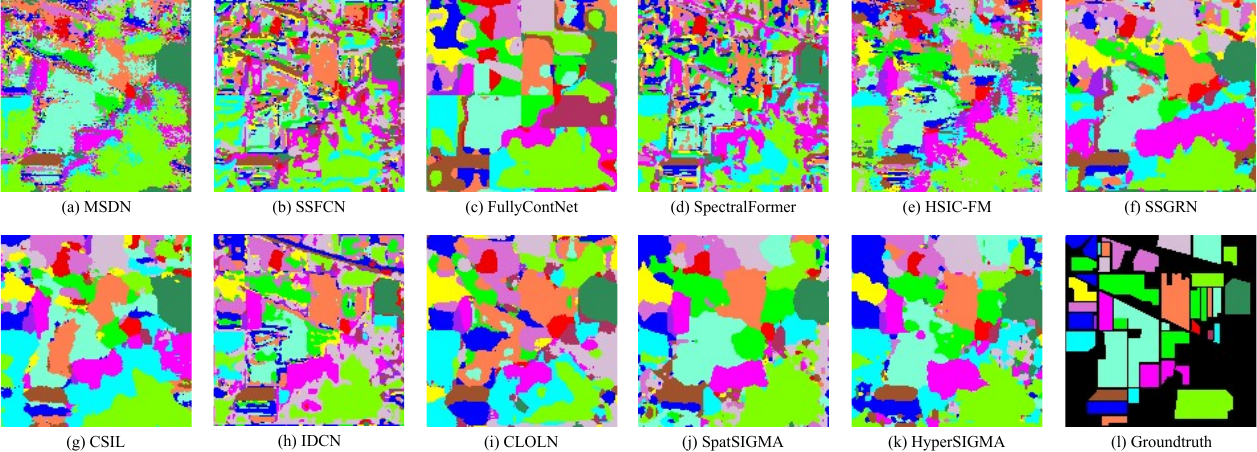}
  \caption{Predicted classification maps from various methods on the Indian Pines dataset \cite{indian_pines}.}
\label{fig:IP_ClassificationMap}
\vspace{-0.5em}
\end{figure*}

\noindent\textbf{Spectral Mask Ratio} Concretely, we pre-trained spectral subnetworks with different mask ratios ranging from 0.15 to 0.9, in intervals of 0.15, for 400 epochs with a batch size of 2,048. We used the AdamW optimizer \cite{adamw}, setting the learning rate to 0.00015 and weight decay to 0.05, with all other settings following those of MAE \cite{mae}. For convenience, the number of tokens \(N_{spec}\) was set equal to \(C\), mapping each channel to a token. After pre-training, we fine-tuned these networks on the Indian Pines (IP) and Pavia University (PU) datasets for HSI classification tasks. For IP, we used 10 samples per category for training, and for PU, 20 samples per category, with the remaining samples used for testing. Each input sample included all channels, and a 33$\times$33 patch centered on the target pixel was fed into the network. \(N_{spec}\) remained set to 100. The network output, with a shape of \(N_{spec} \times D\), was passed through an average pooling layer to reduce the dimensionality to \(D\), followed by a linear layer for classification. We used cross-entropy loss for training and evaluated performance using overall accuracy (OA). The results, shown in Fig. \ref{spec_mask_ratio}, indicate that the best performance across all datasets was achieved with \(R_{spec}=0.75\). Therefore, we used this mask ratio for the subsequent pre-training of spectral subnetworks.

\noindent\textbf{Pre-training Cost} The pre-training required significant time and computational resources, as detailed in Table \ref{trn_cost}. The differences between SpatViT and SpecViT within the same version mainly stem from the variations in embedding layers. Additionally, we employed Gradient Checkpointing when GPU memory was insufficient. 

\section{Hyperspectral Image Classification \label{sec_hyper_cls}}

\begin{table}[t]
  \caption{Pre-training costs for various ViT versions in spatial and spectral subnetworks.}
  \centering

  \begin{threeparttable}
    \resizebox{\linewidth}{!}{
  \begin{tabular}{lcccc}
  \hline
  SubNetwork  & \#Param.(M) & \#GPU & Memory(MB)\tnote{1} & Time (days) \\
  \hline
  SpatViT-Base  & 90 & 64  & 3,744 & 2.3  \\
  SpatViT-Large & 309 & 64  & 8,390 & 4.4 \\
  SpatViT-Huge  & 638 & 128 & 13,419  & 7.6 \\
  \hline
  SpecViT-Base  & 88 & 32 & 6,224 &  3.3 \\
  SpecViT-Large & 307 & 64 & 9,347 & 4.3 \\
  SpecViT-Huge  & 635 & 128 & 13,360 & 7.6 \\
  \hline
\end{tabular}
 }
 \begin{tablenotes}
  \scriptsize
  \item[1] Using gradient checkpointing if Out-of-memory (OOM) is observed.
\end{tablenotes}
\end{threeparttable}
 \label{trn_cost}
\end{table}

\begin{figure*}[h]
  \centering
  \includegraphics[width=0.9\linewidth]{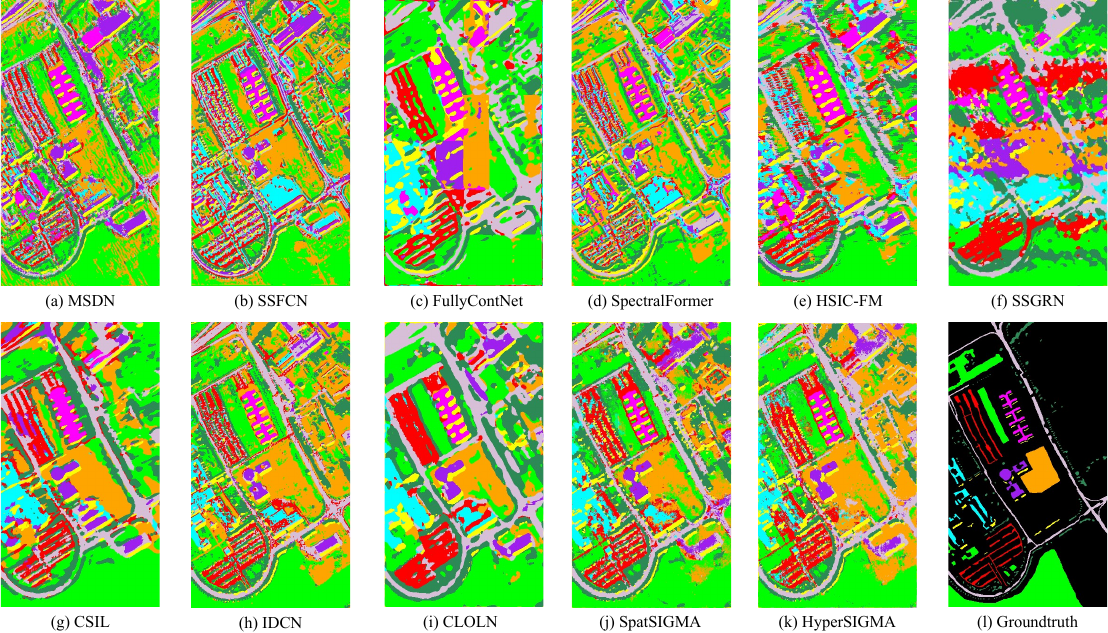}
  \caption{Predicted classification maps from various methods on the Pavia University dataset\protect\footnotemark.}
\label{fig:PU_ClassificationMap}
\vspace{-1em}
\end{figure*}
\footnotetext{\url{https://www.ehu.eus/ccwintco/index.php/Hyperspectral\_Remote\_Sensing\_Scenes}}

\begin{table}[t]
\centering
    \caption{
    Fine-tuning accuracies of the spatial subnetwork with varying numbers of sampling points on the Indian Pines (IP) and Pavia University (PU) datasets. FA: Full Attention. NSP: Number of Sampling Points. \textbf{\color{red}{Best}} and \textbf{\color{blue}{2nd-best}} results are highlighted.
    }
\label{tab: numbers of sampling points}
\begin{tabular}{lccc}
\hline
Method &  NSP &  IP & PU \\
\hline
FA \cite{vit} & - & 58.25 & 77.54 \\
SSA & 4 & \textbf{\color{blue}{61.54}} & 80.81 \\
SSA & 8 & \textbf{\color{red}{63.92}} &\textbf{\color{red}{82.00}}\\
SSA & 12 & 60.57 &\textbf{\color{blue}{ 81.12 }} \\
SSA & 16 &  60.85 & 80.43 \\
\hline
\end{tabular}
\vspace{-1em}
\end{table}

\begin{table}[t]
\centering
    \caption{
    Fine-tuning accuracies of the spatial subnetwork with various attentions on the Indian Pines (IP) and Pavia University (PU) datasets. FA: Full Attention; WS: Window Size; DR: Downsampling Rate; NAB: Number of Attention Buckets; NSP: Number of Sampling Points. \textbf{\color{red}{Best}} and \textbf{\color{blue}{2nd-best}} results are highlighted.
    }
\begin{tabular}{lcccccc}
\hline
Method & WS  & NAB & DR  & NSP  &  IP & PU \\
\hline
FA \cite{vit} & - & - & - & - &  58.25 & 77.54 \\
WMHSA \cite{swint} &4 &- & - &- & 59.18 &  77.68\\
WMHSA \cite{swint} &8 & - & - &- & 56.79  & 78.84  \\
VSA \cite{vsa} &4 &- & - &- & 60.54  & 78.62 \\
VSA  \cite{vsa} & 8 &- & - &- & 60.72 & 80.37 \\
RVSA \cite{rvsa} & 4 &- & - &- & \textbf{\color{blue}{62.74}} & 79.49 \\
RVSA \cite{rvsa} & 8 & - & - & - & 60.61 & 79.74 \\
 NLSA \cite{nlsa} & - & 4 & - & - & 58.98 & 79.09 \\
 NLSA \cite{nlsa} & - & 8 & - & - & 61.08 & \textbf{\color{blue}{80.54}} \\
DMHA \cite{dat} & - & - & 4 &- & 47.06 & 69.93 \\
DMHA \cite{dat} & - & - & 8 &- & 45.78 & 67.98 \\
\hline
SSA &- & - &- & 8 & \textbf{\color{red}{63.92}} & \textbf{\color{red}{82.00}} \\

\hline
\end{tabular}
\label{tab: different attentions}
\end{table}

\begin{figure*}[t]
  \centering
  \includegraphics[width=0.9\linewidth]{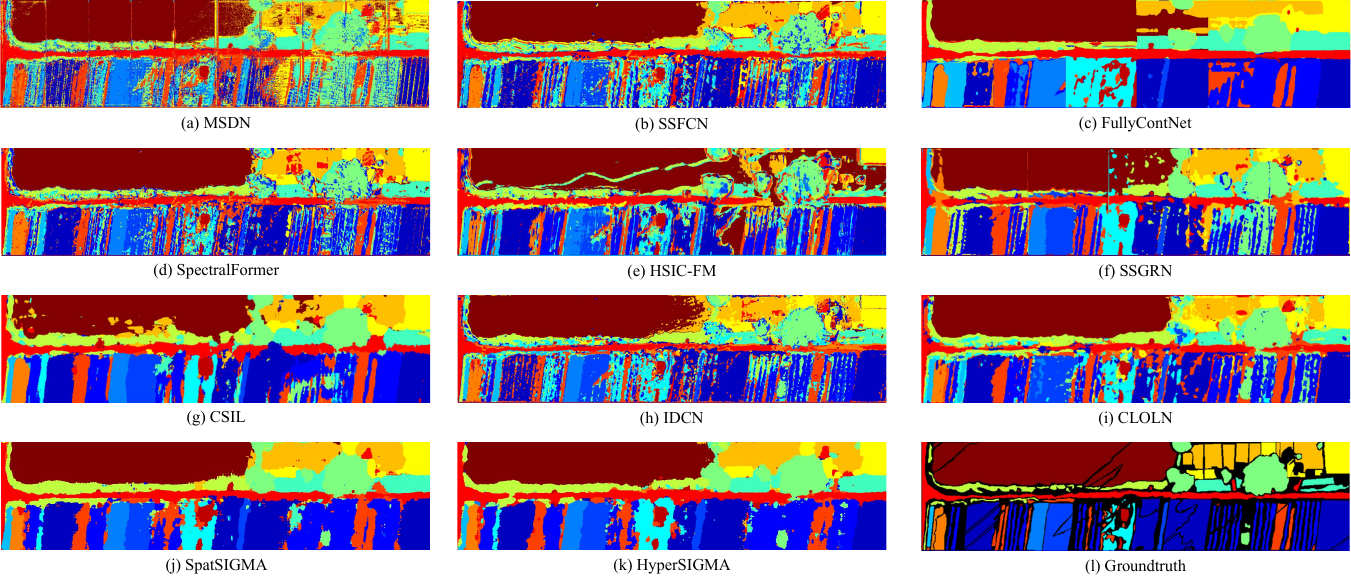}
  \caption{Predicted classification maps from various methods on the HanChuan dataset \cite{whu_hi1}.}
\label{fig:HC_ClassificationMap}
\vspace{-0.5em}

\end{figure*}

\begin{table}[t]
    \centering
    \caption{
    Fine-tuning accuracies of the spatial subnetwork using various pre-training weights on Indian Pines (IP) and Pavia University (PU) datasets. \textbf{\color{red}{Best}} and \textbf{\color{blue}{2nd-best}} results are highlighted.
    }
    \begin{tabular}{lcc}
    \hline
    Pre-training Dataset & IP & PU \\
    \hline
    Random & 58.44 & \textbf{\color{blue}{78.64}} \\
    ImageNet & \textbf{\color{blue}{58.45}} & 77.21 \\
    MillionAID & 49.44 & 70.73 \\
    HyperGlobal-450K & \textbf{\color{red}{63.92}} & \textbf{\color{red}{82.00}} \\
    \hline
    \end{tabular}
    \label{tab: pretraining weights}
\end{table}

\begin{table*}[h]
\centering
\newcommand{\tabincell}[2]{\begin{tabular}{@{}#1@{}}#2\end{tabular}}
\caption{Performance (\%) of various methods across different HSI classification datasets. \textbf{\color{red}{Best}} and \textbf{\color{blue}{2nd-best}} results are highlighted.}
\label{tab:hsi_classification_appendix}
\resizebox{\linewidth}{!}{
\begin{tabular}{l|ccccccccc|cc}
\hline
Metric & MSDN \cite{MSDN} & SSFCN\cite{ssfcn} & FullyContNet\cite{fullycontnet} & SpectralFormer\cite{spectralformer} & HSIC-FM\cite{HSIC-FM} & SSGRN\cite{ssgrn} & CSIL\cite{yang_csil} & IDCN\cite{idcn} & CLOLN\cite{CLOLN}  & SpatSIGMA & HyperSIGMA\\ 
\hline 
\bfseries \textit{Indian Pines} &   \multicolumn{11}{c}{}\\
\hline
OA & 57.54  &  41.93 &  71.11 & 50.02  & 36.02 & 69.58 &   66.53  &  71.12  & 72.75  &   \textbf{\color{blue}{85.08}} & \textbf{\color{red}{85.54}} \\
AA &   52.66 & 58.43  &  79.56 & 63.73 & 53.36 & \textbf{\color{red}{81.28}} & 80.94 &  \textbf{\color{blue}{80.99}} & 72.58  &  78.30 & 76.68 \\
Kappa & 51.81  & 35.65  & 67.77  &  44.14 & 30.33 & 65.98  & 62.71  & 67.40  & 69.05 &  \textbf{\color{blue}{83.04}}  & \textbf{\color{red}{83.58}} \\
\hline 
\bfseries \textit{Pavia University} &   \multicolumn{11}{c}{}\\
\hline
OA & 76.48  &  78.88  & 80.31  & 75.37  & 77.28 & 81.45 & 88.23  & 91.64  & 93.11  &  \textbf{\color{blue}{93.36}} &\textbf{\color{red}{93.52}}\\
AA & 75.18  & 82.28  &  86.85 & 80.88 & 77.09 & 85.35  & 90.88  & \textbf{\color{red}{93.19}}   & \textbf{\color{blue}{91.94}} & 88.82 & 87.78 \\
Kappa   & 69.52  &  72.99 & 75.01  & 68.56  & 70.36 & 76.36 & 84.76  & 89.05   &\textbf{\color{blue}{90.88}} & 90.66  & \textbf{\color{red}{90.93}} \\
\hline 
\bfseries \textit{HanChuan} &   \multicolumn{11}{c}{} \\
\hline
OA &  73.40 & 63.35  & 78.80  & 82.60 & 66.21 & 90.43 & 88.55  & 84.15  & 86.73 &  \textbf{\color{blue}{94.03}} & \textbf{\color{red}{94.44}} \\
AA & 67.43  & 61.50  &  76.71 & 79.58 & 63.31 & \textbf{\color{red}{92.08}} &  \textbf{\color{blue}{91.20}}  &  83.49 & 79.96  &  88.96  & 90.41 \\
Kappa    & 69.09  & 57.79  & 75.56  & 79.74  & 60.55 & 88.88 & 86.75  & 81.64 & 84.56 &  \textbf{\color{blue}{93.16}} & \textbf{\color{red}{93.62}}  \\
\hline 
\bfseries \textit{HongHu} &   \multicolumn{11}{c}{} \\
\hline
OA & 78.55  & 71.62  & 67.12 & 85.33 & 70.47 & 82.19 & 91.86  &  89.19 & 87.89 &  \textbf{\color{blue}{94.35}} & \textbf{\color{red}{94.87}} \\
AA & 73.06  & 66.51  &  73.81 & 83.20 & 71.03 & 87.31 & \textbf{\color{red}{94.64}}  &\textbf{\color{blue}{88.15}}  &  78.25 & 88.10 &  87.18 \\
Kappa    &  73.45 &  65.67 & 61.34 &  81.76 & 64.73 & 78.21 &  89.89  &86.59 &  84.83  & \textbf{\color{blue}{93.04}}  & \textbf{\color{red}{93.69}} \\
\hline
\bfseries \textit{Houston} &   \multicolumn{11}{c}{} \\
\hline
OA & 72.18 & 72.39  & 51.07 & 77.21 & 54.43 & 68.62 &  66.11 & 85.34 & 85.95  & \textbf{\color{red}{87.33}} & \textbf{\color{blue}{86.80}} \\
AA & 74.78 &  72.23 & 55.77 & 79.00  & 53.30 & 70.84 &  68.96  & 86.94 & \textbf{\color{red}{88.86}}  &  \textbf{\color{blue}{88.74}} & 88.27 \\
Kappa & 69.93  &  70.12 & 47.00 & 75.32 & 50.80 & 65.99 & 63.21  & 84.11 & 84.78  &  \textbf{\color{red}{86.30}} & \textbf{\color{blue}{85.72}} \\
\hline
\bfseries \textit{ZY1-02D} &   \multicolumn{11}{c}{} \\
\hline
OA & 88.57 &  82.46 & 84.47 & 72.03 & 76.98 & 77.46 &  92.49 & 92.29 & 80.43 & \textbf{\color{blue}{94.72}} & \textbf{\color{red}{94.92}} \\
AA & 88.13 &  80.61 & 83.69 & 63.06  & 74.75 & 82.28 & 91.31 & 92.89 & 73.88 & \textbf{\color{red}{95.70}} & \textbf{\color{blue}{93.95}} \\
Kappa & 86.47  & 79.36  & 81.65 & 67.20  &  73.31 & 73.80  & 91.09 & 90.82 & 76.91 & \textbf{\color{blue}{93.74}} & \textbf{\color{red}{95.15}} \\
\hline
\end{tabular}
}
\end{table*}

\begin{figure*}[t]
  \centering
  \includegraphics[width=0.9\linewidth]{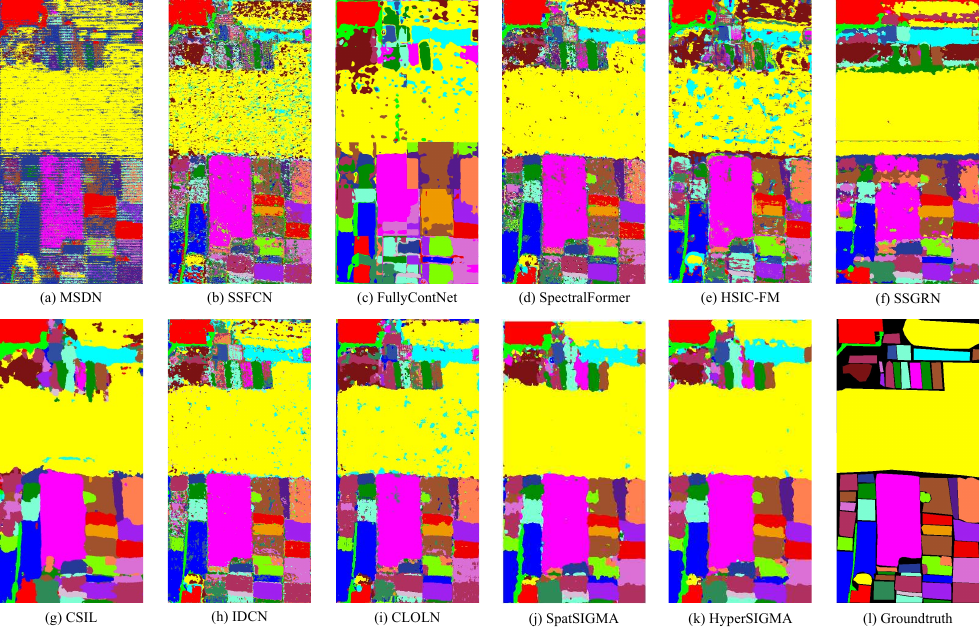}
  \caption{Predicted classification maps from various methods on the HongHu Wdataset \cite{whu_hi1}.}
\label{fig:HH_ClassificationMap}
\end{figure*}

\subsection{Implementation Details\label{subsec_cls_setting}}

\begin{figure*}[t]
    \centering
    \begin{minipage}{\textwidth}
        \centering
  \includegraphics[width=0.9\linewidth]{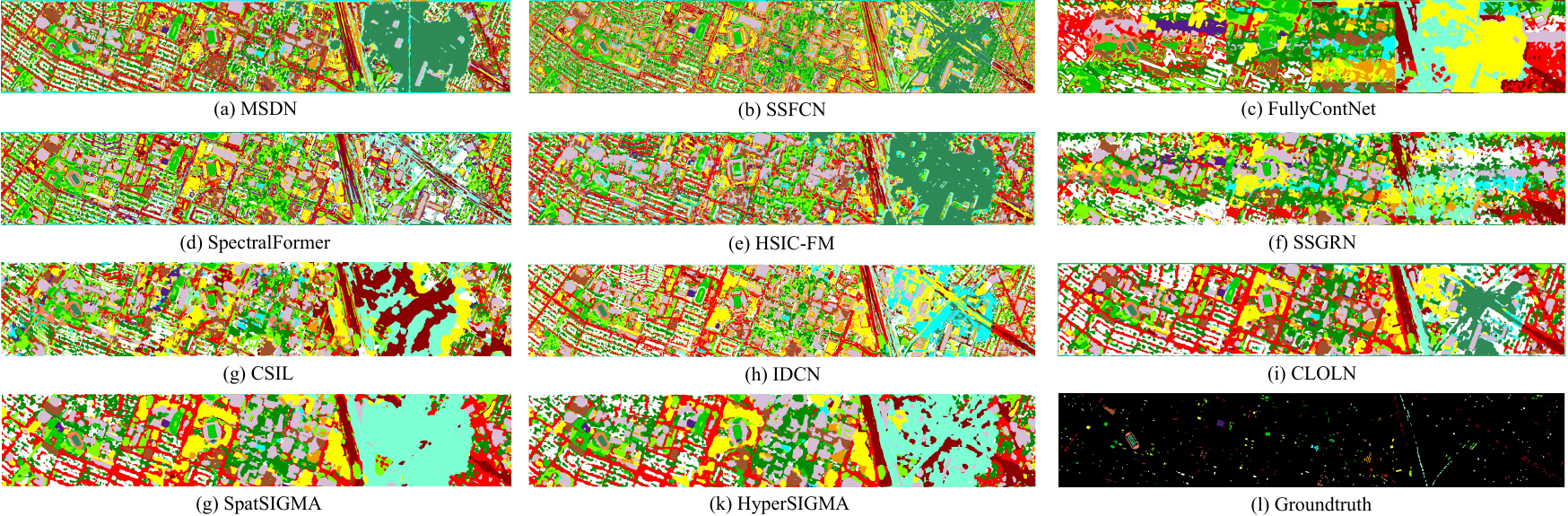}
  \caption{Predicted classification maps from various methods on the Houston dataset\cite{houston}}
\label{fig:HU_ClassificationMap}
    \end{minipage}
    \hfill
    \vspace{0.5em}
    \begin{minipage}{\textwidth}
        \centering
  \includegraphics[width=0.9\linewidth]{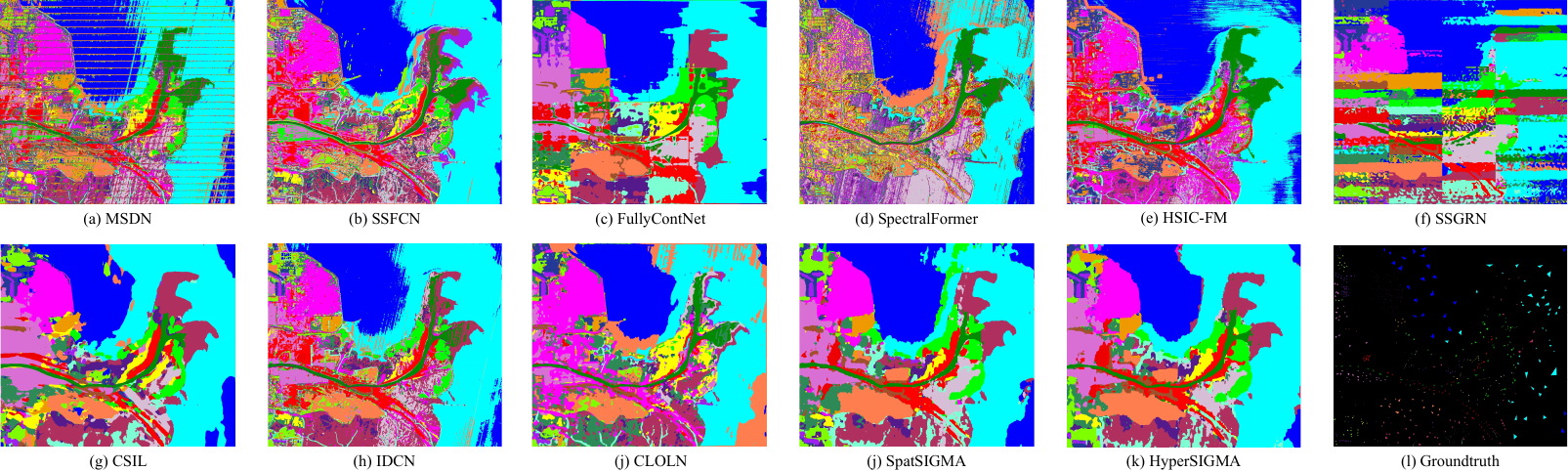}
  \caption{Predicted classification maps from various methods on the ZY1-O2D dataset \cite{huanghe}.}
\label{fig:ZY_ClassificationMap}
    \end{minipage}
\end{figure*}

\noindent\textbf{Network Structure}  Fig. \ref{fig: HSIClassificationFlowchart} illustrates the application of HyperSIGMA to this task. Spatial and spectral tokens are generated from the patch embedding layer. After processing through HyperSIGMA, four spatial feature maps $\bm{F}_i$ from the $i\cdot n/4$th layers ($i=1,\cdots,4$) are separately upsampled to the input size and enhanced by the final spectral feature $\bm{V}$ via four distinct SEM modules. The output features from the SEM modules are then concatenated and processed by fully convolutional layers.

\noindent\textbf{Experimental Details} In the experiments, 33$\times$33 patches are used for patch-level classification. For patch-level segmentation, besides 32$\times$32 patches and a spatial patch size of 2 are adopted for the Pavia University dataset with the consideration of tiny objects, other datasets apply 128$\times$128 patches, where the spatial patch size is set to 2 for Houston and ZY datasets, and 8 for other datasets. For the Indian Pines, Pavia University, HanChuan, HongHu, and ZY1-02D datasets, 10, 20, 50, 50, and 20 labeled samples per class are randomly selected as the training set, the remaining labeled samples form the test set, while the Houston dataset follows the official train/test split. The model is trained for 500 epochs with a learning rate of 0.00006, and the number of tokens in the spectral branch is set to 100 ($N_{spec}=100$). 

\subsection{Ablation Studies}

 \noindent\textbf{Network Structure} We slightly modify the structure from Fig. \ref{fig: HSIClassificationFlowchart}. Only the features ($\bm{F}_4$) from the last layer are used directly for classification, omitting any subsequent layers.

\noindent\textbf{Number of Sampling Points (NSP)} NSP is a critical hyperparameter in SSA, significantly influencing contextual feature extraction. We performed experiments to analyze its impact, as shown in Table \ref{tab: numbers of sampling points}. Here, ``FA'' denotes the use of full attention across all SpatSIGMA layers, while ``SSA'' indicates replacing some FAs with SSA. SSA helps reduce data redundancy and generally enhances classification performance compared to FA, thereby validating the efficacy of our approach. Based on the results, we selected 8 sampling points for subsequent experiments.

\begin{figure*}[t]
    \centering
    \begin{minipage}{0.48\textwidth}
        \centering
        \includegraphics[width=\linewidth]{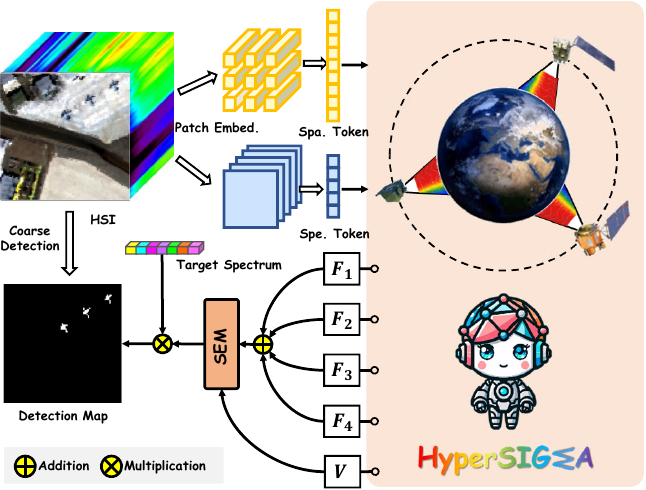} 
  \caption{Diagram illustrating the application of HyperSIGMA for hyperspectral target detection.}
\label{fig: htd_framework}
    \end{minipage}
    \hfill
    \begin{minipage}{0.48\textwidth}
        \centering
        \includegraphics[width=\linewidth]{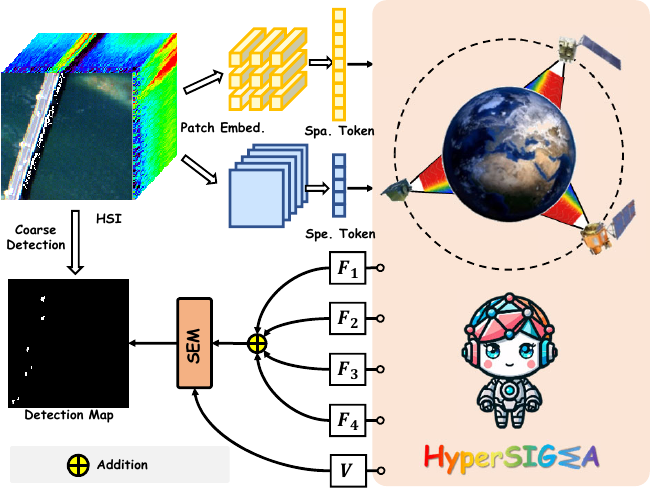} 
  \caption{Diagram illustrating the application of HyperSIGMA for hyperspectral anomaly detection.}
\label{fig: had_framework}
    \end{minipage}
\end{figure*}

\begin{figure*}[t]
  \centering
  \includegraphics[width=0.9\linewidth]{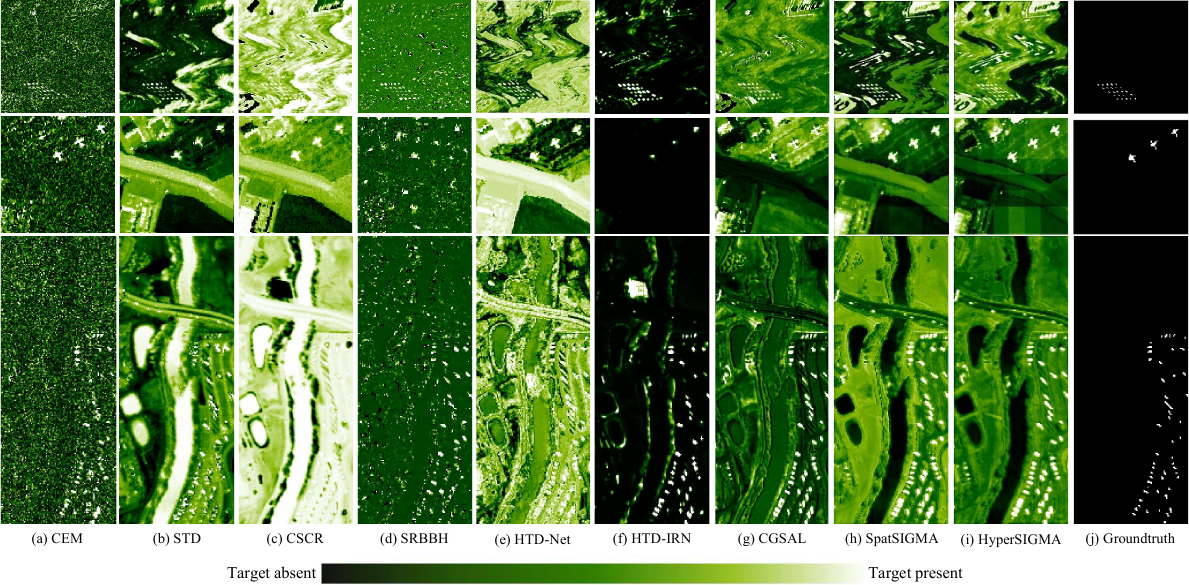}
  \caption{Detection probability maps predicted by various methods for the HTD tasks. Top row: Mosaic \cite{htd_mosaic}. Middle row: AVIRIS\cite{htd_aviris}. Bottom row: Renourban \cite{htd_renourban}.}
\label{fig: htd_map}
\end{figure*}

\noindent\textbf{Comparison of Different Attentions} To further validate the superiority of SSA, we conducted a comparative analysis by replacing SSA with other attention mechanisms: VSA \cite{vsa}, RVSA \cite{rvsa}, NLSA\cite{nlsa}, WMHSA \cite{swint}, and DMHA \cite{dat}. VSA, RVSA, and WMHSA were evaluated with window sizes of 4 and 8, NLSA separately adopted 4 and 8 attention buckets, while DMHA used downsampling rates of 4 and 8. Table \ref{tab: different attentions} summarizes the experimental results in terms of OA. SSA consistently achieved superior classification results. RVSA performed well, benefiting from its design tailored for RS overhead views. NLSA also obtained competitive accuracies by capturing locality-sensitive contexts. In contrast, DMHA underperformed due to its limited focus on hyperspectral objects within the global deformable region.

\noindent\textbf{Comparison of Pre-training Weight Strategies} Next, we compare various strategies for initializing model weights, including random initialization, pre-training on ImageNet \cite{imagenet,rsp}, MillionAID \cite{millionaid,rsp}, and HyperGlobal-450K. Results are summarized in Table \ref{tab: pretraining weights}. Loading weights pre-trained on different datasets occasionally led to performance degradation, likely due to domain gaps. Interestingly, pre-training with MillionAID, despite also being RS images, yielded lower accuracies, underscoring the uniqueness of hyperspectral data. Conversely, pre-training with HyperGlobal-450K significantly improved accuracy, highlighting its importance in HSI interpretation.

\subsection{More Results}

\noindent\textbf{Additional Metrics} Besides OA, we provide more metrics of the classification result, including average accuracy (AA) and the Kappa coefficient (Kappa), as shown in Table \ref{tab:hsi_classification_appendix}. It can be observed that our models show limitations in achieving high AA on the first five datasets, particularly in classes with fewer pixels, suggesting challenges in capturing details of small objects due to large patch sizes. 

\noindent\textbf{Visualization} Furthermore, qualitative evaluations through visualization of classification maps (Fig.~\ref{fig:IP_ClassificationMap}-\ref{fig:ZY_ClassificationMap}) for the Indian Pines, Pavia University, HanChuan, HongHu, Houston and ZY1-02D datasets, demonstrate that our models effectively mitigate issues such as salt and pepper noise, over-smoothing, and misclassifications observed in comparative methods. These visualizations reveal clear and discriminative boundaries, highlighting the effectiveness of spectral enhancement in HyperSIGMA for refining boundaries. 

\section{Hyperspectral Target and Anomaly Detection \label{sec_hyper_det}}

\begin{figure*}[h]
  \centering
  \includegraphics[width=0.9\linewidth]{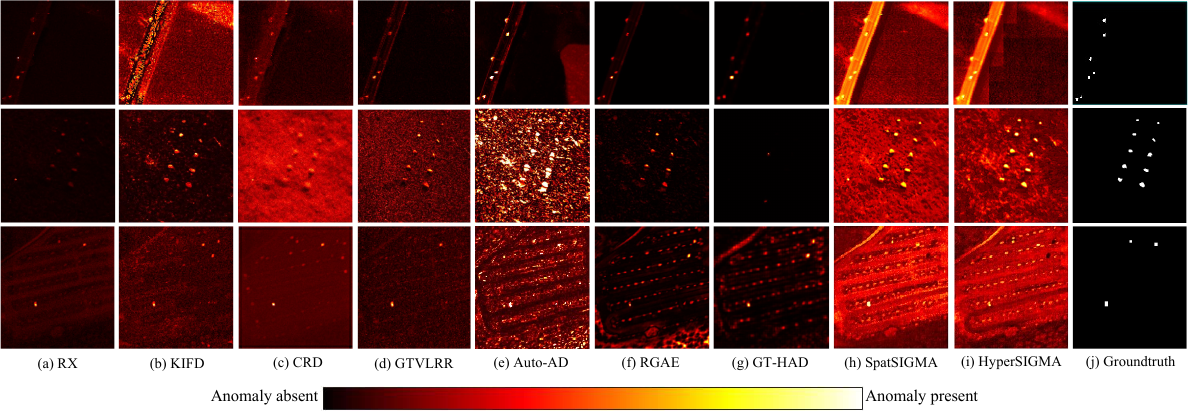}
  \caption{Detection probability maps predicted by various methods for the HAD tasks. Top row: Pavia\cite{had_pavia}. Middle row: Cri\cite{had_cri}. Bottom row: Viareggio \cite{had_viareggio}.}
\label{fig: had_map}
\vspace{-1em}
\end{figure*}

\begin{figure*}[t]
  \centering
  \includegraphics[width=0.9\linewidth]{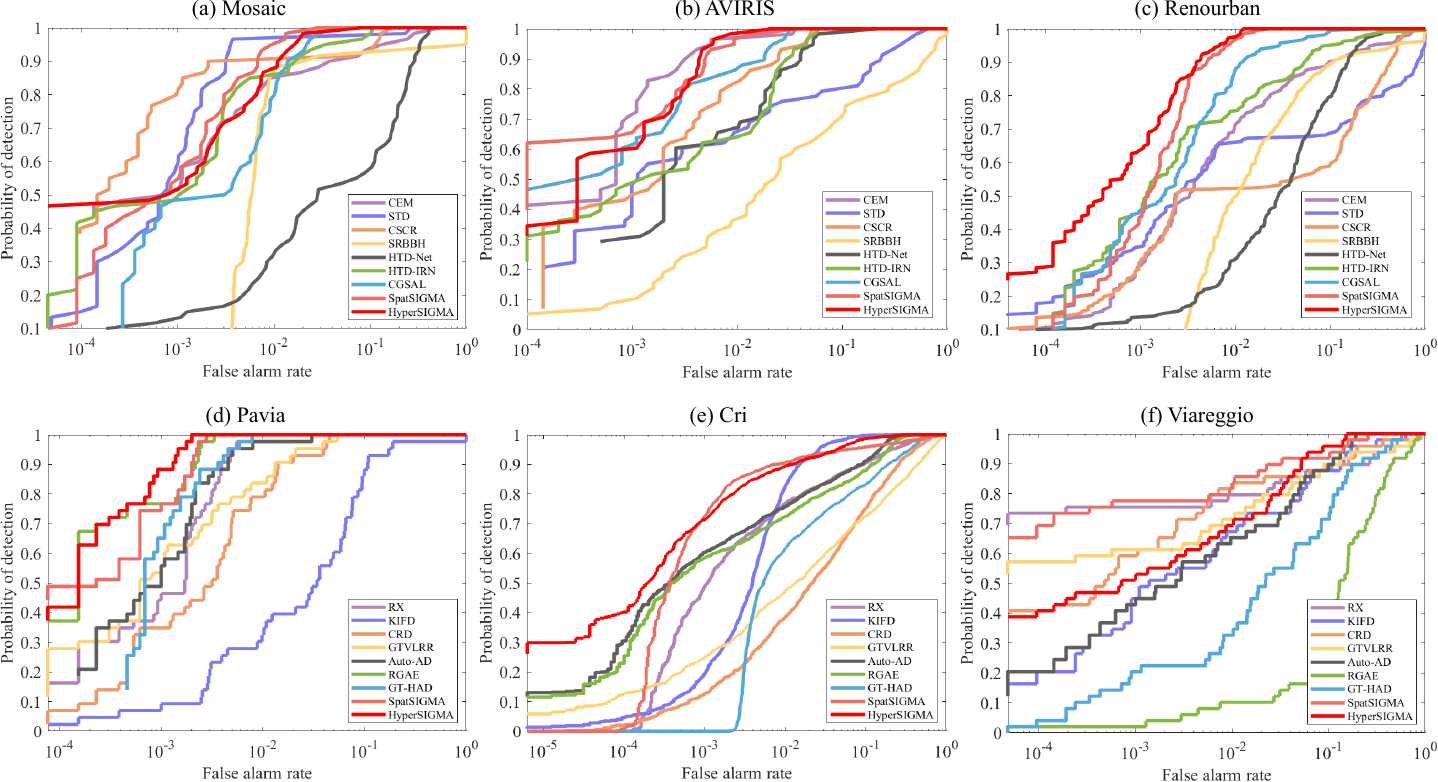}
  \caption{ROC curves for various methods applied to the HTD and HAD tasks.}
\label{fig: hd_roc_curves}
\end{figure*}

\subsection{Implementation Details}

\noindent\textbf{Network Framework} Specifically, we start by using CEM \cite{cem} for coarse detection \cite{htd_vit}. Pixels with the highest 1.5\% probabilities are labeled as pseudo detections, while the bottom 30\% are set as the background by following \cite{htd_vit}. The remaining pixels are ignored during segmentation loss calculation. Fig. \ref{fig: htd_framework} illustrates the entire network framework. For segmentation, a linear layer matches the target spectrum $\mathbf{T}$'s dimension to $\mathbf{F}^{\star}$, i.e., from $\mathbb{R}^{C}$ ($C$ is the number of HSI channels) to $\mathbb{R}^{D_1}$. We then compute the similarities between each pixel of $\mathbf{F}^{\star}$ and $\mathbf{T}$:
\begin{equation}
  \mathbf{S} = \mathbf{F}^{\star}\mathbf{T}. 
\end{equation}
$\mathbf{S} \in \mathbb{R}^{H' \times W' \times 1}$ denotes the predicted target probability map. To calculate the segmentation losses for the background and target classes, we construct a segmentation probability matrix $\mathbf{S}_M \in \mathbb{R}^{H' \times W' \times 2}$ by concatenating $1-\mathbf{S}$ and $\mathbf{S}$ along the channel dimension. We adopt the cross-entropy loss. Additionally, prior to SEM, we utilize multiscale spatial features $\bm{F}_i, i=1,2,3,4$ from the $i \cdot (n/4)$th layers to enhance the discrimination of the generated $\mathbf{F}_{spat}$.

Given the similarities between HAD and HTD tasks, we use a similar network for HAD implementation (Fig. \ref{fig: had_framework}). The primary difference is in generating coarse detection labels, where we utilize the classical HAD method RX \cite{rx} and the highest 0.15\% probabilities are labeled as pseudo detections. Since HAD does not involve the target spectrum, $\mathbf{F}^{\star}$ is used for segmentation after passing through a 1$\times$1 convolution. 

\noindent\textbf{Experimental Hyperparameters} For segmentation, HSIs are divided into 32$\times$32 patches with an overlap of 16, given that the SpatViT's patch size is 1. The patch embedding layers' weights are trained from scratch, with $H'=H$ and $W'=W$. For the SpecViT, $N_{spec}$ is set to 100. The networks are trained for 10 epochs using AdamW \cite{adamw} with a learning rate of 0.00006. The batch size is set to 2 for HTD and 1 for HAD.

\subsection{More Results}

We qualitatively visualize the detection results of different methods. As shown in Fig. \ref{fig: htd_map}-\ref{fig: had_map}, our models outperform others by effectively separating targets from backgrounds, accurately identifying the entire target areas, and assigning high detection confidences to these regions. However, we notice that some unrelated object regions in our results have high probabilities, especially in the HTD task. This is likely due to inaccurate pseudo labels from coarse detection, which mislead the model. Our results also exhibit traces of grid splicing due to sliding window-based inference. This effect is more pronounced in HyperSIGMA than in SpatSIGMA. We hypothesize that, while fusing spectral information enhances representational ability, the resulting channel changes may increase the instability of spatial features.

\begin{table}[t]
    \centering
    \caption{Change detection accuracy (\%) of various methods on the Hermiston dataset. \textbf{\color{red}{Best}} and \textbf{\color{blue}{2nd-best}} results are highlighted.}
    \begin{tabular}{lccccc}
    \hline
        Method & OA & Kappa & F1 & Precision & Recall \\ \hline
        CVA \cite{hsi_cd_cva} & 92.72 & 76.70 & 81.03 & \textbf{\color{blue}{98.19}} & 68.98 \\ 
        ISFA \cite{hsi_cd_isfa} & 90.23 & 67.16 & 72.62 & \textbf{\color{red}{98.52}} & 57.50  \\ 
        GETNET \cite{hsi_cd_getnet} & 95.46  & 87.56  & 90.54 & 85.36 & 96.38    \\ 
        ML-EDAN \cite{hsi_cd_mledan} & 94.71 & 85.57 & 89.03 & 83.56 & 95.29  \\ 
        BIT \cite{hsi_cd_bit} & 83.92 & 60.70 & 71.28 & 59.69  & 88.50\\ 
        EMS-Net \cite{hsi_cd_emsnet} & 92.93 & 81.05 & 85.67& 80.66 & 92.04  \\
        CSA-Net \cite{hsi_cd_csanet} & 92.85 & 80.60 & 85.29 & 79.54 & 91.95 \\ 
        SST-Former\cite{hsi_cd_sstformer}  & 94.49 & 85.18 & 88.80 & 81.90 & 96.97  \\ 
        GlobalMind\cite{hsi_cd_globalmind}  & 95.56 & 87.81  & 90.71 & 85.82 & 96.20 \\ 
        \hline
        SpatSIGMA & \textbf{\color{red}{96.24}}  & \textbf{\color{red}{89.62}}  & \textbf{\color{red}{92.08}}  & 87.53  & \textbf{\color{blue}{97.13}}  \\ 
        HyperSIGMA & \textbf{\color{blue}{96.06}}  & \textbf{\color{blue}{89.17}}   & \textbf{\color{blue}{91.74}}  & 86.88  & \textbf{\color{red}{97.19}} \\ \hline
    \end{tabular}
    \label{table:HERMISTON}
\end{table}

To comprehensively compare detection performance, we present the ROC curves of various methods in Fig. \ref{fig: hd_roc_curves}. These results show that our models remain competitive across different false alarm rates (FAR). In the HTD task, HyperSIGMA excels at higher FARs, effectively identifying most target pixels. For the HAD task, HyperSIGMA consistently outperforms other methods across nearly all FARs. Overall, these findings demonstrate HyperSIGMA's effectiveness and versatility in hyperspectral detection tasks.

\section{Hyperspectral Change Detection \label{sec_hyper_cd}}

\subsection{Implementation Details}

\begin{table}[t]
    \centering
    \caption{Change detection accuracy (\%) of various methods on the Farmland dataset. \textbf{\color{red}{Best}} and \textbf{\color{blue}{2nd-best}} results are highlighted.}
    \begin{tabular}{lccccc}
    \hline
        Method & OA & Kappa & F1 & Precision & Recall  \\ \hline
        CVA \cite{hsi_cd_cva} & 95.48 & 89.27 & 92.49& 89.23 & 96.00   \\ 
        ISFA \cite{hsi_cd_isfa} & 95.75 & 89.96 & 93.01 & 88.89 & 97.52   \\ 
        GETNET \cite{hsi_cd_getnet} & 96.58 & 91.93 & 94.39 & \textbf{\color{red}{99.06}} & 90.15  \\ 
        ML-EDAN \cite{hsi_cd_mledan} & 96.35 & 91.32 & 93.93 & 90.81& 97.26   \\ 
        BIT \cite{hsi_cd_bit} & 90.21 & 77.46 & 84.56 & 92.38 & 77.96   \\ 
        EMS-Net \cite{hsi_cd_emsnet} & 95.68 & 89.86 & 92.97 & 88.07 & \textbf{\color{blue}{98.45}}  \\ 

        CSA-Net \cite{hsi_cd_csanet}  & 95.92 & 90.34 & 93.27 & 89.47 & 97.42   \\ 
        SST-Former \cite{hsi_cd_sstformer} & 96.59 & 91.91 & 94.34 & 91.02 & 97.92   \\ 
        GlobalMind \cite{hsi_cd_globalmind} & 97.00 & 92.84 & 94.98 & 92.19 & 97.95    \\ 
        \hline
        SpatSIGMA & \textbf{\color{blue}{97.19}} & \textbf{\color{blue}{93.30}} & \textbf{\color{blue}{95.31}} & 92.25 & \textbf{\color{red}{98.59}}  \\ 
        HyperSIGMA & \textbf{\color{red}{97.30}}  & \textbf{\color{red}{93.56}}  & \textbf{\color{red}{95.49}}  & \textbf{\color{blue}{92.76}} & 98.37  \\ \hline
    \end{tabular}
    \label{table:FARMLAND}
\end{table}

\begin{figure*}[tb]
\centering
\includegraphics[width=0.9\linewidth]{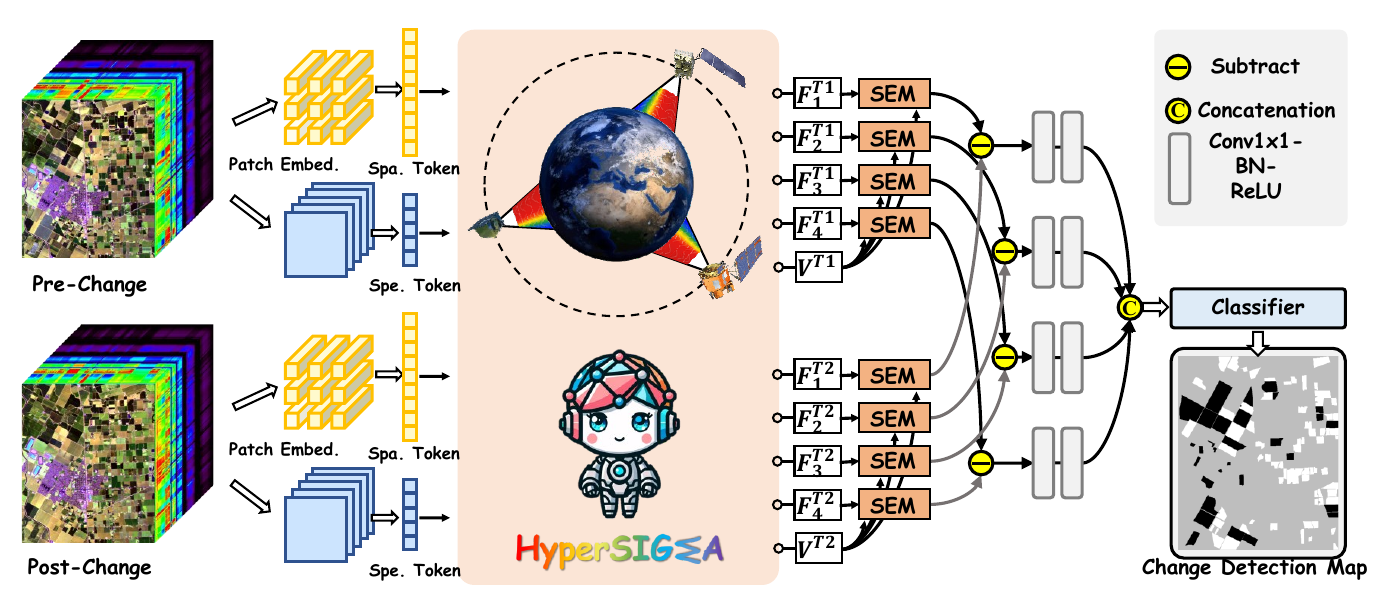}
\caption{Diagram illustrating the application of HyperSIGMA for hyperspectral change detection.}
\label{fig: CD_FlowChart}
\end{figure*}

\noindent\textbf{Network Structure} The HyperSIGMA model's architecture for this task is depicted in Fig.~\ref{fig: CD_FlowChart}. Firstly, we merge four spectral features $\bm{V}_i$ from layers $i\cdot (n/4), i=1,2,3,4$ of the spectral subnetwork, obtaining $\bm{V}$, i.e., $\sum_{i=1}^4 \bm{V}_i = \bm{V}$. Then, features $\bm{F}_i$ from layers $i\cdot (n/4), i=1,2,3,4$ of the spatial branch are separately combined with $\bm{V}$ through four SEMs. Afterward, the spatiotemporal fusion of two temporal phases is achieved by computing subtraction results, which are then passed through two convolutional layers for information integration. The final change detection result is produced using a classifier on the concatenated features.

\begin{table}[t]
    \centering
    \caption{Change detection accuracy (\%) of various methods on the BayArea dataset. \textbf{\color{red}{Best}} and \textbf{\color{blue}{2nd-best}} results are highlighted.}
    \begin{tabular}{lccccc}
    \hline
        Method & OA & Kappa & F1 & Precision & Recall  \\ \hline
        CVA \cite{hsi_cd_cva} & 87.23 & 74.62 & 87.09 & 94.74 & 80.57   \\ 
        ISFA \cite{hsi_cd_isfa} & 89.17 & 78.48 & 89.05 & 96.95 & 82.34  \\ 
        GETNET  \cite{hsi_cd_getnet} & 95.87 & 91.73 & 96.07 & 97.86 & 94.35   \\ 
        ML-EDAN \cite{hsi_cd_mledan} & 97.24 & 94.47 & 97.41 & 97.97 & 96.85   \\ 
        BIT \cite{hsi_cd_bit} & 94.37 & 88.69 & 94.69 & 95.45 & 93.96   \\ 
        EMS-Net \cite{hsi_cd_emsnet} & 97.54 & 95.06 & 97.67 & 96.86 & 98.52   \\ 
        CSA-Net   \cite{hsi_cd_csanet} & 97.68 & 95.34 & 97.80 & \textbf{\color{blue}{98.75}} & 96.88  \\ 
        SST-Former \cite{hsi_cd_sstformer} & 97.06 & 94.10 & 97.23& 97.87 & 96.60   \\ 
        GlobalMind \cite{hsi_cd_globalmind} & 98.15 & 96.29 & 98.26& \textbf{\color{red}{98.94 }} & 97.58   \\ 
        \hline
        SpatSIGMA & \textbf{\color{blue}{98.81}} & \textbf{\color{blue}{97.60}} & \textbf{\color{blue}{98.89}} & 98.57 &  \textbf{\color{red}{99.20}}  \\ 
        HyperSIGMA &  \textbf{\color{red}{98.85}} &  \textbf{\color{red}{97.70}} &  \textbf{\color{red}{98.93}} &  98.66& \textbf{\color{blue}{99.21}}  \\ \hline
    \end{tabular}
    \label{table:BAY}
\end{table}

\noindent\textbf{Experimental Hyperparameters} We select 5$\times$5 patches as input when training on Hermiston and Farmland, and 15$\times$15 patches for BayArea and Barbara. For the spatial subnetwork, the patch size of embedding layers is 1 for Hermiston and Farmland, and 2 for BayArea and Barbara, to accommodate their greater variability.

Following the protocol of SST-Former \cite{hsi_cd_sstformer}, we randomly sampled 500 unchanged and 500 changed pixels from each dataset for training. To ensure a fair comparison, the training patches have a very low overlapping rate related to the testing set, aligning to \cite{hsi_cd_sstformer}. Our models were trained for 50 epochs using a learning rate of 0.00006 and a batch size of 32. The spectral branch utilized 144 tokens to capture detailed spectral information ($N_{spec} = 144$).

\begin{figure*}[!t]
\centering
\includegraphics[width=0.85\linewidth]{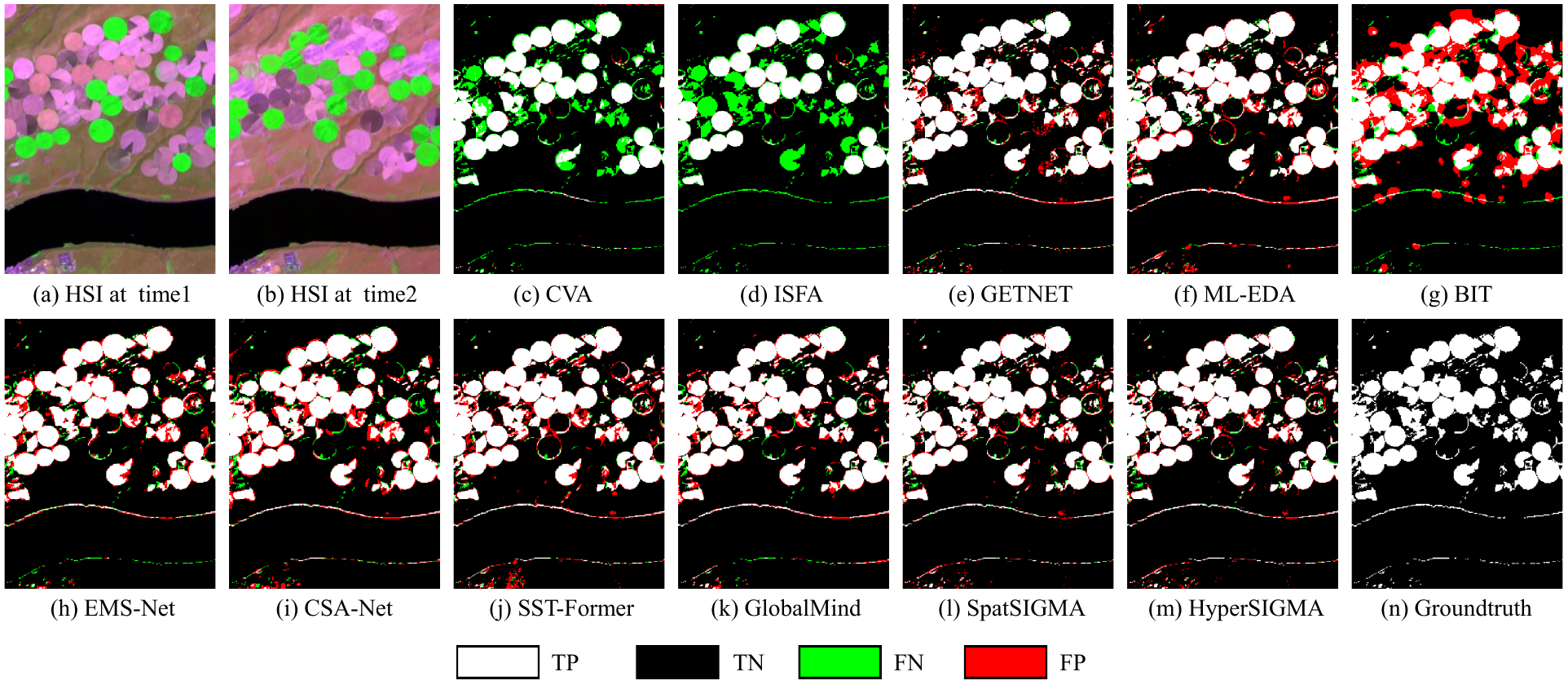}
\caption{Change detection maps from various methods on the Hermiston dataset.}
\label{fig: CD_Hermiston}
\end{figure*}

\begin{table}[t]
    \centering
    \caption{Change detection accuracy (\%) of various methods on the Santa Barbara dataset. \textbf{\color{red}{Best}} and \textbf{\color{blue}{2nd-best}} results are highlighted.}
    \begin{tabular}{lccccc}
    \hline
        Method & OA & Kappa & F1 & Precision & Recall  \\ \hline
        CVA \cite{hsi_cd_cva} & 87.80 & 74.03 & 83.76  & 87.92 & 79.97 \\ 
        ISFA  \cite{hsi_cd_isfa}, & 89.12 & 76.75 & 85.35 & 90.71 & 80.59  \\ 
        GETNET \cite{hsi_cd_getnet} & 97.35 & 94.43 & 96.59& 97.65 & 95.56   \\ 
        ML-EDAN \cite{hsi_cd_mledan} & 98.28 & 96.39 & 97.80 & 98.26 & 97.35  \\ 
        BIT \cite{hsi_cd_bit} & 96.93 & 93.61 & 96.19 & 94.08 & 98.41   \\ 
        EMS-Net  \cite{hsi_cd_emsnet} & 97.88 & 95.56 & 97.32 & 98.03 & 96.68  \\ 
        CSA-Net  \cite{hsi_cd_csanet} & 98.61 & 97.09 & 98.24 & 97.64& 98.86   \\ 
        SST-Former \cite{hsi_cd_sstformer} & 98.10 & 96.02 & 97.58 & 97.98  & 97.18 \\ 
        GlobalMind \cite{hsi_cd_globalmind} & 98.65 & 97.17 & 98.28  & 98.51 & 98.06 \\ 
        \hline
        SpatSIGMA & \textbf{\color{blue}{99.17}} & \textbf{\color{blue}{98.27}} & \textbf{\color{blue}{98.95}} & \textbf{\color{blue}{98.87}} & \textbf{\color{red}{99.04}}   \\ 
        HyperSIGMA & \textbf{\color{red}{99.25}}  & \textbf{\color{red}{98.43}}  & \textbf{\color{red}{99.04}}  & \textbf{\color{red}{99.17}}  & \textbf{\color{blue}{98.93}}\\
        \hline
    \end{tabular}
    \label{table:BARBARA}
\end{table}

\begin{figure*}[!t]
\centering
\includegraphics[width=0.85\linewidth]{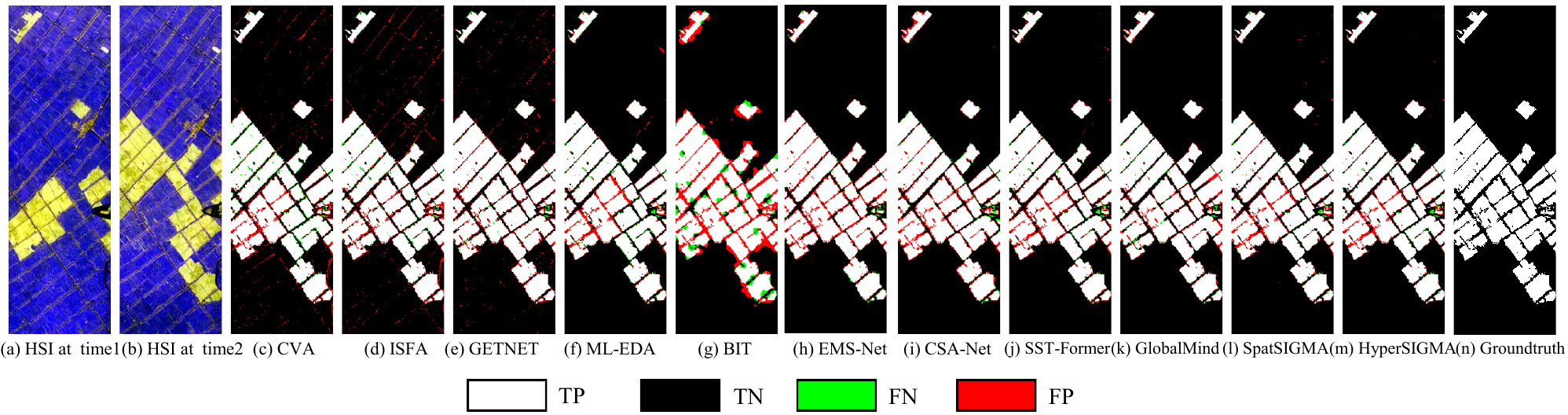}
\caption{
Change detection maps from various methods on the Farmland dataset.
}
\label{fig:CDmap_Farmland}
\end{figure*}

\begin{figure*}[!t]
\centering
\includegraphics[width=0.85\linewidth]{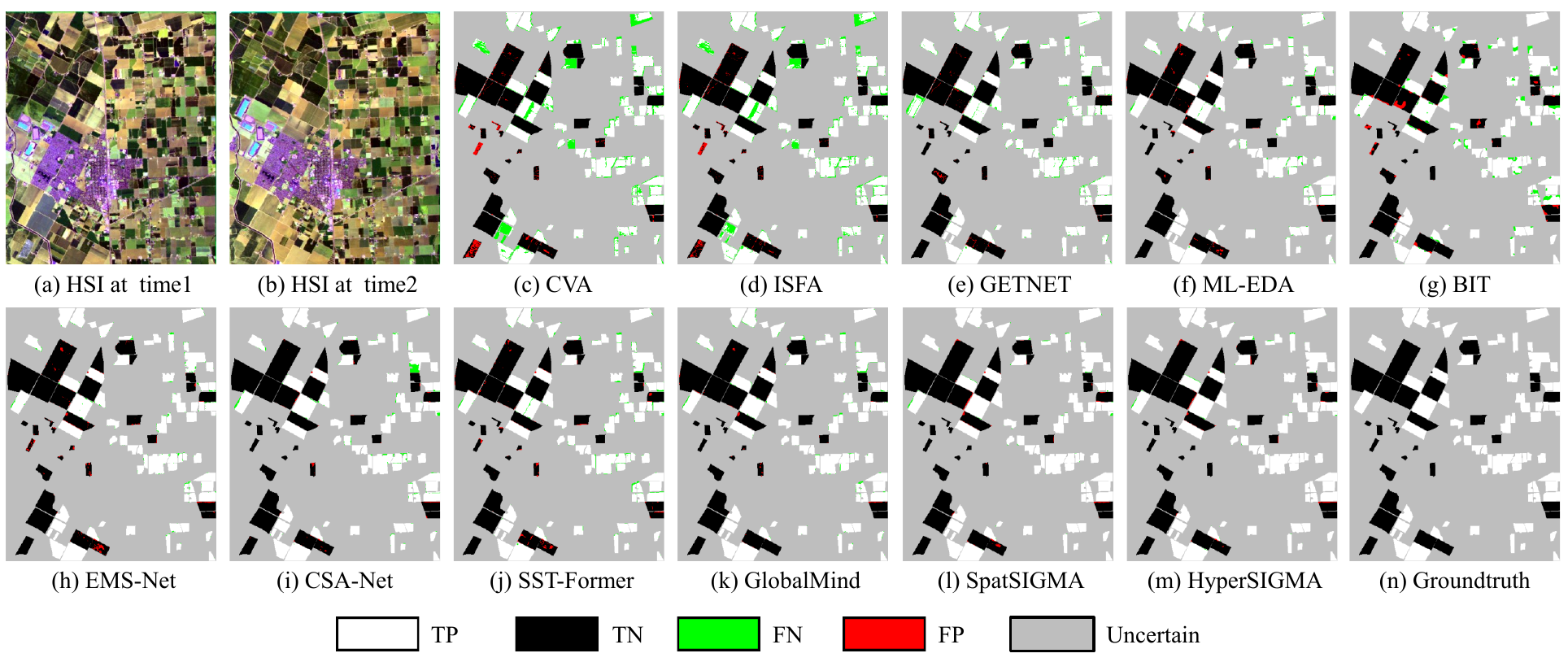}
\caption{Change detection maps from various methods on the Bay Area dataset.}
\label{fig: CD_BayArea}
\end{figure*}

\begin{figure*}[t]
\centering
\includegraphics[width=0.85\linewidth]{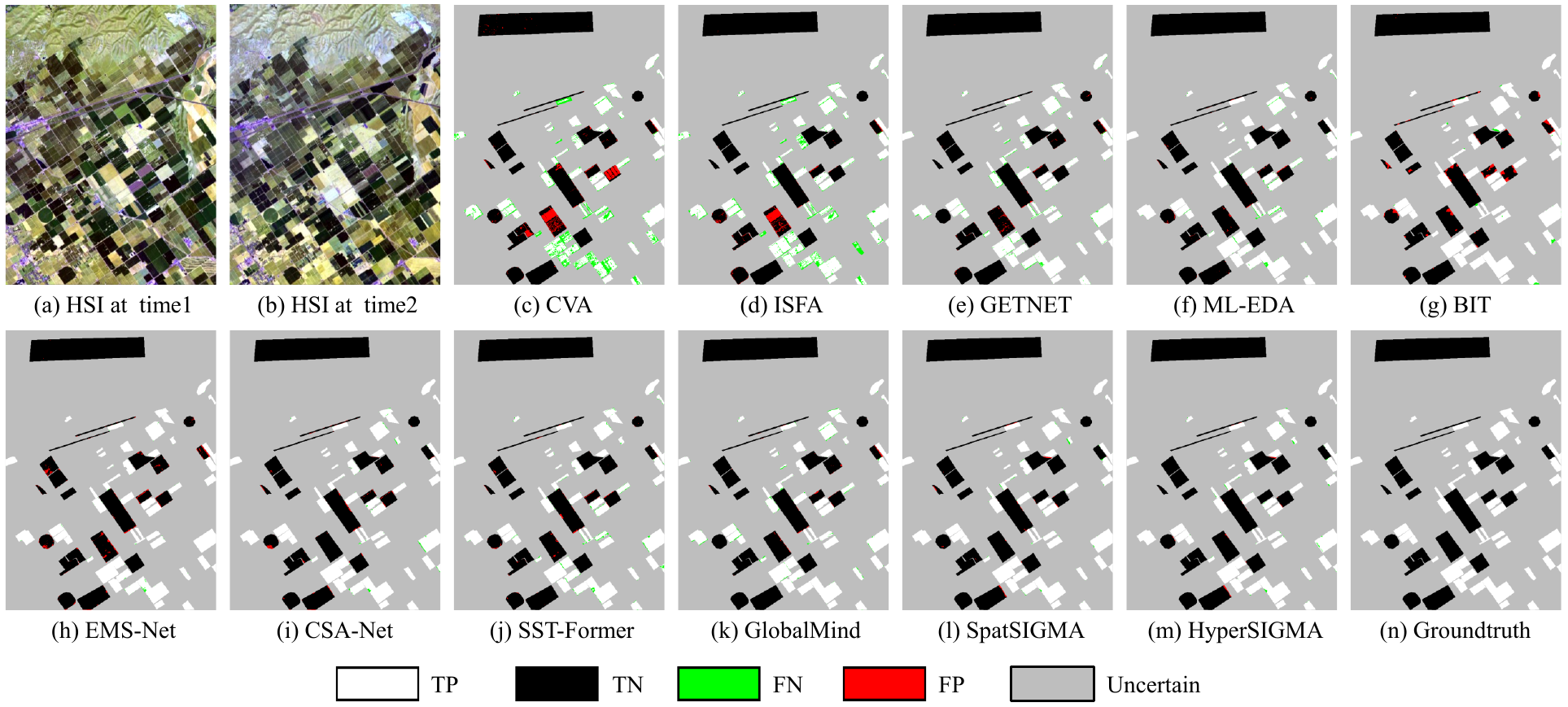}
\caption{Change detection maps from various methods on the Santa Barbara dataset.}
\label{fig: CD_Barbara}
\end{figure*}

\begin{table*}[t]
\centering
\caption{
Quantitative comparison of endmember and abundance prediction performance across various methods on the Urban dataset. \textbf{\color{red}{Best}} and \textbf{\color{blue}{2nd-best}} results are highlighted.
}
\label{hsi_unmixing_appendix}
\resizebox{\linewidth}{!}{
\begin{tabular}{l|ccccccc|cc}
\hline
Metric & FCLS \cite{fcls} & ELMM \cite{elmm} & SUnSAL \cite{sunsal} & CNNAEU \cite{cnnaeu} & CyCU \cite{cycu} & DeepTrans \cite{deeptrans} & EGUnet\cite{egunet}  & SpatSIGMA & HyperSIGMA\\ 

\hline 
\bfseries \textit{Abundance} &   \multicolumn{7}{c}{}\\
\hline
Asphalt &0.1918&0.0211&0.3910&0.0307&0.1029 &\textbf{\color{red}{0.0138}} &0.1690 &0.0152 & \textbf{\color{blue}{0.0146}} \\
Grass &0.2211&0.0873&0.1633 &\textbf{\color{red}{0.0286}} &0.1154&0.0872&0.1626 &0.0358 & \textbf{\color{blue}{0.0317}} \\
Tree&0.1155&0.0657&0.1563 & 0.0148 &0.0579&0.0860&0.2098 &\textbf{\color{blue}{0.0135}} &\textbf{\color{red}{0.0114}}  \\
Roof &0.0330&0.0229&0.0329&0.0126&0.0520&0.0083&0.1514 & \textbf{\color{blue}{0.0059}} &\textbf{\color{red}{0.0052}}  \\

\hline 
\bfseries \textit{Endmember} &   \multicolumn{7}{c}{}\\
\hline
Asphalt &0.2283& - & - &0.0809&0.2121&0.0820&0.0777 &\textbf{\color{red}{0.0633}}  &\textbf{\color{blue}{0.0679}}  \\
 Grass &0.3929&- & -&\textbf{\color{red}{0.0372}} &0.5107&0.3001&0.4443&0.1178 &\textbf{\color{blue}{0.1062}} \\
 Tree &0.3045&- & -&0.1482&0.1324&0.0863&0.5625  &\textbf{\color{blue}{0.0270}} &\textbf{\color{red}{0.0266}} \\
 Roof &0.8237&- & -&0.0795&0.3256&0.1362&1.2339  &\textbf{\color{red}{0.0312}} &\textbf{\color{blue}{0.0327}}\\

\hline
\end{tabular}
}
\end{table*}

\begin{figure*}[t]
\centering
\includegraphics[width=0.9\linewidth]{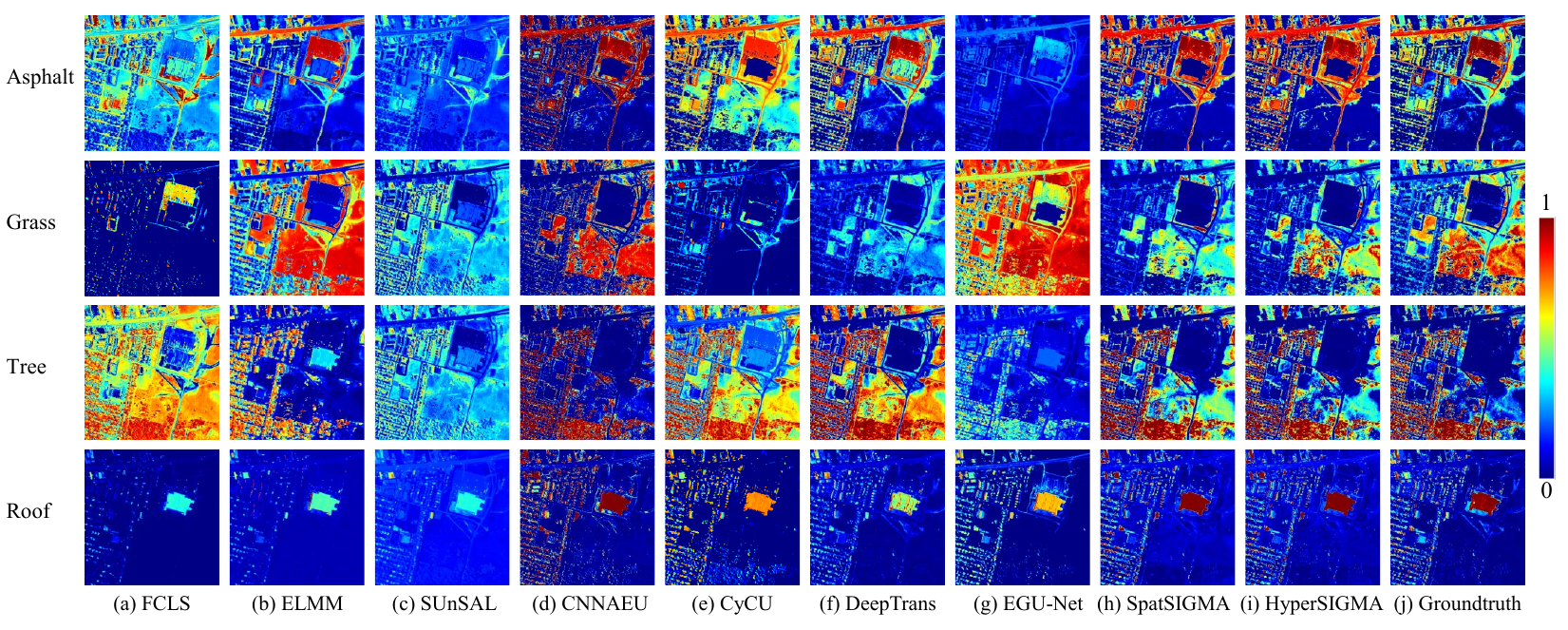}
\caption{Abundance prediction results of different methods for hyperspectral unmixing on the Urban dataset.}
\label{fig: unmixing_Urban_abundance}
\end{figure*}

\begin{figure}[t]
\centering
\includegraphics[width=\linewidth]{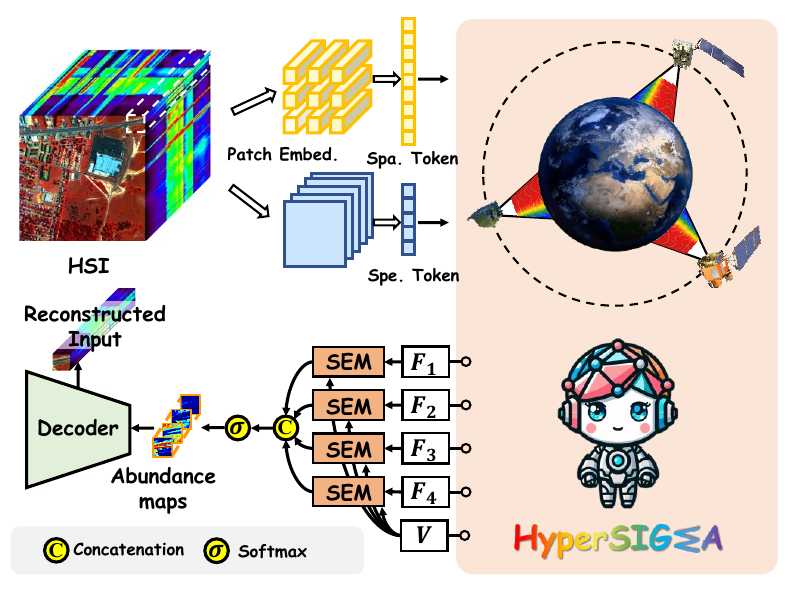}
\caption{Diagram illustrating the application of HyperSIGMA for hyperspectral unmixing.}
\label{fig: unmixing_flowchart}
\end{figure}

\subsection{More Results}
\noindent\textbf{Additional Metrics} Tables \ref{table:HERMISTON}-\ref{table:BARBARA} present more comprehensive evaluation results beyond F1 score, encompassing OA, Kappa, precision, and recall. We observe that the precision of our models still requires improvement, as they may misclassify unchanged areas as changed. This issue might stem from using relatively small patches for change detection, as in \cite{hsi_cd_sstformer}. We speculate that using larger patches to increase context perception could mitigate this problem.

\noindent\textbf{Visualization} Fig.~\ref{fig: CD_Hermiston}-\ref{fig: CD_Barbara} show the change detection results for the Hermiston, Farmland, Bay Area, and Santa Barbara datasets, highlighting True Positives (white), True Negatives (black), False Positives (red), False Negatives (green), and Uncertain areas (gray). Our models detect most changes accurately with fewer false negatives compared to other methods, resulting in superior recall.

\section{Hyperspectral Unmixing \label{sec_hyper_unmixing}}

\subsection{Implementation Details} 

\noindent\textbf{Task Formulation} Assume the input and output of the network are $X\in \mathbb{R}^{H \times W \times C}$ and $\hat{X}\in \mathbb{R}^{H \times W \times C}$, we have
\begin{equation}
Z=f_{E}(X) \quad \hat{X} = f_{D}(Z),
\end{equation}
where $H,W,C$ are the height, width, and channel number of the input HSI, respectively. The low-dimensional feature $Z \in \mathbb{R}^{H' \times W' \times C_a}$ has $C_a$ channels, where typically $C_a \ll C$. $f_{E}$ and $f_{D}$ represent the encoder and decoder, respectively. Each HSI pixel is considered a weighted sum of endmembers (linear mixture model \cite{spectral_mixing}), with $\hat{X}$ representing the reconstructed spectrum. As described in \cite{cnnaeu}, the decoder $f_D$ is a single convolutional layer, specifically $\hat{X} = W^D(Z)$, where $W^D \in \mathbb{R}^{C_a \times C}$ is the weight matrix. In the unmixing task, $W^D$ is referred to as the endmember matrix and $Z$ as the abundance matrix, indicating that each HSI pixel contains $C_a$ endmembers.

We employ the spectral angular distance (SAD) loss \(L_{SAD}\) to quantify the reconstruction error between \(X\) and \(\hat{X}\). To ensure physical validity, we impose non-negativity constraints on both the endmembers and abundances, with the latter also constrained by a sum-to-one condition. To promote sparsity, we apply an \(L_{1/2}\) norm constraint on the abundances. Additionally, we introduce first-order total variation norm (TV-norm) regularization \(L_{TV}\) on the endmembers to achieve a smooth spectrum. The final loss function is formulated as follows:
\begin{equation}
J= \frac{1}{HW}\sum\limits_{i=1}^{HW}\left[L_{SAD}(x_i,\hat{x}_i)+\alpha\cdot L_{1/2}(z_i)\right]+\beta\cdot L_{TV}(W^D),
\end{equation}
where three losses are separately defined as:
\begin{equation}
  \begin{aligned}
L_{SAD}(x_i,\hat{x}_i)&=\arccos\left(\frac{x_i\hat{x}_i^T}{\|x_i\|_2\|\hat{x}_i\|_2}\right), \\
L_{1/2}(z_i)&=\sum_{j=1}^{C_a}z_{ij}^{1/2}, \\
L_{TV}(W^D)&=\sum_{j=1}^{C_a}\sum_{c=1}^{C-1}\left| W^D_{j,c+1}- W^D_{j,c}\right|,
  \end{aligned}
\end{equation}
where \( x_i, \hat{x}_i, z_i \) for \( i=1, \ldots, HW \) are the \( i \)-th elements of \( X, \hat{X}, Z \), respectively. Note that before being fed into \( f_D \), the size of \( Z \) is recovered by bilinear upsampling, i.e., \( H' \rightarrow H \) and \( W' \rightarrow W \).

\begin{figure}[h]
\centering
\includegraphics[width=\linewidth]{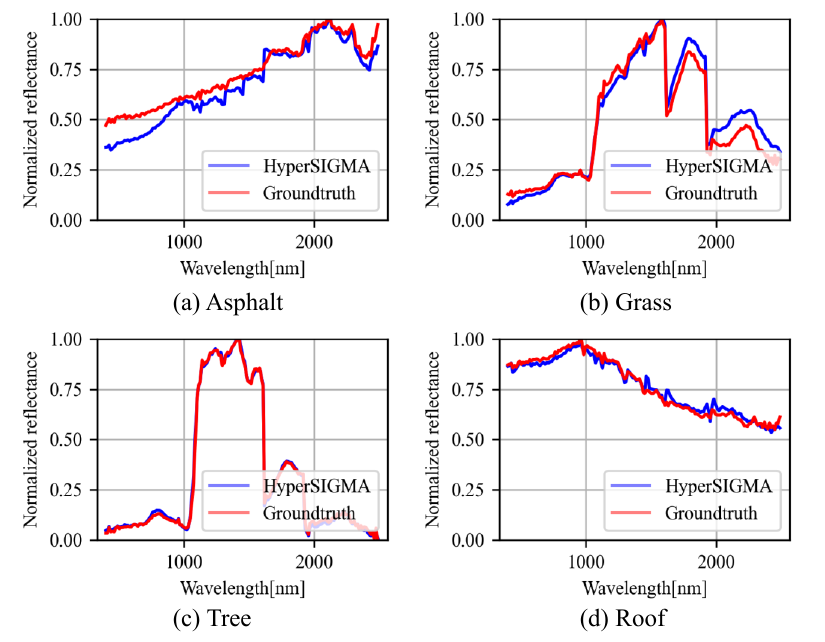}
\caption{Endmember prediction results of HyperSIGMA for hyperspectral unmixing on the Urban dataset.}
\label{fig: unmixing_Urban_endmember}
\end{figure}

\noindent\textbf{Fine-tuning Network Architecture} We use HyperSIGMA as the feature extractor, denoted as \(f_E\). The architecture for using HyperSIGMA for hyperspectral unmixing is shown in Fig. \ref{fig: unmixing_flowchart}. Similar to the classification task, to comprehensively understand HSI scenes, we utilize four feature maps from the spatial branch, \(\bm{F}_i\) (where \(i=1,2,3,4\)), and the final layer of the spectral branch, \(\bm{V}\). These maps are fused using different SEM modules and then concatenated for information aggregation. The aggregated features pass through 1$\times$1 convolutions to reduce the dimension to \(C_a\). Finally, the abundance matrix is generated using a softmax function.

\noindent\textbf{Experimental settings} In our experiments, we extracted 64$\times$64 patches and fed them into the network, using a patch size of 2 in the spatial ViT. We set the number of tokens $N_{spec}$ to 64. The decoder used a convolutional kernel size of 1, following \cite{deeptrans}. Based on preliminary experiments, we set $\alpha$ to 0.35 and $\beta$ to 0.1. The model was trained for 200 epochs with a batch size of 32.

\begin{table*}[h]
\centering
\caption{
Quantitative comparison of different HSI denoising methods on the WDC Mall dataset. \textbf{\color{red}{Best}} and \textbf{\color{blue}{2nd-best}} results are highlighted.
}
\label{tab:denoising_appendix}
\resizebox{\linewidth}{!}{
\begin{tabular}{cccccccccccc}
\hline
Cases & Index & Noisy & LLRT \cite{DBLP:conf/cvpr/ChangYZ17} & NGMeet\cite{DBLP:conf/cvpr/0003YLYZ19} & LRTFL0\cite{xiong2019hyperspectral} & E-3DTV \cite{peng2020enhanced} & DS2DP \cite{miao2022hyperspectral} & QRNN3D \cite{DBLP:journals/tnn/WeiFH21} & SST\cite{DBLP:conf/aaai/LiFZ23}  & SpatSIGMA & HyperSIGMA \\ \hline
\multirow{3}{*}{Case 1} & PSNR & 11.226 & 28.888 & 29.591 & 27.375 & 27.166 & 28.428 & 29.232 & 31.133 & \textbf{\color{blue} 31.211} & \textbf{\color{red} 31.236} \\
                        & SSIM & 0.367  & 0.956  & 0.964  & 0.935  & 0.934  & 0.949  & 0.954  & 0.973  & \textbf{\color{blue} 0.974}  & \textbf{\color{red} 0.975} \\
                        & SAM  & 0.670   & 0.106  & 0.105  & 0.151  & 0.116  & 0.120   & 0.104  & 0.086  & \textbf{\color{blue} 0.083}  & \textbf{\color{red} 0.082} \\ \hline
\multirow{3}{*}{Case 2} & PSNR & 11.315 & 22.296 & 24.918 & 25.983 & 26.884 & 27.607 & 28.542 & \textbf{\color{blue} 28.756} & 28.669 & \textbf{\color{red} 28.961} \\
                        & SSIM & 0.362  & 0.814  & 0.878  & 0.905  & 0.928  & 0.941  & 0.949  & 0.960  & \textbf{\color{blue} 0.962}  & \textbf{\color{red} 0.963} \\
                        & SAM  & 0.707  & 0.231  & 0.254  & 0.198  & 0.117  & 0.131  & 0.132  & 0.111  & \textbf{\color{blue} 0.107}  & \textbf{\color{red} 0.106} \\ \hline
\multirow{3}{*}{Case 3} & PSNR & 11.105 & 22.174 & 23.788 & 25.581 & 26.709 & 27.097 & 28.152 & 28.216 & \textbf{\color{blue} 28.394} & \textbf{\color{red} 28.735} \\
                        & SSIM & 0.353  & 0.828  & 0.875  & 0.909  & 0.933  & 0.939  & 0.945  & 0.956  & \textbf{\color{blue} 0.959}  & \textbf{\color{red} 0.961} \\
                        & SAM  & 0.717  & 0.221  & 0.259  & 0.199  & 0.118  & 0.142  & 0.134  & \textbf{\color{blue} 0.115}  & \textbf{\color{red} 0.106}  & \textbf{\color{red} 0.106} \\ \hline
\multirow{3}{*}{Case 4} & PSNR & 10.916 & 22.076 & 24.363 & 25.560 & 26.587 & 27.310 & 28.037 & \textbf{\color{blue} 28.738} & 28.506 & \textbf{\color{red} 28.781} \\
                        & SSIM & 0.342  & 0.857  & 0.860  & 0.897  & 0.926  & 0.937  & 0.944  & 0.958  & \textbf{\color{blue} 0.959}  & \textbf{\color{red} 0.961} \\ 
                        & SAM  & 0.714  & 0.228  & 0.267  & 0.205  & 0.119  & 0.135  & 0.134  & \textbf{\color{blue} 0.109}  & \textbf{\color{red} 0.106}  & \textbf{\color{red} 0.106} \\ 
\hline
\end{tabular}
}
\end{table*}

\begin{figure*}[t]
  \centering
  \includegraphics[width=\linewidth]{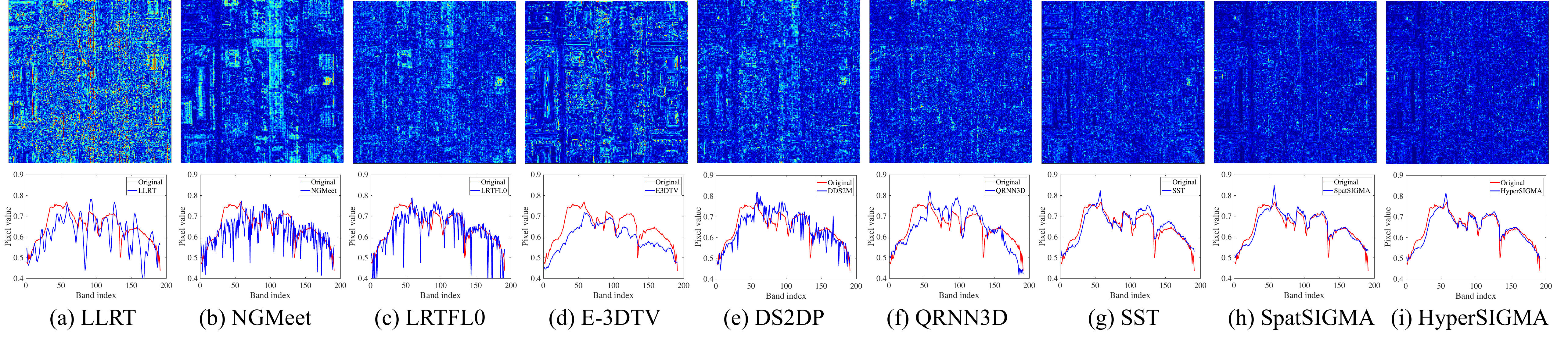}
  \vspace{-6mm}
  \caption{
  HSI denoising results using different methods on the WDC Mall dataset for the case of in the main text. Top row: average error maps across all bands. Bottom row: spectral curves at spatial location (98, 166) of the reconstructed results.
  }
\label{fig:denoising}
\end{figure*}

\begin{figure}[h]
\centering
\includegraphics[width=\linewidth]{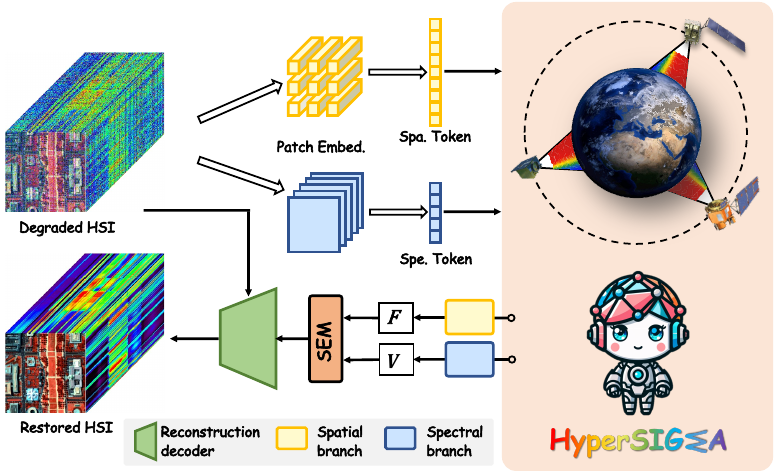}
\caption{Diagram illustrating the application of HyperSIGMA for HSI denoising.}
\label{fig: Image_denoising}
\end{figure}

\begin{figure*}[!t]
  \centering
  \includegraphics[width=\linewidth]{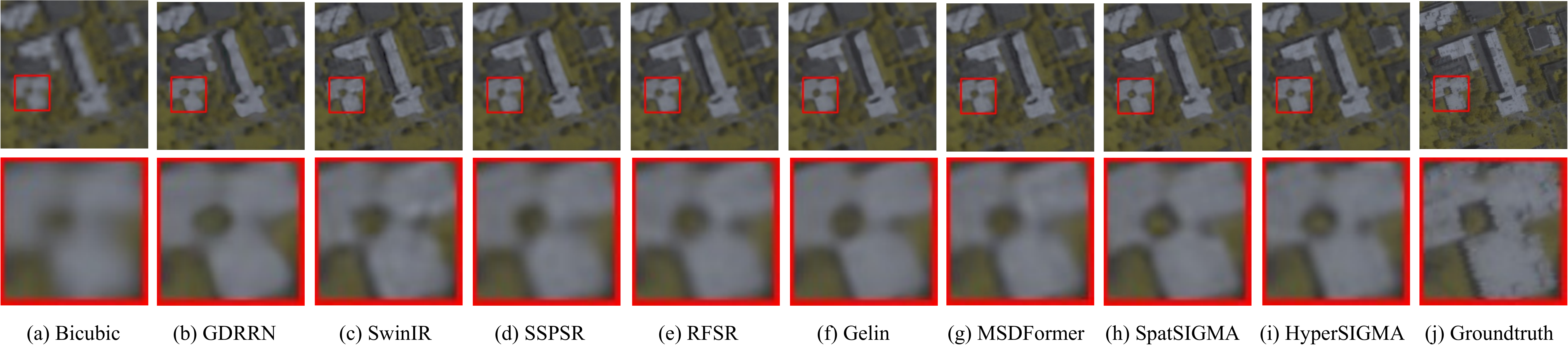}
  \vspace{-6mm}
  \caption{
  HSI super-resolution results on the Houston dataset for spectral bands 29-26-19 (R-G-B) with an 8$\times$ scale factor.
  }
\label{fig:super}
\end{figure*}

\subsection{More Results}
\noindent\textbf{Predictive Performance for Each Endmember} Table \ref{hsi_unmixing_appendix} provides the detailed model performance for each endmember, involving abundance and endmember predictions. It can be seen that our models have advantages on almost all endmembers, especially for obvious foreground endmembers, such as Tree and Roof.

\noindent\textbf{Qualitative Results} Fig. \ref{fig: unmixing_Urban_abundance} shows that HyperSIGMA accurately characterizes the actual distributions of mixed ground objects in HSIs. Additionally, Fig. \ref{fig: unmixing_Urban_endmember} illustrates that HyperSIGMA effectively captures pure endmember signatures, closely aligning with the ground truth, especially for the tree endmember.

\section{Hyperspectral Image Denoising \label{sec_hyper_denoising}}

\subsection{Implementation Details}

\noindent\textbf{Model Architecture} The HyperSIGMA model's architecture is shown in Fig. \ref{fig: Image_denoising}. First, the HSI with added noise is input into the model. In HyperSIGMA, we use only the final layer's spatial feature and spectral feature from their respective subnetworks. These features are then fed into additional Spatial and Spectral branches, obtaining $\bm{F}$ and $\bm{V}$. Specifically, to handle the high-dimensional output features, we stack four spatial/spectral transformer layers with an embedding dimension of 191 and a patch size of 1 to reduce the dimensions. These spatial/spectral layers are trained from scratch, producing outputs that better fit the fusion stage and reconstruction decoder. The two features are then combined via an SEM module. Before reconstruction, the input image is concatenated with the spatial-spectral fusion feature. The final image restoration is achieved through a decoder with two convolutional layers.

\noindent\textbf{Experimental Details} We select a central area of size 192$\times$192 for testing, while the remaining regions are used for training. Each iteration involves selecting 64$\times$64 patches, and the patch size of the spatial subnetwork is set to 2.

\noindent\textbf{Different Noise Cases} To fully assess the model, we also test the model under other noise cases: 

$ Case\ 1 $ \textit{(i.i.d Gaussian Noise)}: In this case, i.i.d. zero-mean Gaussian noise with a variance of 70 is added to all bands. 

$ Case\ 2 $ \textit{(non-i.i.d Gaussian Noise + Impulse Noise)}: This case includes unknown non-i.i.d. Gaussian noise with variances between 10 and 70, affecting different bands at different levels. Additionally, impulsive noise is introduced, following an i.i.d. zero-mean Laplacian distribution with a density parameter ranging from 0.1 to 0.7. 

$ Case\ 3 $ \textit{(non-i.i.d Gaussian Noise + Impulse Noise + Stripes)}: To increase complexity, stripes are added to $ Case\ 2 $. Stripes randomly affect 30\% of the bands, with each affected band having 10 to 15 stripes. The stripe elements are assigned values randomly generated between 0.6 and 0.8. 

$ Case\ 4 $ \textit{(non-i.i.d Gaussian Noise + Impulse Noise + Deadline)}: Here, the stripes in $ Case\ 3 $ are replaced by deadlines. Deadlines randomly affect 30\% of the bands, with each selected band containing 10 to 15 deadlines. The spatial width of the deadlines ranges from 1 to 3 pixels. 

The case presented in the main text is actually the combination of $ Case\ 3 $ and $ Case\ 4 $.

\subsection{More Results}

\noindent\textbf{Visualization} For the case in the main text, Fig. \ref{fig:denoising} visualizes the comparisons of spatial and spectral reconstruction performance. The results further highlight the advantages of our models, showing minimal spatial reconstruction errors and the least deviation in spectral curves compared to Ground Truth. In the error maps, a higher concentration of blue pixels indicates smaller errors relative to Ground Truth, while greater similarity in spectral curves signifies better reconstruction. Both SpatSIGMA and HyperSIGMA clearly outperform other methods, with HyperSIGMA performing better due to the utilization of spectral information, aligning with the quantitative results.

\noindent\textbf{Extra Cases} We present more quantitative results in Table \ref{tab:denoising_appendix}. It can be seen that our models consistently outperform other methods, especially in the more challenging Cases 3 and 4. We attribute this to the superior feature representations by pre-training models, which can be effectively adapted to denoising through a specially designed decoder optimized for low-dimensional tasks.

\section{Hyperspectral Image Super-Resolution \label{sec_hyper_super_resolution}}

\begin{table}[!t]
\centering
\caption{Quantitative comparison of different methods on the Houston dataset at 4$\times$ scale factors. \textbf{\color{red}{Best}} and \textbf{\color{blue}{2nd-best}} results are highlighted.}
\label{tab:super_appendix}
\resizebox{\linewidth}{!}{
    \begin{tabular}{lcccccc}
    \hline
        Method & PSNR & SSIM & SAM & CC & RMSE & ERGAS \\ 
        \hline
        Bicubic  & 43.027 & 0.961 & 2.545 & 0.974 & 0.009 & 2.909 \\ 
        GDRRN \cite{li2018single}  & 44.296 & 0.973 & 2.535 & 0.976 & 0.007 & 2.470 \\ 
        SwinIR \cite{liang2021swinir}  & 46.097 & \textbf{\color{blue}{0.981}} & 1.946 & 0.986 & \textbf{\color{blue}{0.006}} & 2.004 \\ 
        SSPSR \cite{jiang2020learning}  & 45.602 & 0.978 & 1.965 & 0.985 & \textbf{\color{blue}{0.006}} & 2.138 \\ 
        RFSR \cite{wang2021hyperspectral}   & 45.868 & 0.979 & 1.830 & 0.986 & \textbf{\color{blue}{0.006}} & 2.066 \\ 
        Gelin \cite{wang2022group}  & 45.872 & 0.979 & 1.876 & 0.986 & \textbf{\color{blue}{0.006}} & 2.078 \\
        MSDFormer \cite{msdformer}   & 46.201 & \textbf{\color{blue}{0.981}} & 1.777 & 0.987 &  \textbf{\color{blue}{0.006}} & 1.996 \\
        \hline
        SpatSIGMA  & \textbf{\color{blue}{46.337}} & \textbf{\color{red}{0.982}} & \textbf{\color{blue}{1.789}} & \textbf{\color{blue}{0.980}} & \textbf{\color{red}{0.005}} & \textbf{\color{blue}{1.954}} \\  
        HyperSIGMA  & \textbf{\color{red}{46.455}} & \textbf{\color{red}{0.982}} & \textbf{\color{red}{1.782}} & \textbf{\color{red}{0.988}} & \textbf{\color{red}{0.005}} & \textbf{\color{red}{1.923}} \\ 
        \hline 
    \end{tabular}
    }
\end{table}

\subsection{Implementation Details} 

\noindent\textbf{Network Structure} The architecture for performing HSI super-resolution is similar to HSI denoising, as shown in Fig. \ref{fig: Image_denoising}. The main differences are: (1) the patch size of the spatial subnetwork is set to 1, and (2) after the reconstruction decoder, we use PixelShuffle \cite{DBLP:conf/cvpr/ShiCHTABRW16} to upsample the reconstruction result (e.g., a scale factor of 8$\times$ corresponds to upsampling 3 times). This upsampling operation ensures the network outputs the expected size.

\noindent\textbf{Experimental Details} All patches are uniformly downsampled to 32 $\times$ 32. Therefore, if for 8$\times$ super-resolution, we extract 256 $\times$ 256 patches.

\subsection{More Results}

\begin{figure*}[!t]
  \centering
  \includegraphics[width=\linewidth]{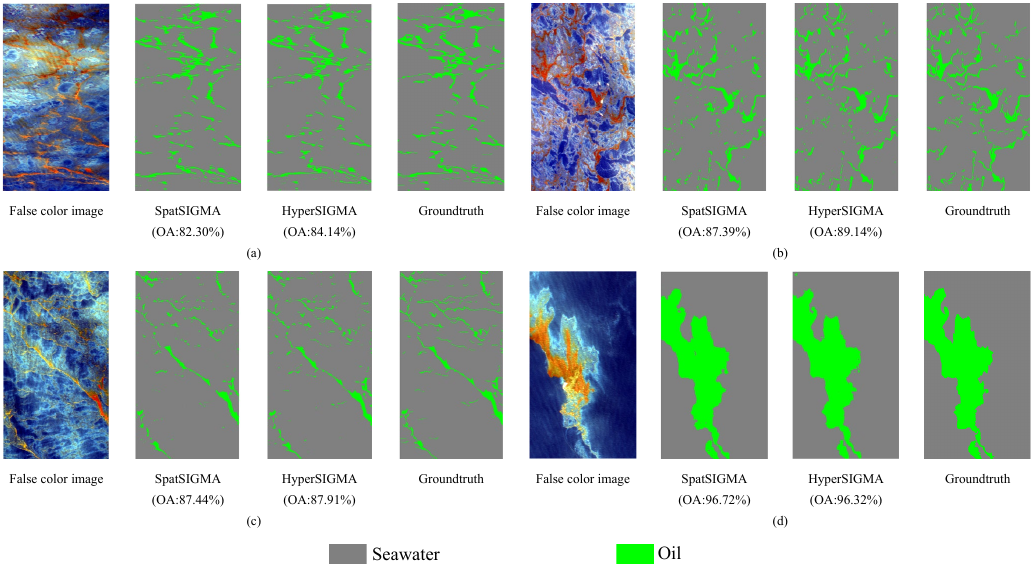}
  \caption{Visualization of detected oil leakage in various Gulf of Mexico regions by our models. (a) GM07. (b) GM08. (c) GM09. (d) GM17.}
\label{fig:oil_other_examples}
\end{figure*}

\noindent\textbf{Qualitative Results} Fig. \ref{fig:super} illustrates the effectiveness of our methods at an 8$\times$ scale factor, showing clearer visual results compared to other competitors, particularly in the highlighted zoomed-in areas. 

\noindent\textbf{Other Upsampling Ratios} We additionally conduct an experiment with a 4$\times$ super-resolution, as shown in Table \ref{tab:super_appendix}. It can be seen that our models demonstrate clear performance advantages.

\section{Model Scalability Experiments}
\label{sec_model_scalability}

\subsection{Implementation Details}

\noindent\textbf{Using Different ViT Backbones} The experimental settings on the Indian Pines dataset are consistent with those used in Sec.~\ref{subsec_cls_setting}. For the Xiongan dataset, models were trained for 300 epochs with a batch size of 512. The training set included 100 randomly selected samples per class, with the remainder used for evaluation.

\subsection{More results}

\begin{table}[t]
  \caption{Classification accuracies of models using different pre-training weights on the HanChuan dataset.}
  \resizebox{\linewidth}{!}{
    \begin{tabular}{lccccc}
    \hline
    Model & SpatViT & SpecViT & OA & AA & Kappa \\
    \hline
    SpatSIGMA-Spec  &  Spec   & - & 92.93  & 86.51 & 91.89 \\
    SpatSIGMA  &  Spat   & - & 94.03 & 88.96 & 93.16 \\
    \hline
    HyperSIGMA-Interchange  &  Spec   & Spat & 90.64 & 83.35 & 89.30 \\
    HyperSIGMA-DualSpec  &  Spec   & Spec & 91.51 & 84.21 & 90.27 \\ 
    HyperSIGMA-DualSpat  &  Spat   & Spat & 93.53 & 88.78 & 92.58 \\
    HyperSIGMA &  Spat   & Spec   &94.44  & 90.41 & 93.62 \\
    \hline
    \end{tabular}
    }
    \label{table:weight_flex}
\end{table}

\noindent\textbf{Loading Different Weights} In HyperSIGMA, both the spatial and spectral subnetworks utilize a ViT structure, allowing pre-trained weights from one branch to be reused in the other. We explored this by fine-tuning on the HanChuan dataset, and initializing SpatSIGMA and HyperSIGMA with different pre-training weights, as detailed in Table \ref{table:weight_flex}. Using spectral ViT pre-training weights in SpatSIGMA led to a performance drop, highlighting the necessity of spatial pre-training (i.e., matched) weights for the spatial subnetwork. HyperSIGMA also showed reduced accuracy with mismatched weights, with the worst performance occurring when both subnetworks used mismatched weights. Comparing SpatSIGMA-Spec and HyperSIGMA-DualSpec, we found that matched weights in the spectral branch did not enhance accuracy if the spatial branch weights were mismatched. Conversely, comparing SpatSIGMA and HyperSIGMA-DualSpat revealed that mismatched spectral branch weights caused a milder performance decline. These results indicate that the spatial branch has stricter weight initialization requirements. Ultimately, SpatSIGMA and HyperSIGMA performed best with matched weights, underscoring the importance of structure-weight matching and confirming the auxiliary role of the spectral subnetworks.

\section{Cross-modal Transferability Experiments}
\label{sec_cross_modal}

\noindent\textbf{Network Structure} For collaborative classification, we adopt the same segmentation structure for HyperSIGMA following previous classification fine-tuning experiments, as shown in Fig. \ref{fig: HSIClassificationFlowchart}, where we input 128 $\times$ 128 patches, with the spatial patch size is set to 2. For the multispectral CD task, the structure of HyperSIGMA is similar to that shown in Fig. \ref{fig: CD_FlowChart}, but incorporates a simple feature pyramid network (FPN) like \cite{Chen2023Exchange} to upsample features layer by layer, gradually recovering spatial details for the final prediction. 

\noindent\textbf{Experimental Details} Data processing and label split of collaborative classification and CD tasks follow the protocol in \cite{spdda} and \cite{spectralgpt}, respectively.

\section{More Results of Oil Leakage Detection \label{sec_real_world}}

Fig. \ref{fig:oil_other_examples} shows the visualization results of oil leakage detection using HyperSIGMA on other regions of the Gulf of Mexico.

\section{More Complexity Analyses and Experimental Details}
\label{sec_model_complexity}

\noindent\textbf{Detailed Complexity Calculation Process} Given a 2-D feature map $\mathbf{X}\in\mathbb{R}^{N\times D'}$ that has $N$ tokens with dimensions in $D'$, we first generate query, key and value maps $\mathbf{Q}=\{q_1, \cdots, q_N\}$, $\mathbf{K}=\{k_1, \cdots, k_N\}$, and $\mathbf{V}=\{v_1, \cdots, v_N\}$, requiring a complexity of $O(3ND'^2)$. Then, we predict $N_p$ positions (including X and Y coordinates) for each query, spending $O(2NN_pD')$ computational complexity. The new $D'$ dimension key and value vectors in $\mathbf{K}', \mathbf{V}' \in \mathbb{R}^{N\times N_p \times D'}$ are both sampled by the bilinear interpolation considering four nearest points, and this process totally costs $O(8NN_pD')$ complexity. The multiplication $\otimes$ between $\mathbf{Q}$ and $\mathbf{K}'$, and $\mathbf{Q}\otimes\mathbf{K}'$ and $\mathbf{V}'$, costing totally $O(2NN_pD')$ computational complexity, while the softmax process in $\mathbf{Q}\otimes\mathbf{K}'$ need a complexity of $O(NN_p)$. Therefore, the computational complexity of SSA is $O(3ND'^2+12NN_pD'+NN_p)$.

\noindent\textbf{Experimental Settings} We use the Indian Pines dataset, which contains $N = 21,025$ pixels, and perform category prediction for each pixel via patch-level classification. For model input, we select a surrounding patch of size $p \times p$, resulting in an input size of $N \times p \times p$. We conduct experiments with different patch sizes $p$, maintaining a constant batch size of 8. To assess inference time, we measure the total time required to predict all pixels and then calculate the average time per pixel by dividing the total time by $N$.

\section{Datasheet for HyperGlobal-450K
\label{sec_datasheet}}
\subsection{Motivation} 
\noindent 1. \textit{For what purpose was the dataset created?} \vspace{0.5\baselineskip}\\
\noindent\textbf{A1:} The HyperGlobal-450K dataset is designed to facilitate the development of large foundational models for HSI processing and analysis using self-supervised learning. Given the scarcity of large-scale, high-quality hyperspectral datasets with global coverage, this dataset integrates data from multiple sensors to build a model with powerful generalization ability. HyperGlobal-450K provides a comprehensive, diverse, and temporally extensive HSI resource, driving innovation in RS technologies and models, and supporting applications such as land cover classification and environmental monitoring.\vspace{0.5\baselineskip}

\noindent 2. \textit{Who created the dataset (e.g., which team, research group) and on behalf of which entity (e.g., company, institution, organization)? } \vspace{0.5\baselineskip}\\
\noindent\textbf{A2:} The dataset was created by:

Di Wang (Wuhan University),

Meiqi Hu (Wuhan University),

Yao Jin (Wuhan University),

Yuchun Miao (Wuhan University),

Jiaqi Yang (Wuhan University),

Yichu Xu (Wuhan University),

Xiaolei Qin (Wuhan University),

Jiaqi Ma (Wuhan University),

Lingyu Sun (Wuhan University),

Chenxing Li (Wuhan University),

Chuan Fu (Chongqing University),

Hongruixuan Chen (University of Tokyo),

Chengxi Han (Wuhan University),

Naoto Yokoya (University of Tokyo),

Jing Zhang (Wuhan University),

Minqiang Xu (National Engineering Research Center of Speech and Language Information Processing),

Lin Liu (National Engineering Research Center for Speech and Language Information Processing),

Lefei Zhang (Wuhan University),

Chen Wu (Wuhan University),

Bo Du (Wuhan University),

Dacheng Tao (Nanyang Technological University),

Liangpei Zhang (Wuhan University).\vspace{0.5\baselineskip}

\noindent 3. \textit{Who funded the creation of the dataset? }\vspace{0.5\baselineskip}\\
\noindent\textbf{A3:} The dataset creation was funded by the affiliations of the authors involved in this work.
\subsection{Composition}
\noindent 1. \textit{What do the instances that comprise the dataset represent (e.g., documents, photos, people, countries)? Are there multiple types of instances(e.g., movies, users, and ratings; people and interactions between them; nodes and edges)? Please provide a description. }\vspace{0.5\baselineskip}\\
\noindent\textbf{A1:} This dataset is composed of the image subsets captured by the EO-1 satellite and the Gaofen-5 satellite. In these subsets, each sample is a single HSI. 

\noindent 2. \textit{How many instances are there in total (of each type, if appropriate)?}\vspace{0.5\baselineskip}\\
\noindent\textbf{A2:} HyperGlobal-450K contains 447,072 HSIs, including 247,072 EO-1 image patches and 200,000 GF-5 samples.\vspace{0.5\baselineskip}

\noindent 3. \textit{Does the dataset contain all possible instances or is it a sample (not necessarily random) of instances from a larger set? If the dataset is a sample, then what is the larger set? Is the sample representative of the larger set (e.g., geographic coverage)? If so, please describe how this representativeness was validated/verified. If it is not representative of the larger set, please describe why not (e.g., to cover a more diverse range of instances, because instances were withheld or unavailable).}\vspace{0.5\baselineskip}\\
\noindent\textbf{A3:} 
HyperGlobal-450K is sampled from larger sets, which contains so far all images photoed by EO-1 and GF-5 satellites. For the EO-1 instance, since the locations are selected to cover the globe path/row pairs in the Worldwide Reference System as far as possible, it can represent the large EO-1 set to some extent. The GF-5 instances in HyperGlobal-450K can represent most of the large GF-5 set, since they encompass a variety of landscapes in China, such as Grasslands, forests, farmlands, etc. \vspace{0.5\baselineskip}

\noindent 4. \textit{What data does each instance consist of? “Raw” data (e.g., unprocessed text or images)or features? In either case, please provide a description.}\vspace{0.5\baselineskip}\\
\noindent\textbf{A4:} Each instance consists of a clipped HSI patch. \vspace{0.5\baselineskip}

\noindent 5. \textit{Is there a label or target associated with each instance? If so, please provide a description.}\vspace{0.5\baselineskip}\\
\noindent\textbf{A5:} No, the dataset is intended for self-supervised learning. Therefore, each instance is an individual RS image and does not possess any form of annotations. \vspace{0.5\baselineskip}

\noindent 6. \textit{Is any information missing from individual instances? If so, please provide a description, explaining why this information is missing (e.g., because it was unavailable). This does not include intentionally removed information, but might include, e.g., redacted text.}\vspace{0.5\baselineskip}\\
\noindent\textbf{A6:} No. \vspace{0.5\baselineskip}

\noindent 7. \textit{Are relationships between individual instances made explicit (e.g., users’ movie ratings, social network links)? If so, please describe how these relationships are drawn.}\vspace{0.5\baselineskip}\\
\noindent\textbf{A7:} Yes, the relationships between different instances are shown in filenames. \vspace{0.5\baselineskip}

\noindent 8. \textit{Are there recommended data splits (e.g., training, development/validation, testing)? If so, please provide a description of these splits, explaining the rationale behind them.}\vspace{0.5\baselineskip}\\
\noindent\textbf{A8:} Yes. The entire dataset is designated for self-supervised learning algorithms. Therefore, we recommend the whole dataset be utilized for self-supervised learning research. \vspace{0.5\baselineskip}

\noindent 9. \textit{Are there any errors, sources of noise, or redundancies in the dataset? If so, please provide a description.}\vspace{0.5\baselineskip}\\
\noindent\textbf{A9:} No. \vspace{0.5\baselineskip}

\noindent 10. \textit{Is the dataset self-contained, or does it link to or otherwise rely on external resources (e.g., websites, tweets, other datasets)? If it links to or relies on external resources, a) are there guarantees that they will exist, and remain constant, over time; b) are there official archival versions of the complete dataset (i.e., including the external resources as they existed at the time the dataset was created); c) are there any restrictions (e.g., licenses, fees) associated with any of the external resources that might apply to a future user? Please provide descriptions of all external resources and any restrictions associated with them, as well as links or other access points, as appropriate.}\vspace{0.5\baselineskip}\\
\noindent\textbf{A10:} The dataset is self-contained as it is derived from raw satellite imagery downloaded from official sources, but users should refer to the original platforms for licensing terms and note that the availability of the raw data may vary over time. \vspace{0.5\baselineskip}

\noindent 11. \textit{Does the dataset contain data that might be considered confidential (e.g., data that is protected by legal privilege or by doctor-patient confidentiality, data that includes the content of individuals non-public communications)? If so, please provide a description.}\vspace{0.5\baselineskip}\\
\noindent\textbf{A11:} No. \vspace{0.5\baselineskip}

\noindent 12. \textit{Does the dataset contain data that, if viewed directly, might be offensive, insulting, threatening, or might otherwise cause anxiety? If so, please describe why.}\vspace{0.5\baselineskip}\\
\noindent\textbf{A12:} No. \vspace{0.5\baselineskip}

\subsection{Collection process} 
\noindent 1. \textit{How was the data associated with each instance acquired? Was the data directly observable (e.g., raw text, movie ratings), reported by subjects (e.g., survey responses), or indirectly inferred/derived from other data (e.g., part-of-speech tags, model-based guesses for age or language)? If data was reported by subjects or indirectly inferred/derived from other data, was the data validated/verified? If so, please describe how.} \vspace{0.5\baselineskip}\\
\noindent\textbf{A1:} The data associated with each instance is directly observable, as they are HSIs and can be acquired from EO-1 or GF-5 hyperspectral satellite imagery.
\vspace{0.5\baselineskip}

\noindent 2. \textit{What mechanisms or procedures were used to collect the data (e.g., hardware apparatus or sensor, manual human curation, software program, software API)? How were these mechanisms or procedures validated?} \vspace{0.5\baselineskip}\\
\noindent\textbf{A2:} The dataset consists of HSIs that were downloaded from \href{https://www.usgs.gov/centers/eros/science/usgs-eros-archive-earth-observing-one-eo-1-hyperion}{USGS} and \href{https://grid.cpeos.org.cn/}{CPEOS}. The images were filtered based on cloud cover, location, and spectral bands. The cloud cover selection is performed online. The Python scripts are compiled to perform location and band selection. The satellite sensors were pre-calibrated, and all the scripts were thoroughly tested to ensure reliability.
\vspace{0.5\baselineskip}

\noindent 3. \textit{If the dataset is a sample from a larger set, what was the sampling strategy (e.g., deterministic, probabilistic with specific sampling probabilities)?} \vspace{0.5\baselineskip}\\
\noindent\textbf{A3:} The EO-1 instances were sampled from the aforementioned large set using three strategies: (1) Deterministic cloud cover selection: screening the data with cloud cover percentage less than 5\% from 2011-2017 and excluding the data with unknown cloud cover percentage. (2) Random location selection: randomly selecting one image each year from each path/row location derived from the Worldwide Reference System (WRS). (3) Deterministic band selection: removing bad bands and water vapor absorption bands. For GF-5, the instances were sampled from related large sets by deterministic band and location selections.\vspace{0.5\baselineskip}

\noindent 4. \textit{Who was involved in the data collection process (e.g., students, crowdworkers, contractors) and how were they compensated (e.g., how much were crowdworkers paid)} \vspace{0.5\baselineskip}\\
\noindent\textbf{A4:} The data was collected and verified by the authors.\vspace{0.5\baselineskip}

\noindent 5. \textit{Over what timeframe was the data collected? Does this timeframe match the creation timeframe of the data associated with the instances (e.g., recent crawl of old news articles)? If not, please describe the timeframe in which the data associated with the instances was created.} \vspace{0.5\baselineskip}\\
\noindent\textbf{A5:} The data download takes approximately one month. Then, performing location selection costs about 1 day, while it takes about 7 days for band selection and clipping.
\vspace{0.5\baselineskip}

\subsection{Preprocessing/cleaning/labeling} 
\noindent 1. \textit{Was any preprocessing/cleaning/labeling of the data done (e.g., discretization or bucketing, tokenization, part-of-speech tagging, SIFT feature extraction, removal of instances, processing 5 of missing values)? If so, please provide a description. If not, you may skip the remainder of the questions in this section.}\vspace{0.5\baselineskip}\\
\noindent\textbf{A1:} The EO-1 and GF-5B hyperspectral images are clipped into hyperspectral patches with a spatial size of 64×64, to facilitate network reading and training.\vspace{0.5\baselineskip}

\noindent 2. \textit{Was the “raw” data saved in addition to the preprocessed/cleaned/labeled data (e.g., to support unanticipated future uses)? If so, please provide a link or other access point to the “raw” data.}\vspace{0.5\baselineskip}\\
\noindent\textbf{A2:} No.\vspace{0.5\baselineskip}

\noindent 3. \textit{Is the software used to preprocess/clean/label the instances available? If so, please provide a link or other access point.} \vspace{0.5\baselineskip}\\
\noindent\textbf{A3:} We use Python for clipping.\vspace{0.5\baselineskip}

\subsection{Uses} 
\noindent 1. \textit{Has the dataset been used for any tasks already? If so, please provide a description.}\vspace{0.5\baselineskip}\\
\noindent\textbf{A1:} No.
\vspace{0.5\baselineskip}

\noindent 2. \textit{Is there a repository that links to any or all papers or systems that use the dataset? If so, please provide a link or other access point.
}\vspace{0.5\baselineskip}\\
\noindent\textbf{A2:} N/A.\vspace{0.5\baselineskip}

\noindent 3. \textit{What (other) tasks could the dataset be used for?
} \vspace{0.5\baselineskip}\\
\noindent\textbf{A3:} HyperGlobal-450K can be used for unsupervised pre-training. Besides, it can be used for evaluation on specific tasks, such as image denoising and super-resolution. With further manual labeling, it can also be used for semantic segmentation, anomaly detection, target detection, etc.
\vspace{0.5\baselineskip}

\noindent 4. \textit{Is there anything about the composition of the dataset or the way it was collected and preprocessed/cleaned/labeled that might impact future uses? For example, is there anything that a future user might need to know to avoid uses that could result in unfair treatment of individuals or groups (e.g., stereotyping, quality of service issues) or other undesirable harms (e.g., financial harms, legal risks) If so, please provide a description. Is there anything a future user could do to mitigate these undesirable harms?
} \vspace{0.5\baselineskip}\\
\noindent\textbf{A4:} No.
\vspace{0.5\baselineskip}

\noindent 5. \textit{Are there tasks for which the dataset should not be used? If so, please provide a description.
} \vspace{0.5\baselineskip}\\
\noindent\textbf{A5:} No.
\vspace{0.5\baselineskip}

\subsection{Distribution} 
\noindent 1. \textit{Will the dataset be distributed to third parties outside of the entity (e.g., company, institution, organization) on behalf of which the dataset was created?
}\vspace{0.5\baselineskip}\\
\noindent\textbf{A1:} Yes. The dataset will be publicly available.
\vspace{0.5\baselineskip}

\noindent 2. \textit{How will the dataset be distributed (e.g., tarball on website, API, GitHub)?}\vspace{0.5\baselineskip}\\
\noindent\textbf{A2:} It will be publicly available on the project website at \href{https://huggingface.co/datasets/WHU-Sigma/HyperGlobal-450K}{HyperGlobal-450K}.
\vspace{0.5\baselineskip}

\noindent 3. \textit{When will the dataset be distributed?
}\vspace{0.5\baselineskip}\\
\noindent\textbf{A3:} The dataset will be distributed once the paper is accepted after peer review.
\vspace{0.5\baselineskip}

\noindent 4. \textit{Will the dataset be distributed under a copyright or other intellectual property (IP) license, and/or under applicable terms of use (ToU)? If so, please describe this license and/or ToU, and provide a link or other access point to, or otherwise reproduce, any relevant licensing terms or ToU, as well as any fees associated with these restrictions.
}\vspace{0.5\baselineskip}\\
\noindent\textbf{A4:} It will be distributed under the \href{https://creativecommons.org/licenses/by-nc-sa/4.0/}{Creative Commons Attribution-NonCommercial-ShareAlike 4.0 License}.
\vspace{0.5\baselineskip}

\noindent 5. \textit{Have any third parties imposed IP-based or other restrictions on the data associated with the instances? If so, please describe these restrictions, and provide a link or other access point
to, or otherwise reproduce, any relevant licensing terms, as well as any fees associated with
these restrictions. 
}\vspace{0.5\baselineskip}\\
\noindent\textbf{A5:} No.
\vspace{0.5\baselineskip}

\noindent 6. \textit{Do any export controls or other regulatory restrictions apply to the dataset or to individual instances? If so, please describe these restrictions, and provide a link or other access point to, or otherwise reproduce, any supporting documentation.
}\vspace{0.5\baselineskip}\\
\noindent\textbf{A6:} No.
\vspace{0.5\baselineskip}

\subsection{Maintenance} 
\noindent 1. \textit{Who will be supporting/hosting/maintaining the dataset? }\vspace{0.5\baselineskip}\\
\noindent\textbf{A1:} The authors.
\vspace{0.5\baselineskip}

\noindent 2. \textit{How can the owner/curator/manager of the dataset be contacted (e.g., email address)? 
}\vspace{0.5\baselineskip}\\
\noindent\textbf{A2:} They can be contacted via email available on the project website. 
\vspace{0.5\baselineskip}

\noindent 3. \textit{Is there an erratum? If so, please provide a link or other access point. 
}\vspace{0.5\baselineskip}\\
\noindent\textbf{A3:} No. 
\vspace{0.5\baselineskip}

\noindent 4. \textit{Will the dataset be updated (e.g., to correct labeling errors, add new instances, delete instances)? If so, please describe how often, by whom, and how updates will be communicated to users (e.g., mailing list, GitHub)? 
}\vspace{0.5\baselineskip}\\
\noindent\textbf{A4:} No. 
\vspace{0.5\baselineskip}

\noindent 5. \textit{Will older versions of the dataset continue to be supported/hosted/maintained? If so, please describe how. If not, please describe how its obsolescence will be communicated to users. 
}\vspace{0.5\baselineskip}\\
\noindent\textbf{A5:} N/A. 
\vspace{0.5\baselineskip}

\noindent 6. \textit{If others want to extend/augment/build on/contribute to the dataset, is there a mechanism for them to do so? If so, please provide a description. Will these contributions be validated/verified? If so, please describe how. If not, why not? Is there a process for communicating/distributing these contributions to other users? If so, please provide a description. 
}\vspace{0.5\baselineskip}\\
\noindent\textbf{A6:} N/A. 
\vspace{0.5\baselineskip}

\section{Per-class Accuracies on Hyperspectral Image Classification Datasets\label{sec:detailed_hsic}}

We present the per-class classification accuracies of various methods on the Indian Pines, Pavia University, HanChuan, HongHu, Houston and ZY1-02D datasets in Tables \ref{Indian_Pines}-\ref{ZY1-02D}.

\begin{table*}[h]
    \centering
    \caption{Per-class classification accuracies of various methods on the Indian Pines dataset.}
    \begin{threeparttable}
    \resizebox{\linewidth}{!}{
    \begin{tabular}{cccccccccccc}
    \hline
        Class\tnote{1} & MSDN \cite{MSDN} & SSFCN \cite{ssfcn} & FullyContNet \cite{fullycontnet} & SpectralFormer \cite{spectralformer} & HSIC-FM \cite{HSIC-FM} & SSGRN \cite{ssgrn} & CSIL \cite{yang_csil} & IDCN \cite{idcn} & CLOLN \cite{CLOLN} & SpatSIGMA & HyperSIGMA \\ \hline
        1 & 55.08  & 88.89  & 82.41  & 84.26  & 92.59  & 100.00  & 99.07 & 86.02  & 81.46  & 73.80  & 44.28  \\ 
        2 & 52.60  & 19.56  & 50.73  & 36.84  & 34.39  & 53.87  & 42.38  & 57.50  & 67.76  & 87.62  & 91.99  \\ 
        3 & 38.72  & 29.96  & 57.76  & 39.35  & 27.28  & 32.07  & 62.03  & 53.49  & 49.78  & 81.31  & 88.96  \\ 
        4 & 37.13  & 54.04  & 75.92  & 51.54  & 45.23  & 86.78  & 87.22  & 81.67  & 52.68  & 73.19  & 72.18  \\ 
        5 & 61.44  & 43.62  & 78.65  & 63.57  & 61.66  & 77.80  & 80.20  & 84.25  & 95.84  & 96.69  & 100.00  \\ 
        6 & 75.77  & 51.02  & 73.52  & 64.31  & 41.76  & 63.19  & 88.75  & 86.94  & 98.35  & 94.00  & 88.25  \\ 
        7 & 21.11  & 83.33  & 92.59  & 100.00  & 55.56  & 100.00  & 100.00  & 97.43  & 45.07  & 65.00  & 100.00  \\ 
        8 & 86.78  & 87.75  & 92.52  & 82.19  & 66.17  & 100.00  & 99.72    & 88.91  & 97.78  & 98.28  & 92.88  \\ 
        9 & 11.77  & 96.67  & 100.00  & 96.67  & 100.00  & 100.00  & 100.00 & 100.00  & 46.72  & 15.62  & 8.06  \\ 
        10 & 38.53  & 51.42  & 78.52  & 50.49  & 16.53  & 69.75  & 64.00    & 71.36  & 84.64  & 74.00  & 89.56  \\ 
        11 & 61.94  & 33.44  & 64.16  & 42.52  & 21.06  & 69.12  & 55.01    & 60.35  & 75.00  & 87.04  & 87.23  \\ 
        12 & 39.16  & 30.30  & 57.06  & 29.10  & 34.08  & 65.18  & 51.69    & 56.92  & 61.22  & 71.40  & 68.77  \\ 
        13 & 68.41  & 82.39  & 97.95  & 85.81  & 91.11  & 100.00  & 99.15   & 95.96  & 72.52  & 89.90  & 54.70  \\
        14 & 84.22  & 51.71  & 91.95  & 63.05  & 47.57  & 83.58  & 78.83  &  98.36   & 83.72  & 96.70  & 95.67  \\ 
        15 & 71.70  & 46.45  & 91.31  & 39.98  & 40.51  & 99.20  & 87.32  &  81.48   & 76.42  & 69.49  & 76.77  \\ 
        16 & 38.17  & 84.34  & 87.95  & 89.96  & 78.31  & 100.00  & 99.60  &  100.00  & 58.84  & 78.78  & 67.54  \\ 
        \hline
        OA (\%) & 57.54  & 41.93  & 71.11  & 50.02  & 36.02  & 69.58  & 66.53  &  71.12 & 72.75  & 85.08  & 85.54  \\
        AA (\%) & 52.66  & 58.43  & 79.56  & 63.73  & 53.36  & 81.28  & 80.94  &  80.99 & 72.58  & 78.30  & 76.68  \\
        Kappa (\%) & 51.81  & 35.65  & 67.77  & 44.14  & 30.33  & 65.98  & 62.71 &  67.40 & 69.05  & 83.04  & 83.58 \\ \hline
    \end{tabular}
    }
    \begin{tablenotes}
  \scriptsize
  \item[1] The classes are as follows: 1. Alfalfa, 2. Corn-notill, 3. Corn-min, 4. Corn, 5. Grass-pasture, 6. Grass-tree, 7. Grass-pasture-mowed, 8. Hay-windrowed, 9. Oats,\\ 10. Soybeans-notill, 11. Soybeans-min, 12. Soybeans-clean, 13. Wheat, 14. Woods, 15. Bldg-grass-tree-drives, 16. Stone-stell-towers.
\end{tablenotes}
\end{threeparttable}
    \label{Indian_Pines}
\end{table*}

\begin{table*}[h]
    \centering
    \caption{Per-class classification accuracies of various methods on the Pavia University dataset.}
    \begin{threeparttable}
    \resizebox{\linewidth}{!}{
    \begin{tabular}{cccccccccccc}
    \hline
        Class\tnote{1} & MSDN \cite{MSDN} & SSFCN \cite{ssfcn} & FullyContNet \cite{fullycontnet} & SpectralFormer \cite{spectralformer} & HSIC-FM \cite{HSIC-FM} & SSGRN \cite{ssgrn} & CSIL \cite{yang_csil} & IDCN \cite{idcn} & CLOLN \cite{CLOLN} & SpatSIGMA & HyperSIGMA \\ \hline
        1 & 91.48  & 65.00  & 67.52  & 62.66  & 67.56  & 68.40  & 75.37 &  92.47 & 98.47  & 92.56 & 90.38 \\
        2 & 88.88  & 80.74  & 76.88  & 76.77  & 85.05  & 81.05  & 88.50 &  90.94 & 96.41  & 98.77 & 99.30  \\
        3 & 46.08  & 80.55  & 89.07  & 77.22  & 59.15  & 87.59  & 91.44 &  85.81 & 92.50  & 91.84 & 85.65 \\
        4 & 89.81  & 94.03  & 80.95  & 93.65 & 81.89  & 82.88  & 90.24  &  95.11  & 91.08  & 70.70  & 78.58 \\
        5 & 85.69  & 93.69  & 98.16  & 100.00  & 97.31  & 100.00  & 100.00 &  99.94 & 98.23 & 94.70  & 93.53  \\
        6 & 47.77  & 79.96  & 86.34  & 70.71  & 60.20  & 94.09  & 99.18 &  93.82  & 91.33  & 99.61 & 98.90  \\ 
        7 & 68.50  & 85.65  & 95.47  & 86.26  & 74.07  & 94.35  & 97.86 & 96.32  & 99.12  & 98.02 & 97.03 \\ 
        8 & 75.76  & 67.06  & 91.23  & 62.76  & 74.89  & 72.25  & 82.77 & 84.40  & 83.36 & 78.30  & 80.91 \\ 
        9 & 82.61  & 97.27 & 96.04  & 97.88  & 93.67  & 87.59  & 92.56 & 99.92  & 87.94  & 74.84  & 65.70  \\ 
        \hline
        OA (\%) & 76.48  & 78.88  & 80.31  & 75.37  & 77.28  & 81.45  & 88.23 & 91.64  & 93.11  & 93.36 & 93.52 \\ 
        AA (\%) & 75.18  & 82.28  & 86.85  & 80.88  & 77.09  & 85.35  & 90.88 &  93.19 & 91.94  & 88.82 & 87.78 \\ 
        Kappa (\%) & 69.52  & 72.99  & 75.01  & 68.56  & 70.36  & 76.36  & 84.76 & 89.05  & 90.88 & 90.66 & 90.93 \\ \hline
    \end{tabular}
    }
    \begin{tablenotes}
  \scriptsize
  \item[1] The classes are as follows: 1. Asphalt, 2. Meadows, 3. Gravel, 4. Trees, 5. Metal Sheets, 6. Bare Soil, 7. Bitumen, 8. Bricks, 9. Shadow.
\end{tablenotes}
\end{threeparttable}
    \label{Pavia_University}
\end{table*}

\begin{table*}[h]
    \centering
    \caption{Per-class classification accuracies of various methods on the HanChuan dataset.}
     \begin{threeparttable}
    \resizebox{\linewidth}{!}{
    \begin{tabular}{cccccccccccc}
    \hline
        Class\tnote{1} & MSDN \cite{MSDN} & SSFCN \cite{ssfcn} & FullyContNet \cite{fullycontnet} & SpectralFormer \cite{spectralformer} & HSIC-FM \cite{HSIC-FM} & SSGRN \cite{ssgrn} & CSIL \cite{yang_csil} & IDCN \cite{idcn} & CLOLN \cite{CLOLN} & SpatSIGMA & HyperSIGMA \\ \hline
        1 & 79.00  & 65.41  & 85.50  & 79.49  & 66.77  & 86.12  & 88.10 & 87.01 & 96.15  & 98.19 & 98.65 \\
        2 & 70.37  & 64.57  & 56.54  & 77.59  & 78.02  & 85.31  & 83.49 & 82.38 & 92.53  & 97.49  & 97.01 \\
        3 & 35.64  & 61.81  & 26.70  & 82.40  & 61.22  & 96.46  & 96.77 & 88.48 & 59.43  & 90.27 & 94.61 \\
        4 & 95.09  & 88.94  & 89.90  & 94.59  & 85.86  & 96.32  & 98.87 & 97.24 & 88.02  & 97.96 & 95.67 \\
        5 & 96.93  & 84.38  & 98.93  & 97.42  & 99.62  & 100.00  & 99.94& 97.36 & 74.39  & 80.79 & 74.71 \\
        6 & 25.19  & 39.45  & 80.81  & 50.50  & 47.52  & 98.68  & 92.27 & 75.96 & 52.29  & 63.62 & 91.70  \\
        7 & 90.69  & 58.22  & 94.21  & 86.77  & 52.86  & 97.77  & 86.93 & 75.75 & 89.69  & 77.89 & 86.55 \\
        8 & 35.39  & 50.78  & 84.43  & 64.84  & 50.93  & 89.93  & 87.34 & 66.21 & 91.43  & 97.42 & 96.37  \\
        9 & 47.10  & 39.03  & 59.88  & 69.49  & 52.03  & 77.81  & 84.80 & 75.01 & 70.69  & 85.72  & 87.34 \\
        10 & 60.12  & 52.82  & 63.25  & 79.63  & 36.21  & 91.98  & 92.99& 96.83  & 88.67  & 97.59 & 96.92  \\
        11 & 50.42  & 30.92  & 85.45  & 77.70  & 15.60  & 90.65  & 91.23& 71.41  & 94.82  & 94.93 & 89.86 \\
        12 & 64.04  & 70.41  & 95.32  & 80.58  & 68.76  & 99.94  & 99.99& 82.08  & 72.72  & 95.36 & 93.58 \\
        13 & 52.09  & 41.63  & 76.02  & 69.09  & 45.48  & 90.20  & 88.26& 80.65  & 58.12  & 75.72 & 75.56 \\
        14 & 84.82  & 69.20  & 79.48  & 73.45  & 72.95  & 81.19  & 83.26& 72.29  & 90.08  & 96.69 & 88.38 \\
        15 & 97.02  & 89.50  & 63.38  & 92.11  & 94.60  & 95.67  & 96.47& 96.01  & 60.62  & 74.02 & 80.07 \\
        16 & 94.95  & 76.87  & 87.49  & 97.69  & 84.46  & 95.28  & 88.45& 92.20  & 99.61  & 99.73 & 99.63  \\
        \hline
        OA (\%) & 73.40  & 63.35  & 78.80  & 82.60  & 66.21  & 90.43  & 88.55& 84.15  & 86.73  & 94.03  & 94.44 \\
        AA (\%) & 67.43  & 61.50  & 76.71  & 79.58  & 63.31  & 92.08  & 91.20& 83.49  & 79.96  & 88.96 & 90.41 \\
        Kappa (\%) & 69.09  & 57.79  & 75.56  & 79.74  & 60.55  & 88.88  & 86.75& 81.64  & 84.56  & 93.16  & 93.62 \\ \hline
    \end{tabular}
    }
        \begin{tablenotes}
  \scriptsize
  \item[1] The classes are as follows: 1. Strawberry, 2. Cowpea, 3. Soybean, 4. Sorghum, 5. Water spinach, 6. Watermelon, 7. Greens, 8. Trees, 9. Grass, 10. Red roof, \\11. Gray roof, 12. Plastic, 13. Bare soil, 14. Road, 15. Bright object, 16. Water.
\end{tablenotes}
\end{threeparttable}
    \label{HanChuan}

\end{table*}

\begin{table*}[h]
    \centering
    \caption{Per-class classification accuracies of various methods on the HongHu dataset.}
    \begin{threeparttable}
    \resizebox{\linewidth}{!}{
    \begin{tabular}{cccccccccccc}
    \hline
        Class & MSDN \cite{MSDN} & SSFCN \cite{ssfcn} & FullyContNet \cite{fullycontnet} & SpectralFormer \cite{spectralformer} & HSIC-FM \cite{HSIC-FM} & SSGRN \cite{ssgrn} & CSIL \cite{yang_csil} & IDCN \cite{idcn} & CLOLN \cite{CLOLN} & SpatSIGMA & HyperSIGMA \\ \hline
        1 & 73.82  & 80.74  & 78.29  & 94.51  & 66.32  & 96.73  & 93.05&  94.27 & 95.20  & 98.16  & 98.10  \\
        2 & 89.69  & 78.79  & 60.18  & 91.06  & 85.91  & 79.86  & 95.90&  92.06 & 80.97  & 76.91  & 61.49  \\
        3 & 64.26  & 67.70  & 85.21  & 79.22  & 75.81  & 91.26  & 93.96&  81.48 & 98.00  & 95.27  & 97.22  \\ 
        4 & 90.11  & 81.41  & 73.83  & 93.12  & 73.81  & 85.84  & 91.47&  94.74 & 99.48  & 99.71  & 99.71  \\ 
        5 & 66.95  & 60.53  & 92.27  & 89.18  & 68.98  & 99.91  & 98.87&  92.02 & 68.49  & 75.39  & 74.17  \\
        6 & 88.81  & 76.67  & 68.22  & 89.66  & 82.36  & 69.17  & 93.98&  94.35 & 96.32  & 96.67  & 98.30  \\ 
        7 & 34.58  & 48.47  & 26.87  & 64.63  & 44.84  & 62.22  & 79.68&  73.56 & 91.40  & 94.90  & 94.75  \\ 
        8 & 34.49  & 34.17  & 81.17  & 62.40  & 58.40  & 98.52  & 98.43&  73.12 & 54.66  & 89.38  & 90.82  \\ 
        9 & 87.52  & 76.45  & 56.46  & 95.01  & 74.54  & 89.46  & 94.27&  94.75 & 89.31  & 98.78  & 95.30  \\
        10 & 73.75  & 58.24  & 66.86  & 65.30  & 79.34  & 58.49  & 93.03& 79.07 & 84.71  & 97.34  & 96.76  \\ 
        11 & 41.96  & 39.76  & 51.95  & 57.98  & 38.83  & 68.51  & 85.52& 79.78  & 66.22  & 88.16  & 87.71  \\
        12 & 63.45  & 60.67  & 67.14  & 68.95  & 59.34  & 88.85  & 94.19& 77.52  & 69.82  & 77.51  & 89.24  \\
        13 & 68.56  & 50.10  & 56.42  & 65.17  & 54.05  & 74.81  & 89.42& 66.16  & 80.22  & 81.72  & 92.23  \\
        14 & 64.67  & 70.85  & 70.70  & 79.28  & 73.51  & 84.84  & 96.10& 93.22  & 78.09  & 94.88  & 85.45  \\ 
        15 & 90.44  & 78.99  & 93.03  & 95.55  & 92.40  & 100.00  & 99.96& 96.34  & 44.52  & 69.35  & 59.97  \\
        16 & 94.65  & 81.44  & 66.56  & 90.75  & 87.12  & 98.98  & 97.51&  93.95 & 97.24  & 95.94  & 96.50  \\
        17 & 89.00  & 84.21  & 92.71  & 90.44  & 90.02  & 99.96  & 99.27&  95.17 & 88.21  & 96.73  & 82.55  \\
        18 & 90.21  & 65.99  & 82.70  & 90.05  & 67.97  & 97.12  & 98.58&  93.79 & 77.60  & 87.03  & 83.30  \\
        19 & 65.36  & 52.88  & 66.13  & 87.60  & 60.16  & 77.47  & 91.04& 90.70  & 70.47  & 83.77  & 84.64  \\ 
        20 & 88.86  & 63.27  & 93.02  & 94.83  & 74.19  & 99.33  & 99.63&  93.74 & 67.00  & 80.40  & 84.95  \\ 
        21 & 59.57  & 66.82  & 99.53  & 91.94  & 86.25  & 100.00  & 100.00& 94.32  & 66.89  & 83.65  & 79.08  \\
        22 & 86.60  & 79.98  & 93.03  & 93.65  & 68.44  & 99.42  & 98.20& 95.25  & 56.61  & 76.60  & 85.84  \\
        \hline
        OA (\%) & 78.55  & 71.62  & 67.12  & 85.33  & 70.47  & 82.19  & 91.86 & 89.19 & 87.89  & 94.35  & 94.87  \\
        AA (\%) & 73.06  & 66.51  & 73.81  & 83.20  & 71.03  & 87.31  & 94.64  & 88.15 & 78.25  & 88.10  & 87.18  \\
        Kappa (\%) & 73.45  & 65.67  & 61.34  & 81.76  & 64.73  & 78.21  & 89.89 &  86.59 & 84.83  & 93.04  & 93.69 \\ \hline
    \end{tabular}
    }
\begin{tablenotes}
  \scriptsize
  \item[1] The classes are as follows: 1. Red roof, 2. Road, 3. Bare soil, 4. Cotton, 5. Cotton firewood, 6. Rape, 7. Chinese cabbage, 8. Pakchoi, 9. Cabbage, 10. Tuber mustard, \\11. Brassica parachinensis, 12. Brassica chinensis, 13. Small Brassica chinensis, 14. Lactuca sativa, 15. Celtuce, 16. Film covered lettuce, 17. Romaine lettuce, \\18. Carrot, 19. White radish, 20. Garlic sprout, 21. Broad bean, 22. Tree.
\end{tablenotes}
\end{threeparttable}
    \label{HongHu}
\end{table*}

\begin{table*}[h]
    \centering
    \caption{Per-class classification accuracies of various methods on the Houston dataset.}
    \begin{threeparttable}
    \resizebox{\linewidth}{!}{
    \begin{tabular}{cccccccccccc}
    \hline
        Class & MSDN \cite{MSDN} & SSFCN \cite{ssfcn} & FullyContNet \cite{fullycontnet} & SpectralFormer \cite{spectralformer} & HSIC-FM \cite{HSIC-FM} & SSGRN \cite{ssgrn} & CSIL \cite{yang_csil} & IDCN \cite{idcn} & CLOLN \cite{CLOLN} & SpatSIGMA & HyperSIGMA \\ \hline
        1 & 96.62 & 81.89 & 50.21 & 81.92 & 76.29 & 74.64 & 78.44 & 90.64 & 97.38 & 83.81 & 81.22 \\ 
        2 & 92.43 & 92.61 & 39.63 & 93.83 & 62.75 & 75.85 & 72.93 & 87.22 & 98.51 & 84.93 & 86.76 \\ 
        3 & 20.06 & 62.05 & 91.88 & 94.59 & 32.81 & 83.96 & 97.62 & 84.98 & 54.84 & 97.27 & 98.85 \\ 
        4 & 96.53 & 95.52 & 42.93 & 94.13 & 60.13 & 94.60 & 66.29 & 95.99 & 97.80 & 83.36 & 84.57 \\ 
        5 & 83.44 & 85.80 & 56.53 & 97.95 & 83.49 & 94.41 & 93.21 & 94.69 & 99.65 & 100.00 & 100.00 \\ 
        6 & 65.21 & 77.86 & 72.96 & 88.11 & 59.44 & 93.01 & 94.64 & 85.16 & 91.71 & 98.46 & 99.08 \\ 
        7 & 89.41 & 75.78 & 61.13 & 81.56 & 70.37 & 71.08 & 74.16 & 74.66 & 89.36 & 75.55 & 76.80 \\ 
        8 & 62.23 & 62.93 & 19.12 & 62.27 & 28.58 & 57.64 & 29.60 & 69.66 & 93.59 & 76.13 & 81.13 \\ 
        9 & 88.61 & 65.72 & 40.98 & 59.08 & 63.17 & 45.23 & 48.54 & 84.88 & 90.59 & 80.83 & 73.40 \\ 
        10 & 57.21 & 55.18 & 53.80 & 55.21 & 30.53 & 52.32 & 64.90 & 87.95 & 89.02 & 69.44 & 70.94 \\ 
        11 & 69.53 & 51.93 & 84.95 & 63.28 & 37.95 & 52.85 & 63.98 & 75.60 & 74.61 & 99.76 & 97.09 \\ 
        12 & 71.96 & 65.42 & 37.56 & 65.42 & 53.19 & 66.86 & 58.92 & 79.80 & 81.11 & 99.68 & 100.00 \\ 
        13 & 68.79 & 70.06 & 66.20 & 60.23 & 53.45 & 74.74 & 58.60 & 73.40 & 92.79 & 81.88 & 74.20 \\ 
        14 & 78.40 & 85.83 & 98.92 & 92.44 & 70.04 & 100.00 & 99.87 & 82.76 & 91.39 & 100.00 & 100.00 \\ 
        15 & 81.29 & 55.04 & 19.80 & 95.07 & 17.34 & 25.37 & 32.84 & 96.17 & 90.59 & 100.00 & 100.00 \\ 
        \hline
OA & 72.18 & 72.39  & 51.07 & 77.21 & 54.43 & 68.62 &  66.11 & 85.34 & 85.95  & 87.33 & 86.80 \\
AA & 74.78 &  72.23 & 55.77 & 79.00  & 53.30 & 70.84 &  68.96  & 86.94 & 88.86  &  88.74 & 88.27 \\
Kappa & 69.93  &  70.12 & 47.00 & 75.32 & 50.80 & 65.99 & 63.21  & 84.11 & 84.78  & 86.30 & 85.72 \\
  \hline
    \end{tabular}
    }
\begin{tablenotes}
  \scriptsize
  \item[1] The classes are as follows: 1. Healthy grass, 2. Stressed grass, 3. Synthetic grass, 4. Trees, 5. Soil, 6. Water, 7. Residential, 8. Commercial, 9. Road, 10. Highway, \\11. Railway, 12. Parking Lot 1, 13. Parking Lot 2, 14. Tennis Court, 15. Running Track. 
\end{tablenotes}
\end{threeparttable}
    \label{Houston}
\end{table*}

\begin{table*}[h]
    \centering
    \caption{Per-class classification accuracies of various methods on the ZY1-02D dataset.}
    \begin{threeparttable}
    \resizebox{\linewidth}{!}{
    \begin{tabular}{cccccccccccc}
    \hline
        Class & MSDN \cite{MSDN} & SSFCN \cite{ssfcn} & FullyContNet \cite{fullycontnet} & SpectralFormer \cite{spectralformer} & HSIC-FM \cite{HSIC-FM} & SSGRN \cite{ssgrn} & CSIL \cite{yang_csil} & IDCN \cite{idcn} & CLOLN \cite{CLOLN} & SpatSIGMA & HyperSIGMA \\ \hline
        1 & 85.39 & 72.02 & 54.78 & 41.49 & 90.45 & 77.45 & 79.58 & 93.59 & 87.05 & 88.49 & 90.57 \\ 
        2 & 81.28 & 70.24 & 68.69 & 32.99 & 68.25 & 77.89 & 89.49 & 91.47 & 87.35 & 92.01 & 94.86 \\ 
        3 & 99.82 & 94.04 & 94.28 & 96.34 & 80.16 & 83.80 & 97.83 & 96.43 & 97.46 & 98.62 & 99.24 \\ 
        4 & 94.60 & 86.27 & 77.00 & 50.59 & 89.91 & 78.52 & 89.79 & 95.07 & 49.06 & 98.10  & 92.25 \\ 
        5 & 93.83 & 89.24 & 90.32 & 89.22 & 88.67 & 77.09 & 97.75 & 94.92 & 94.24 & 97.30  & 98.20  \\ 
        6 & 96.61 & 91.67 & 94.53 & 86.72 & 85.68 & 89.84 & 100.00 & 96.29 & 69.87 & 100.00   & 100.00   \\ 
        7 & 54.76 & 56.58 & 63.89 & 32.69 & 49.85 & 62.10 & 76.67 & 76.56 & 84.51 & 78.55 & 70.33 \\ 
        8 & 90.48 & 79.05 & 81.43 & 55.24 & 69.52 & 90.00 & 93.81 & 98.00    & 66.16 & 88.00    & 100.00   \\ 
        9 & 95.49 & 95.15 & 95.58 & 49.74 & 97.79 & 89.54 & 93.88 & 93.90  & 92.32 & 100.00   & 99.74 \\ 
        10 & 87.17 & 73.99 & 96.88 & 59.95 & 69.74 & 73.08 & 91.72 & 95.18 & 80.06 & 95.59 & 95.31 \\ 
        11 & 93.71 & 80.70 & 72.37 & 73.39 & 86.40 & 80.70 & 82.02 & 96.95 & 78.89 & 94.71 & 85.96 \\ 
        12 & 78.41 & 62.97 & 72.48 & 32.95 & 50.34 & 81.06 & 86.77 & 88.70  & 81.68 & 96.13 & 94.95 \\ 
        13 & 99.46 & 88.44 & 93.28 & 92.47 & 89.78 & 94.35 & 98.39 & 98.39 & 64.97 & 92.30  & 100.00   \\ 
        14 & 81.50 & 72.83 & 86.92 & 62.06 & 52.71 & 68.29 & 81.03 & 80.50  & 90.81 & 91.94 & 96.34 \\ 
        15 & 78.88 & 79.65 & 73.26 & 55.62 & 40.50 & 100.00 & 94.57 & 88.59 & 16.46 & 100.00   & 100.00   \\ 
        16 & 99.19 & 98.21 & 100.00 & 75.61 & 66.02 & 99.02 & 100.00 & 100.00   & 17.04 & 100.00   & 100.00   \\ 
        17 & 96.46 & 83.78 & 100.00 & 74.93 & 68.44 & 100.00 & 100.00 & 99.64 & 94.24 & 100.00   & 100.00   \\ 
        18 & 70.96 & 65.27 & 76.42 & 42.39 & 68.52 & 63.41 & 85.83 & 81.51 & 45.13 & 93.63 & 100.00   \\ 
        19 & 98.77 & 96.31 & 86.90 & 95.51 & 96.71 & 59.50 & 98.41 & 93.63 & 92.43 & 99.88 & 100.00   \\ 
        20 & 85.83 & 75.97 & 95.00 & 61.39 & 85.69 & 100.00 & 88.75 & 98.48 & 87.97 & 97.72 & 96.25 \\ 
        \hline
OA & 88.57 &  82.46 & 84.47 & 72.03 & 76.98 & 77.46 &  92.49 & 92.29 & 80.43 & 94.72 & 94.92 \\
AA & 88.13 &  80.61 & 83.69 & 63.06  & 74.75 & 82.28 & 91.31 & 92.89 & 73.88 & 95.70 & 93.95 \\
Kappa & 86.47  & 79.36  & 81.65 & 67.20  &  73.31 & 73.80  & 91.09 & 90.82 & 76.91 & 93.74 & 95.15 \\
\hline
    \end{tabular}
    }
\begin{tablenotes}
  \scriptsize
   \item[1] The classes are as follows: 1. Reed, 2. Spartina alterniflora, 3. Clear sea area, 4. Tamarix, 5. Turbid sea area, 6. Oil field, 7. Dry pond, 8. Bare land, 9. Salt field \\ crystallization pool,  10. Swamp, 11. Ttidal ditch, 12. Cultivated land, 13. Pit and pond, 14. Salt field sedimentation tank, 15. Sparse reed, 16. Black locust, \\17. Fish pond, 18. Suaeda, 19. River, 20. Salt field evaporation pond.
\end{tablenotes}
\end{threeparttable}
    \label{ZY1-02D}
\end{table*}

\end{document}